\documentclass[twocolumn]{ICCKjournal} 
\usepackage{amsmath}
\usepackage{amsfonts}
\usepackage{amsmath}
\usepackage{amssymb}
\usepackage[none]{hyphenat}
\usepackage{enumitem}
\usepackage{tikz}
\usepackage{lipsum}
\usepackage{multicol}
\usepackage{float}
\usepackage[ruled]{algorithm2e}
\usepackage{xcolor}
\newcommand{\major}[1]{{\color{black}{#1}}}
\usetikzlibrary{positioning,arrows.meta,shapes,calc}
\usetikzlibrary{trees}
\usetikzlibrary{matrix,positioning}
\usetikzlibrary{positioning,arrows.meta,shapes,calc}
\usepackage{graphicx}

\usetikzlibrary{shapes.geometric, arrows.meta, positioning, fit, calc, backgrounds}
\usetikzlibrary{arrows.meta, positioning, calc}
\usetikzlibrary{arrows.meta}
\graphicspath{{res/},{media/},{figures/}}

\AtBeginEnvironment{thebibliography}{\small}
\usepackage{newfloat,array}
\newcolumntype{P}[1]{p{#1}<{\raggedright}}
\DeclareFloatingEnvironment[fileext=prc,placement={htb},name=Procedure]{prc}
\newcounter{procedure}
\NewDocumentEnvironment{procedure}{O{ht} m}{%
\begin{prc}[#1]
\caption{#2}
\begin{mdframed}[%
  backgroundcolor=black!10!white,
  font=\ttfamily,
  roundcorner=2pt]
}{\end{mdframed}\end{prc}}

\articletype{Review}
\submitdate{17 December 2025}
\acceptdate{13 January 2026}
\pubdate{23March 2026}

\jnlsetup{%
    author  = {Cui, Y., Liu, P., \& Chen, L.},
    journal = {Journal of Geoscience and Remote Sensing},
    title   = {Mixture-of-Experts in Remote Sensing: A Survey},  
    ctitle  = {Mixture-of-Experts in Remote Sensing: A Survey},  
    volume  = {1},
    number  = {1},
    year    = {2026},
    pages   = {4-38},
    doi     = {10.62762/JGRS.2025.140654}
}

\makeatother
\begin{document}
\author[1,2]{Yongchuan Cui}[0009-0002-5179-6232]
\author[1,2][liupeng202303@aircas.ac.cn]{Peng Liu}[0000-0003-3292-8551]
\author[1,2]{Lajiao Chen}

\affil[1]{Aerospace Information Research Institute, Chinese Academy of Sciences, Beijing 100094, China}
\affil[2]{School of Electronic, Electrical and Communication Engineering, University of Chinese Academy of Sciences, Beijing 101408, China}


\ifproof
\input{proof}
\fi

\maketitle
\abstract{Remote sensing data analysis and interpretation present unique challenges due to the diversity in sensor modalities and spatiotemporal dynamics of Earth observation data. Mixture-of-Experts (MoE) model has emerged as a powerful paradigm that addresses these challenges by dynamically routing inputs to specialized experts designed for different aspects of a task. However, despite rapid progress, the community still lacks a comprehensive review of MoE for remote sensing. This survey provides the first systematic overview of MoE applications in remote sensing, covering fundamental principles, architectural designs, and key applications across a variety of remote sensing tasks. The survey also outlines future trends to inspire further research and innovation in applying MoE to remote sensing.}


\begin{figure}[b]
	\citationblock
	\vspace*{-.98em}
\end{figure}


\section{Introduction}

Remote sensing has become an indispensable information source for observing the Earth's surface and atmosphere, supporting applications such as land-cover and land-use mapping, agriculture and forestry monitoring, urban planning, climate and environmental surveillance, and disaster management~\cite{zhu2017deeplearning,ma2019deeprs,zhao2023lulcreview}. Modern remote sensing archives cover multiple sensor modalities (optical, Synthetic Aperture Radar (SAR), Light Detection and Ranging (LiDAR), multispectral, hyperspectral), spatial resolutions from sub-meter to kilometre scale, and dense temporal sampling from days to decades~\cite{gao2020fusion,osco2021uav,huang2025rsfm}. This diversity and scale create rich opportunities for detailed Earth observation, but they also pose significant challenges for classical Machine Learning (ML) and standard Deep Learning (DL) pipelines. In particular, the same land-cover class may exhibit very different signatures across sensors, resolutions, seasons and viewing geometries, while different classes can appear very similar in a single modality or band~\cite{zhu2017deeplearning,ma2019deeprs}. Handling such heterogeneity, domain shifts and long-tailed label distributions remains a central problem in operational remote sensing systems, where traditional ML approaches often struggle to generalize across diverse acquisition conditions~\cite{jiang2022cdsurvey,shafique2022cdsurvey,ding2025cddatascarce,peng2025cdoptical}.

Over the past decade, DL has substantially improved performance in many remote sensing tasks by leveraging powerful generic architectures originally developed for Computer Vision (CV) and pattern recognition. Convolutional Neural Networks (CNNs) such as AlexNet~\cite{alexnet} and ResNet~\cite{he2016resnet}, fully convolutional networks for dense prediction~\cite{long2015fcn}, and U-Net-style encoder–decoder models~\cite{unet} have all been adapted to high-resolution aerial and satellite imagery for classification, detection and segmentation~\cite{zhu2017deeplearning,ma2019deeprs,zhao2023lulcreview}. More recently, Transformer-based models and large-scale pre-training have further improved representation quality and cross-task transfer, following developments such as the Vision Transformer and attention-based sequence modeling~\cite{vaswani2017attention,raffel2020t5,huang2025rsfm}. Nevertheless, most existing remote sensing models are still trained as monolithic networks for relatively narrow settings: a fixed set of sensors, a limited range of geographic regions, or a single family of tasks. As a result, they often struggle to generalize across sensors, domains and tasks without extensive re-training or careful domain adaptation~\cite{gao2020fusion,ding2025cddatascarce}.

Mixture-of-Experts (MoE) models provide an alternative way to increase model flexibility and capacity while retaining efficiency. The basic idea, introduced in the early 1990s, is to decompose a complex prediction problem into simpler sub-problems that are handled by multiple expert networks, with a gating function that assigns input-dependent weights or responsibilities to each expert~\cite{jacobs1991adaptive,jordan1994hme,miller1997moe}. Adaptive mixtures of local experts and hierarchical mixtures of experts formalized this \textit{divide-and-conquer} principle in a probabilistic framework, leading to theoretical results on approximation properties, identifiability and consistency for mixtures-of-experts and related generalized linear models~\cite{jiang1999hmeglm,yuksel2012twenty,nguyen2018overview,chamroukhi2017skewt,fung2022moe}. Later surveys provide unified treatments of MoE from both practical and theoretical viewpoints and emphasize that the key design choices concern the form of the experts, the gating strategy, and the way experts are regularized and trained jointly~\cite{yuksel2012twenty,nguyen2018overview}. Compared with a single dense network, an MoE architecture can allocate different subsets of parameters to different regions of the input space, classes, modalities or tasks, making it a natural candidate for heterogeneous remote sensing data.

In recent years, MoE has re-emerged as a central mechanism for scaling up deep neural networks in Natural Language Processing (NLP) and CV. Sparsely-gated MoE layers~\cite{shazeer2017sparsemoe} and GShard~\cite{lepikhin2021gshard} demonstrated that conditional computation can decouple parameter count from per-example computation by activating only a small subset of experts for each token. Switch Transformers~\cite{SwitchTransformers} and GLaM~\cite{du2022glam} further refined gating and load-balancing schemes to train language models with hundreds of billions to over a trillion parameters at practical cost, while MoE via shallow embedding explored channel-wise MoE routing within CNNs~\cite{wang2020deepmoe}. For multi-task and multi-modal learning, M$^3$ViT~\cite{liang2022m3vit} integrates MoE layers into Vision Transformers~\cite{ViT} to reduce interference between tasks while keeping inference efficient, Mod-Squad~\cite{chen2023modsquad} treats experts as reusable modules that can be shared or specialized across tasks, and MoE-based semantic segmentation frameworks use expert combinations to analyze multi-modal inputs~\cite{pavlitskaya2020moesemseg}. At the systems level, DeepSpeed-MoE~\cite{rajbhandari2022deepspeedmoe} provides an end-to-end framework for training and serving very large MoE models efficiently, and recent surveys on MoE in large language models synthesize developments in routing, optimization and deployment~\cite{cai2025llmmoe}. These results indicate that MoE architectures are especially suitable for scenarios with heterogeneous data and tasks, where conditional computation and expert specialization can be exploited.

\begin{figure}[t]
    \centering
    \includegraphics[width=1\linewidth]{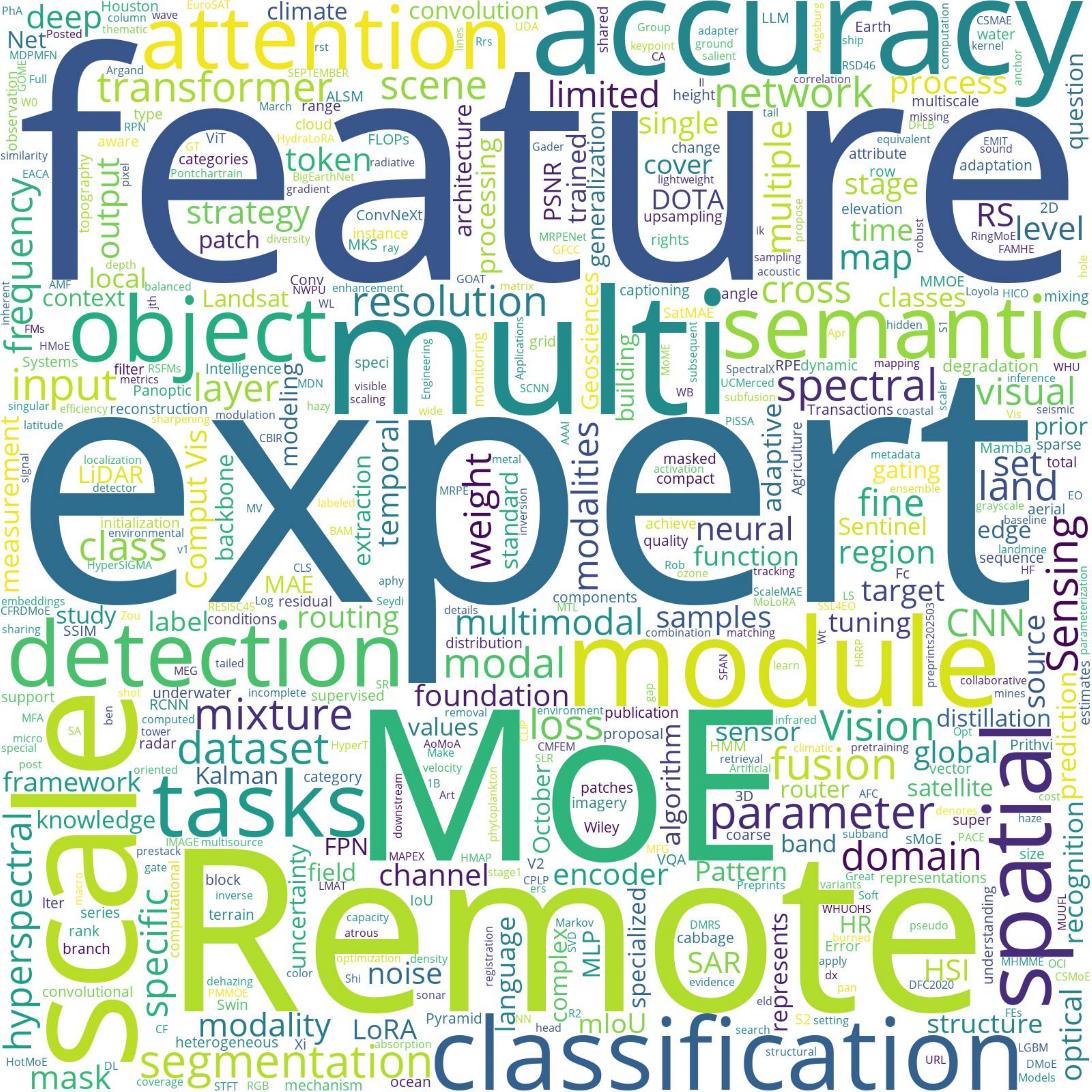}
    \caption{Word cloud of the most frequent words appearing in MoE-related remote sensing papers.}
    \label{fig:wordcloud}
\end{figure}

These properties make MoE particularly attractive for remote sensing, where multi-sensor fusion, long-term time series analysis and multi-task learning are often required in a single workflow. However, explicit MoE usage in remote sensing is still relatively new compared with NLP and general CV. Early work primarily relied on ensembles or multi-classifier systems without an explicit learned gating network~\cite{dou2021tsclassification,jia2018urbanlanduse}, and most deep remote sensing models remain dense. In the last few years, a number of studies have begun to introduce MoE explicitly into remote sensing pipelines. MixtureRS~\cite{liu2025mixturers} replaces dense feed-forward blocks by sparse MoE layers in a cross-modality Transformer for hyperspectral–LiDAR land-use classification. Heterogeneous MoEs architectures have been proposed for remote sensing image super-resolution, with expert groups specialized to different ground-object characteristics and dual routing to adapt reconstruction to local content~\cite{chen2025heterogeneous}. For multi-modal change detection, M$^{2}$CD~\cite{liu2025m} integrates MoE modules into the backbone to explicitly handle the distribution gap between optical and SAR images, while an uncertainty-aware MoE model has been designed to address long-tailed crop type mapping from multi-source imagery~\cite{lu2025uncertaintymoe}. In the context of remote sensing vision–language models, RS-MoE and RSUniVLM~\cite{lin2025rs,liu2024rsunivlm} adopt expert-based routing mechanisms to improve captioning, visual question answering, and multi-granularity reasoning. At an even larger scale, RingMoE~\cite{bi2025ringmoe} proposes a multi-modal MoE foundation model that jointly pre-trains modality-specific and shared experts on massive optical and SAR datasets, and mixture-of-experts networks have also been leveraged for specialized applications such as burned area mapping from multi-temporal satellite imagery~\cite{seydi2024novel}. The word cloud in Fig.~\ref{fig:wordcloud} summarizes the most frequent terms appearing in MoE-related remote sensing papers. \major{The word cloud is based on analysis of 57 core MoE-related remote sensing papers published between 2016 and 2025. Text from paper titles, abstracts, and key technical sections was first tokenized into words using standard word tokenization methods. Word frequency was then computed from the tokenized text, with normalization to account for paper length variations. Standard English stop words were removed using a stop word list to focus on domain-specific technical terms. Literature selection criteria included papers that explicitly apply MoE architectures to remote sensing tasks, published in peer-reviewed journals and conferences as well as recent preprints, given that many MoE-related papers in remote sensing are very recent. The selection covers major application domains (classification, detection, change detection, multi-modal fusion, etc.).} The dominance of words such as \textit{expert}, \textit{feature}, \textit{multi}, \textit{module} and \textit{MoE} confirms that current work is primarily concerned with designing expert modules and feature-processing pipelines, rather than completely new backbone architectures. The prominence of \textit{remote}, \textit{sensing}, \textit{classification}, \textit{detection}, \textit{segmentation}, \textit{semantic} and \textit{object} indicates that most MoE applications concentrate on standard high-level vision tasks (scene and land-cover classification, object detection, semantic segmentation), while terms such as \textit{change}, \textit{time}, and \textit{series} appear but are noticeably smaller, reflecting that temporal and change detection problems are less explored. Frequent occurrence of \textit{multi}, \textit{modal}, \textit{scale}, \textit{spectral}, \textit{spatial} and sensor-related words (\textit{optical}, \textit{SAR}, \textit{hyperspectral}, \textit{LiDAR}) highlights a strong emphasis on multi-modal and multi-scale fusion, where MoE is used to manage heterogeneity across modalities and resolutions. Meanwhile, architectural and training-related terms such as \textit{attention}, \textit{transformer}, \textit{token}, \textit{LoRA}, \textit{foundation}, \textit{uncertainty} and \textit{router} suggest that many methods adapt MoE ideas from large vision/vision–language models and combine them with parameter-efficient tuning or uncertainty modeling. Overall, the word cloud reveals a research landscape in which MoE is mainly deployed as expert-based feature modules for multi-modal, multi-scale classification and segmentation, with relatively fewer works addressing temporal modeling, low-level restoration or large unified foundation models. The MoE paradigm can be adapted to diverse remote sensing problems, but existing works are scattered across tasks and design choices, and a consolidated view is still lacking.

The goal of this survey is to provide a systematic overview of MoE methods for remote sensing, connecting general MoE developments with domain-specific requirements. We first review the fundamentals of MoE architectures, including expert design, gating strategies and training methods, with an emphasis on concepts that are most relevant for geospatial data~\cite{yuksel2012twenty,nguyen2018overview}. We then organize existing remote sensing MoE work by task type (e.g., classification, segmentation, detection, change detection, time series modeling, and vision–language understanding)~\cite{zhu2017deeplearning,ma2019deeprs,huang2025rsfm,liu2025mixturers,chen2025heterogeneous,liu2025m,lu2025uncertaintymoe,lin2025rs,liu2024rsunivlm,bi2025ringmoe,seydi2024novel}. Finally, we discuss open challenges and future directions, including unified multi-modal and multi-task MoE foundations for Earth observation, expert interpretability and analysis, training strategies and efficient deployment on resource-constrained platforms. By linking advances in MoE modeling with the specific characteristics of remote sensing data, we aim to clarify both the current status and the potential of MoE as a general framework for remote sensing analysis.


\section{Fundamentals of Mixture-of-Experts}

\begin{figure}[t]
    \centering
    \includegraphics[width=0.8\linewidth]{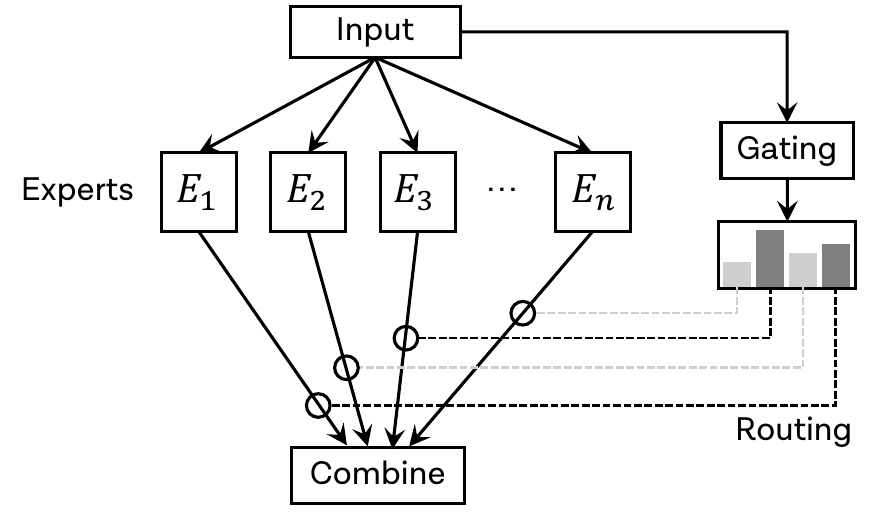}
    \caption{Basic architecture of Mixture-of-Experts (MoE).}
    \label{fig:moe-arch}
\end{figure}

A Mixture-of-Experts (MoE) model consists of a set of expert networks and a gating (routing) network (Fig.~\ref{fig:moe-arch}). The gating network takes an input $x$ (e.g., an image, a token embedding, or a feature vector) and produces nonnegative routing weights $g_1(x),\dots,g_n(x)$ over $n$ experts, typically obtained by normalizing routing scores (e.g., via a softmax). In the dense (soft) case, all experts may be active and the weights satisfy $\sum_{i=1}^n g_i(x)=1$. In the sparse case, only a small subset of experts is selected (e.g., top-$1$ or top-$k$), and the remaining weights are set to zero, i.e., $g_i(x)=0$ for $i\notin \mathcal{S}_k(x)$, where $\mathcal{S}_k(x)$ denotes the selected expert set for input $x$. Formally, if we denote the experts as $E_1,E_2,\dots,E_n$ and the gating function as $G(x)$ producing weights $g_1(x),\dots,g_n(x)$, the MoE output is a typically sparse weighted mixture of expert outputs:
\begin{equation}
\begin{aligned}
y \;&=\;\;\; \sum_{i=1}^{n} g_i(x)\,E_i(x) \\
\;&=\; \sum_{i\in\mathcal{S}_k(x)} g_i(x)\,E_i(x),
\end{aligned}
\end{equation}
where the second equality emphasizes the common sparse-routing implementation. Here $g_i(x)$ is the routing weight assigned to expert $i$ for input $x$. In practice, many MoE models use sparse gating to keep computation efficient even when the total number of experts $n$ is large, e.g., selecting the top-$1$ or top-$k$ experts per input~\cite{SwitchTransformers}. Early MoE formulations used softmax gating trained via Expectation-Maximization (EM)~\cite{jordan1994hme}, whereas modern implementations often use hard or top-$k$ routing trained end-to-end with backpropagation, and may add noise or regularization to encourage load balancing across experts~\cite{shazeer2017sparsemoe,SwitchTransformers}. A commonly used auxiliary objective for this purpose is a load-balancing loss $\mathcal{L}_{\text{balance}}$, which penalizes uneven expert utilization aggregated over a minibatch or the data distribution. Let
\begin{equation}
u_i \;=\; \mathbb{E}_{x\sim\mathcal{B}}\!\left[g_i(x)\right],
\end{equation}
denote the expected routing importance of expert $i$ over a minibatch $\mathcal{B}$, and let $\mathbf{u}=(u_1,\dots,u_n)$. One simple imbalance measure is the Coefficient of Variation (CV) computed across experts:
\begin{equation}
\mathcal{L}_{\text{balance}}
\;=\;
\mathrm{CV}(\mathbf{u})
\;=\;
\frac{\sqrt{\mathrm{Var}_i(\mathbf{u})}}{\mathbb{E}_i[\mathbf{u}]},
\end{equation}
where $\mathrm{Var}_i(\cdot)$ and $\mathbb{E}_i[\cdot]$ denote the variance and the expectation taken over the expert index $i$, respectively. When utilization differs strongly across experts, $\mathrm{CV}(\mathbf{u})$ becomes large, yielding a larger balancing penalty in the overall training objective and thus discouraging the gate from collapsing to a small subset of experts. For training stability, one often uses the squared coefficient of variation.





\subsection{Taxonomy of MoE}

Classical MoE models were introduced as modular regressors or classifiers, in which a set of expert models is combined by a gating network that assigns input-dependent weights to each expert. Early work such as adaptive mixtures of local experts and hierarchical mixtures of experts modeled each expert as a generalized linear model or shallow neural network, with the gate producing a softmax distribution over all experts and parameters estimated by variants of the EM algorithm or maximum likelihood~\cite{jacobs1991adaptive,jordan1994hme,jiang1999hmeglm}. These works mainly focused on approximation properties, statistical consistency, and small- to medium-scale tasks, and treated MoE as a probabilistic mixture with dense activation of the expert set~\cite{yuksel2012twenty,masoudnia2014survey,nguyen2018overview}. Modern MoE systems extend this formulation along the aspect of model scale and sparsity. The transition from dense probabilistic MoE to sparse deep MoE can be described in terms of routing granularity, expert size, number of experts, and layer placement in Large Language Models (LLMs)~\cite{cai2025llmmoe,cai2024survey,gan2025bigdata,dimitri2025survey}. Sparsely-gated MoE layers activate only a small subset of experts (typically top-$k$ by the gating score) for each token, thereby decoupling total parameter count from per-token computation cost. Representative language and vision models include the sparsely-gated MoE layer~\cite{shazeer2017sparsemoe}, GShard~\cite{lepikhin2021gshard}, Switch Transformers~\cite{SwitchTransformers}, GLaM~\cite{du2022glam}, V-MoE~\cite{riquelme2021vmoe} and LIMoE~\cite{mustafa2022limoe}, which demonstrate that conditional computation allows models with hundreds of billions to trillions of parameters to be trained at reasonable computational budgets.

Current MoE research also focuses on how experts are organized around tasks, domains and modalities. In multi-task and multi-objective settings, multi-gate MoE (MMoE) and progressive layered extraction (PLE) architectures use separate gating networks per task, together with shared and task-specific experts, to encourage both parameter sharing and task specialization in recommendation and advertising systems~\cite{ma2018mmoe,tang2020ple}. Follow-up work shows that sparsely activated MoE with task-aware routing can improve transfer to low-resource tasks and robustness when many tasks are trained jointly~\cite{gupta2022sparsemtl,chen2023modsquad,kudugunta2021taskmoe}. In the multimodal setting, MoE backbones such as V-MoE~\cite{riquelme2021vmoe}, LIMoE~\cite{mustafa2022limoe} and Uni-Perceiver-MoE~\cite{zhu2022uniperceiver} combine image and text streams or more general modalities through shared Transformer~\cite{vaswani2017attention} layers with MoE feed-forward blocks, where certain experts become specialized to modalities or sub-domains. For LLMs, architectures like DeepSeekMoE~\cite{dai2024deepseekmoe}, Mixtral~\cite{jiang2024mixtral} and OLMoE~\cite{muennighoff2024olmoe} position MoE layers inside decoder-only Transformers~\cite{vaswani2017attention} to obtain large effective capacity with relatively small active parameter counts per token.

System-level and implementation choices are also crucial for making large MoE models trainable and deployable. Works such as FastMoE~\cite{he2021fastmoe}, DeepSpeed-MoE~\cite{rajbhandari2022deepspeedmoe}, Tutel~\cite{hwang2023tutel}, HetuMoE~\cite{nie2022hetumoe}, FasterMoE~\cite{he2022fastermoe} and MegaBlocks~\cite{gale2023megablocks} focus on efficient distributed training and inference, including all-to-all communication scheduling, CUDA kernel optimizations, expert placement, and parallelism strategies. Pre-gated MoE and related work co-design the algorithm and system by simplifying routing at runtime to reduce memory movement and improve latency, while preserving the advantages of conditional computation~\cite{hwang2024pregatedmoe}. Other works such as BASE layers~\cite{lewis2021base}, hash layers~\cite{roller2021hash} and sparse-upcycling~\cite{komatsuzaki2023sparseupcycling} show that MoE-like conditional computation and expert modularity can be realized either by routing dense blocks or by converting pre-trained dense models into sparse expert collections. Current MoE research can be viewed as varying along model sparsity and scale, task and modality organization, and system realization. The following subsections focus on expert specialization, gating and training within this taxonomy.

\subsection{Expert Specialization}

In classical MoE models, expert specialization is usually interpreted as a partition of the input or covariate space into regions, each handled by a different local model. Adaptive mixtures of local experts and hierarchical mixtures of experts show that the gating network tends to assign nearby inputs to the same expert or leaf in a tree, and theoretical results establish approximation and consistency properties for such hierarchical mixtures-of-experts under suitable assumptions~\cite{jacobs1991adaptive,jordan1994hme,jiang1999hmeglm,yuksel2012twenty,masoudnia2014survey}. These works mostly use low-dimensional inputs and relatively small networks, but already make clear that a successful MoE should distribute data among experts in a way that yields simpler local mappings and avoids redundant experts.

In deep sparse MoE models, specialization emerges at the level of high-dimensional representations. The sparsely-gated MoE layer~\cite{shazeer2017sparsemoe}, GShard~\cite{lepikhin2021gshard}, Switch Transformers~\cite{SwitchTransformers} and GLaM~\cite{du2022glam} all use token-level routing in Transformer~\cite{vaswani2017attention} feed-forward sublayers, and empirical analyses show that experts tend to specialize to language, syntactic patterns or subsets of tokens, although the degree of specialization can vary with routing hyperparameters and auxiliary losses. Vision and multimodal MoE models such as V-MoE~\cite{riquelme2021vmoe} and LIMoE~\cite{mustafa2022limoe} report that some experts focus on particular visual categories, image resolutions or modalities, while others act as more general experts, and that sparse expert utilization is crucial for scaling to large models without severe redundancy. Analytical studies confirm that, under certain conditions, MoE architectures can learn to partition data into clusters or domains and achieve better robustness and generalization than dense counterparts~\cite{chen2022understanding,ho2022gaussianmoe,nguyen2023demystify}.

Expert specialization is particularly important in multi-task and multi-domain applications. MMoE~\cite{ma2018mmoe} and PLE~\cite{tang2020ple} explicitly factor experts into shared and task-specific groups, with multiple gating networks combining them to capture both common and idiosyncratic structure in recommendation tasks. Sparsely activated multi-task MoEs~\cite{gupta2022sparsemtl} and modular MoE~\cite{chen2023modsquad} designs such as Mod-Squad~\cite{kudugunta2021taskmoe} extend this idea by allowing tasks to share only subsets of experts and by routing different tasks or labels to different expert combinations, which can improve resistance to negative transfer and catastrophic forgetting when new tasks are added. Time-MoE pushes this idea to billion-scale time-series models, where experts specialize to temporal dynamics and domains while still being trained within a unified foundation model~\cite{shi2025timemoe}. In these systems, specialization is typically encouraged through task-aware gating, auxiliary load-balancing losses and regularization, rather than by explicit hard constraints on which tasks an expert may serve. For LLMs, recent work has proposed architectural mechanisms that more directly control specialization. DeepSeekMoE separates shared experts, which are always active and capture common skills such as general syntax, from routed experts, which are selectively activated and encouraged to learn rarer capabilities or domain-specific knowledge~\cite{dai2024deepseekmoe}. Mixtral arranges experts in the feed-forward sublayers of a decoder-only Transformer~\cite{vaswani2017attention} and shows that different experts specialize to language families, code versus natural language, or input length regimes, while using only a small number of active experts per token~\cite{jiang2024mixtral}. OLMoE conducts a detailed routing analysis and reports strong specialization across layers and experts, with some experts focusing on particular capability clusters or subsets of the training data distribution~\cite{muennighoff2024olmoe}. Recent works on MoE in LLMs emphasize that this kind of specialization is a key reason why sparse MoE LLMs can match or surpass dense models with substantially fewer active parameters at inference time~\cite{cai2025llmmoe,cai2024survey,dimitri2025survey}. The success of LLMs in NLP has inspired similar MoE architectures for vision and multimodal tasks, demonstrating that the LLM paradigm can be effectively adapted to other domains.

Parameter-efficient adaptation introduces another form of expert specialization, in which the experts are not full feed-forward blocks but low-rank (LoRA)~\cite{hu2022lora} adapter modules. LoRA represents task-specific updates as low-rank matrices added to existing weights, and several works extend this idea to mixtures of LoRA experts. TT-LoRAMoE~\cite{TTLoRAMoE}, MoLE~\cite{wu2024mole} and HMoRA~\cite{liao2024hmora} treat each adapter or group of adapters as an expert and learn gating functions that select among them based on the input or task, which allows multiple domains or instruction styles to be captured within a single base LLM while keeping parameter overhead small. These methods demonstrate that expert specialization can be realized at the level of adapters as well as full layers, and can be learned during fine-tuning without retraining the base model from scratch.

Theoretical work further clarifies when and how MoE architectures specialize. Studies on convergence rates for Gaussian mixtures-of-experts and on the statistical behavior of softmax and sparse top-$k$ gating provide conditions under which experts consistently approximate different parts of the data-generating function and under which gating parameters can be reliably estimated~\cite{ho2022gaussianmoe,nguyen2023demystify,nguyen2023topk,nguyen2023generalgating}. Together with empirical analyses in large-scale systems~\cite{chen2022understanding,du2022glam,dai2024deepseekmoe,muennighoff2024olmoe}, these results support the view that expert specialization is a central mechanism by which MoE models convert increased parameter count into improved accuracy and robustness.

\subsection{Gating Strategies}

Gating strategies determine how inputs are assigned to experts and are therefore central to the behavior of MoE models. Classical MoE work generally used softmax gating functions that output a full probability distribution over experts, combined with EM or gradient-based optimization. In these models, the output is usually a convex mixture of all experts, and the gate is trained to allocate responsibility for each data point across experts; theoretical analyses show how such softmax gates affect approximation rates and identifiability~\cite{jacobs1991adaptive,jordan1994hme,yuksel2012twenty,masoudnia2014survey,nguyen2018overview,ho2022gaussianmoe}. More recent work on the statistical properties of softmax gating and its variants provides more rigorous guarantees for parameter estimation and convergence in Gaussian and generalized linear MoE models~\cite{nguyen2023demystify,nguyen2023topk,nguyen2023generalgating}.

In deep sparse MoE architectures, the dominant strategy is sparse top-$k$ token-level routing. The sparsely-gated MoE layer uses a linear gating network that outputs logits for each expert, applies a softmax, and then routes each token to the top-$k$ experts (often $k=1$ or $2$), with additional noise and auxiliary load-balancing losses to prevent collapse onto a few experts~\cite{shazeer2017sparsemoe}. GShard~\cite{lepikhin2021gshard}, Switch Transformers~\cite{SwitchTransformers} and GLaM~\cite{du2022glam} adopt similar routing, differing in the exact loss design, capacity constraints and implementation details, but all select a small set of experts per token and rely on gradient-based training of the gate. Later work such as ST-MoE studies the stability of such routers, proposes auxiliary objectives such as the $z$-loss, and analyzes how routing schemes and noise influence expert utilization and transfer performance~\cite{zoph2022stmoe,chen2022understanding}. Routing variations have been proposed to improve load balancing, robustness and flexibility. MoE with expert choice routing reverses the usual perspective and lets experts choose a fixed number of tokens rather than tokens choosing experts; this achieves better load balancing and allows the number of experts per token to vary~\cite{zhou2022expertchoice}. Dynamic routing strategies adapt the number of experts or routing pattern to input difficulty, for example by activating more experts for harder examples~\cite{huang2024harder}. Multi-gate designs such as MMoE~\cite{ma2018mmoe} and PLE~\cite{tang2020ple} use separate gating networks per task or tower, which allows different tasks or label spaces to see different mixtures of shared and task-specific experts while using the same underlying expert pool. Hash-based and static routing methods such as hash layers map tokens to experts using deterministic hash functions to eliminate learned gates, and BASE layers demonstrate that a fixed, random assignment can still achieve strong performance when combined with appropriate regularization and training~\cite{lewis2021base,roller2021hash}.

Recent work also aims to make sparse routing differentiable or more amenable to optimization. DSelect-$k$~\cite{hazimeh2021dselectk} replaces the hard top-$k$ operator with a continuous relaxation based on a differentiable selection mechanism, enabling end-to-end gradient-based learning while still producing sparse expert assignments. Lory~\cite{zhong2024lory} and related fully differentiable MoE architectures merge experts in parameter space and route at the level of segments rather than individual tokens, removing the discrete selection step and simplifying router training in autoregressive language models. Other approaches, such as stochastic experts and locality-sensitive hashing-based routing, treat routing as a stochastic process, using randomness to regularize the model and to reduce communication overhead in large distributed systems~\cite{zuo2022stochastic,nie2022hetumoe}. System-level designs such as Tutel and Pre-gated MoE show that simplifying routing computations, for example by precomputing routing decisions or using hierarchical gating, can substantially improve throughput without changing the overall model structure~\cite{hwang2023tutel,hwang2024pregatedmoe}.

Gating strategies and training objectives are closely linked. ST-MoE and subsequent work report that small changes to gating losses, capacity factors and noise can strongly affect training stability and final performance~\cite{zoph2022stmoe,zhou2022expertchoice,chen2022understanding}. Theoretical analyses of softmax and sparse gating provide conditions under which gradient-based training converges to meaningful partitions, while empirical studies emphasize that practical MoE routers must balance model quality, communication cost and implementation simplicity~\cite{cai2025llmmoe,cai2024survey,gan2025bigdata,dimitri2025survey}.

\subsection{Training Methods}

\major{Training methods for MoE models can be organized along three main dimensions: basic training strategies that establish fundamental load balancing and routing optimization, advanced optimization schemes that introduce sophisticated regularization and balancing mechanisms, and cross-scenario adaptation methods that enable flexible deployment and conversion from dense models.}

\major{Fundamental MoE training relies on standard load balancing and routing optimization approaches. The sparsely-gated MoE layer~\cite{shazeer2017sparsemoe} introduced auxiliary load-balancing losses to prevent expert collapse, using capacity constraints and noise injection to encourage balanced expert utilization. Switch Transformers~\cite{SwitchTransformers} refined these basic strategies with simplified top-$k$ routing and improved load balancing through auxiliary losses that penalize uneven expert usage. GShard~\cite{lepikhin2021gshard} further developed capacity constraints and load balancing mechanisms, ensuring that each expert receives a roughly equal number of tokens per batch. These basic strategies form the foundation for MoE training, focusing on preventing expert collapse and maintaining balanced utilization through simple auxiliary objectives and capacity constraints.}

\major{Recent work on training MoE models increasingly focuses on how routing decisions are made and how expert loads are controlled during optimization. Building on basic training strategies, recent work has developed more sophisticated optimization techniques.Wang et al.\ propose an auxiliary-loss-free load balancing strategy that maintains expert-wise bias terms and updates them based on recent routing statistics, so that expert loads remain balanced without injecting additional gradients into the main objective~\cite{wang2024auxiliarylossfree}. Thaman formulates router balancing as a constrained optimization problem and derives a dual-ascent update with sparsemax gating that enforces target usage ratios while leaving the task loss unchanged~\cite{thaman2025rebalancing}. Omi et al.\ replace uniform balancing losses by a similarity-preserving term that encourages similar tokens to be routed to similar expert sets, which improves convergence speed and reduces redundancy in expert usage~\cite{omi2025similarityrouter}. Complementary to these balancing-oriented approaches, SimSMoE~\cite{do2025simsmoe} explicitly addresses representation collapse by minimizing similarity between expert representations, and shows that such regularization leads to more diverse experts under a fixed FLOPs budget during pretraining and fine-tuning. In parallel, several works use the MoE structure itself as a training regularizer: SMoE-Dropout~\cite{chen2023smoedropout} trains a dense Transformer~\cite{ViT} together with a sparse MoE layer driven by a randomly initialized and frozen router, gradually increasing the number of activated experts so that the final model becomes self-slimmable at inference while mitigating expert collapse during training, and MoEC ~\cite{xie2023moec} introduces expert clusters and variance-based constraints on the routing distribution, together with cluster-level dropout, so that experts within the same cluster specialize to complementary sub-regions of the data and maintain useful diversity even when data per expert is limited. In all of these cases, routing patterns, balancing mechanisms and expert activation schedules are treated as explicit components of the training procedure, rather than as purely passive structures optimized only through the main task loss.} 

\major{Beyond optimizing routing within a fixed MoE architecture, recent work has explored methods for converting dense models to MoE and adapting MoE training to flexible inference-time behavior. On top of changing how routing is optimized, some works study how to convert or upcycle existing dense checkpoints into MoE models and how to adapt MoE training to flexible inference-time behavior and downstream optimization. MoEBERT~\cite{zuo2022moebert} starts from a pre-trained BERT~\cite{BERT}, partitions each feed-forward block into multiple experts, and trains the router with layer-wise distillation so that the converted MoE model matches the dense teacher while enabling sparse inference. D2DMoE~\cite{szatkowski2024d2dmoe} regularizes activation sparsity in a dense model and then converts groups of neurons into experts with a dynamic-$k$ routing rule, learning routers that predict expert contributions and activating a variable number of experts per token to obtain large inference speedups with limited additional training. ToMoE~\cite{gao2025tomoe} converts dense LLMs into MoE models through dynamic structural pruning that exposes experts implicitly present in the dense network and then fine-tunes the router and pruned structure to recover or improve the original accuracy. For text embedding models, Nussbaum and Duderstadt train Nomic Embed~v2~\cite{nussbaum2025trainingsparse} as a sparse MoE encoder using contrastive and distillation objectives, together with routing constraints that ensure balanced expert usage, showing that MoE-style training is also effective for retrieval-style encoders and not only for causal language models. Training procedures have also been adapted to support flexible inference-time configurations and reinforcement-learning fine-tuning: Elastic MoE~\cite{gu2025elasticmoe} explicitly randomizes both the number and composition of active experts during pretraining so that the same checkpoint can later be evaluated with different expert budgets at inference, encouraging experts to collaborate under many activation patterns and enlarging the range over which activating more experts improves accuracy, while Ma et al.\ show that in reinforcement learning for MoE LLMs the discrepancy between routing during training and routing during inference can destabilize policy optimization and address this by recording routing distributions from the inference engine and replaying them during training, which aligns the two phases and prevents collapse during reinforcement learning fine-tuning~\cite{ma2025stabilizingmoerl}. Beyond single-model training, Gururangan et al.\ demonstrate that one can train separate expert language models on clusters of documents and combine them as a sparse ensemble at inference time, effectively realizing a data-driven mixture-of-experts without tightly coupling expert training in a single network~\cite{gururangan2023scalingexpert}, and AutoMoE~\cite{jawahar2023automoe} integrates training with neural architecture search by learning where to place experts, how many experts to allocate and how much computation each token should receive under explicit compute constraints, showing that heterogeneous MoE layouts can be discovered automatically in neural machine translation.}


\begin{figure*}[htbp]
    \centering
    \begin{tikzpicture}[
      grow=right,
      level distance=60mm,
      level 1/.style={level distance=35mm, edge from parent path={(\tikzparentnode.east) -- (\tikzparentnode.east -| \tikzchildnode.west) -- (\tikzchildnode.west)}},
      level 2/.style={level distance=70mm, edge from parent path={(\tikzparentnode.east) -- (\tikzparentnode.east -| \tikzchildnode.west) -- (\tikzchildnode.west)}},
      sibling distance=22mm,
      every node/.style={font=\small},
      domainNode/.style={rectangle, rounded corners, draw, thick, fill=blue!5, text width=3.3cm, minimum height=1.4cm, align=center},
      rootNode/.style={rectangle, rounded corners, draw, thick, fill=blue!5, text width=1.5cm, minimum height=14cm, align=center, font=\large},
      refsNode/.style={rectangle, rounded corners, draw, fill=gray!5,  text width=8cm, align=left}
    ]
    
    \node[rootNode]{\rotatebox{90}{Mixture-of-Experts in Remote Sensing}}
      child {
        node[domainNode, yshift=-22mm]{Other\\Specialized\\Applications}
        child {
          node[refsNode]{
            Loyola et al.~\cite{loyola2006applications},
            Li et al.~\cite{li2022pertinent},
            PhA-MoE~\cite{wang2025pha},
          }
        }
      }
      child {
        node[domainNode, yshift=-24mm]{Image\\Restoration and\\Enhancement}
        child {
          node[refsNode]{
            Chen et al.~\cite{chen2025heterogeneous},
            Swin2-MoSE~\cite{rossi2025swin2},
            PhyDAE~\cite{dong2025phydae},
            Shen et al.~\cite{shen2024spatial},
            He et al.~\cite{he2024frequency}
          }
        }
      }
      child {
        node[domainNode, yshift=-10mm]{Multi-Modal\\Fusion and\\Adaptation}
        child {
          node[refsNode]{
            \textit{Multi-Modal Foundation Models:}\\
            RingMoE~\cite{bi2025ringmoe},
            SkySense V2~\cite{zhang2025skysense},
            MAPEX~\cite{hanna2025mapex}\\[4pt]
            \textit{Vision-Language Models:}\\
            Liu et al.~\cite{liu2023unified},
            RS-UniVLM~\cite{liu2024rsunivlm},
            RS-MoE~\cite{lin2025rs},
            SkyMoE~\cite{SkyMoE}\\[4pt]
            \textit{Cross-Domain Adaptation:}\\
            Land-MoE~\cite{chen2025generalizable},
            AMoED~\cite{fu2025adaptive},
            MEDNet~\cite{lin2021mednet},
            Ngo et al.~\cite{ngo2022collaboration},
            SpectralX~\cite{zhang2025spectralx}\\[4pt]
            \textit{Multi-Sensor Data Fusion:}\\
            Pasika et al.~\cite{pasika1999neural},
            Aggarwal et al.~\cite{aggarwal2004multiple},
            MixtureRS~\cite{liu2025mixturers},
            Kong et al.~\cite{kong2025joint},
            AMoE~\cite{he2025adaptive},
          }
        }
      }
      child {
        node[domainNode, yshift=10mm]{Spatiotemporal\\Modeling and\\Analysis}
        child {
          node[refsNode]{
            \textit{Change Detection:}\\
            M$^2$CD~\cite{liu2025m},
            Seydi et al.~\cite{seydi2024novel}\\[4pt]
            \textit{Time-Series Analysis:}\\
            Jiang et al.~\cite{jiang2025knowledge},
            HotMoE~\cite{sun2025hotmoe},
            MoE-MAE~\cite{albughdadi2025lightweight},
            STF-MoE~\cite{li2025stfmoe}
          }
        }
      }
      child {
        node[domainNode, yshift=12mm]{Object\\Detection}
        child {
          node[refsNode]{
            SAFPN~\cite{chai2025scalable},
            FAMHE-Net~\cite{chen2025famhe},
            MRPENet~\cite{lin2025multiple},
            SM3Det~\cite{li2024sm3det},
            MEDNet~\cite{lin2021mednet},
            CM-MMoE~\cite{zhang2025challenging},
            MEG MoE~\cite{qian2025multi}
          }
        }
      }
      child {
        node[domainNode, yshift=18mm]{Image\\Classification}
        child {
          node[refsNode, yshift=10mm]{
            \textit{Land Use \& Land Cover Mapping:}\\
            Kussul et al.~\cite{2016moeKussul},
            Yuksel et al.~\cite{2012moeLandmine},
            Nagarajan et al.~\cite{nagarajan2009multiscale},
            SparseFormer~\cite{chen2024sparseformer},
            MFIAE~\cite{sun2025multi},
            UNetMoE~\cite{2025moeRenclass},
            MSLoRA-Net~\cite{xu2025multi},
            U-MoE Mamba~\cite{li2025u},
            GRAM~\cite{lee2025generalizableslumdetectionsatellite},
            Land-MoE~\cite{chen2025generalizable},
            DMoE-ViT~\cite{lu2025uncertaintymoe}
            \\[4pt]
            \textit{Scene Classification:}\\
            Xie et al.~\cite{xie2022stacked},
            Mensah et al.~\cite{2024moeWildlife},
            DMRS~\cite{DMRS},
            AMoED~\cite{fu2025adaptive},
            Guo et al.~\cite{guo2025confidence}\\[4pt]
            \textit{Hyperspectral Image Classification:}\\
            HyperTransXNet~\cite{HyperTransXNet},
            MambaMoE~\cite{xu2025mambamoe},
            MixtureRS~\cite{liu2025mixturers},
            MaMOL~\cite{MaMOL}
          }
        }
      }
      ;
    
    \end{tikzpicture}
    \vspace{2mm}
    \caption{Overview of Mixture-of-Experts applications in remote sensing.}
    \label{fig:moe-rs}
\end{figure*}
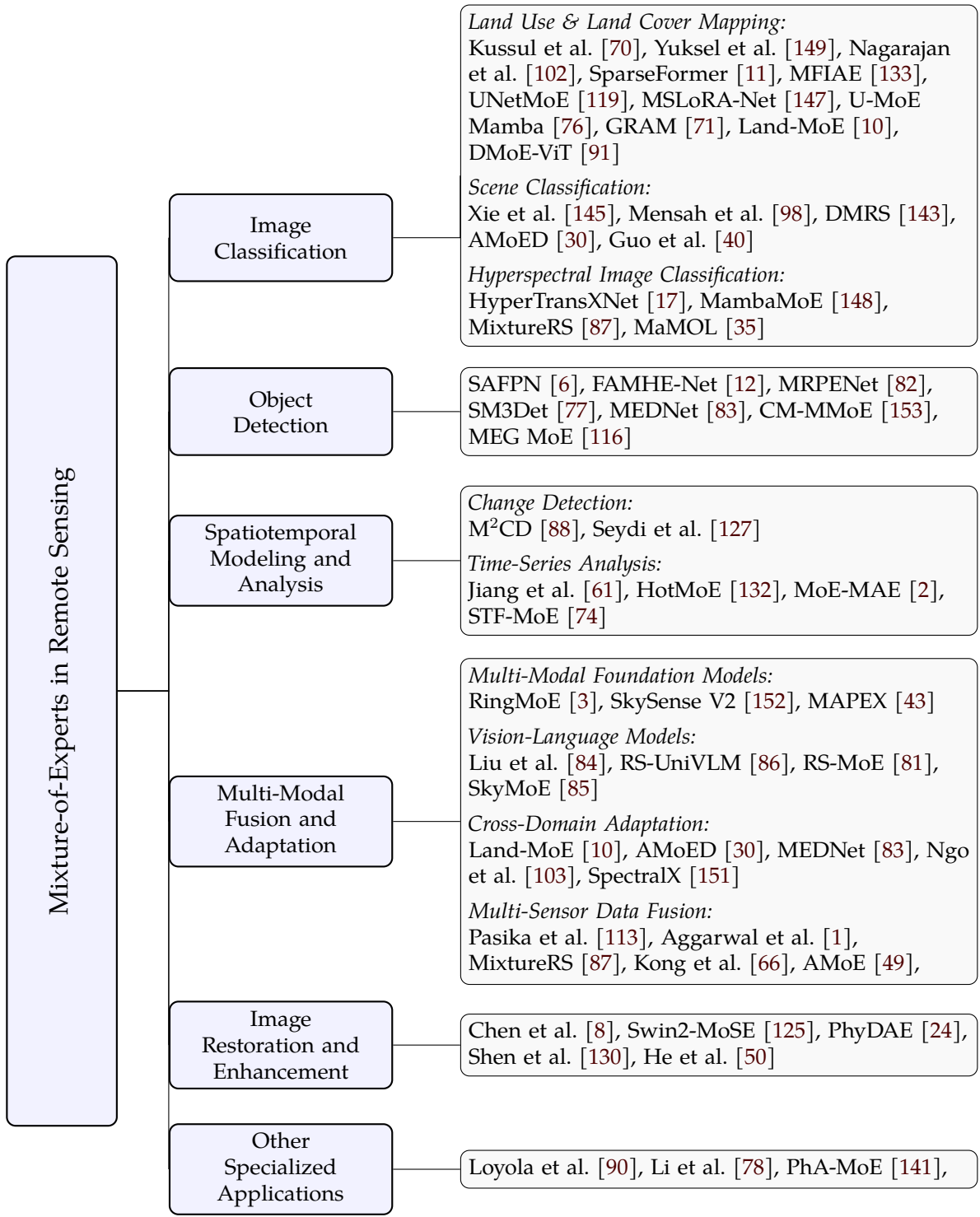

\section{Mixture-of-Experts in Remote Sensing}

Mixture-of-Experts (MoE) models have been applied across a wide range of remote sensing tasks (refer to Fig.~\ref{fig:moe-rs}). In this section, we survey their use in key application domains, including image classification, object detection and segmentation, multi-modal data fusion, change detection and temporal analysis, image restoration, and vision-language tasks. We organize the discussion by task type, highlighting how MoE architectures address specific challenges in each domain and summarizing representative studies. Notably, the idea of combining multiple expert networks in remote sensing is not entirely new, earlier works in the 1990s and 2000s explored multi-network and multi-model approaches for sensor fusion and image analysis~\cite{loyola2006applications,pasika1999neural}, but recent advances in Deep Learning (DL) and sparse gating have greatly expanded MoE's capabilities. Throughout these applications, a common theme is conditional model capacity: MoEs enable the model to activate different subsets of parameters (experts) depending on the input, allowing specialization for diverse data characteristics while keeping overall computation efficient.

\subsection{Image Classification}

Land-cover and scene classification is a foundational task in remote sensing, where each image or each pixel in an image is labeled as a certain category (such as water, urban, forest, agriculture, etc.). This task is challenging because of high intra-class variability, i.e., the same class can appear very different in different conditions, and inter-class similarity, i.e., different land-cover types can have confusingly similar appearances or spectral signatures. Traditional classification approaches struggled with these issues, especially when relying on fixed spectral thresholds or shallow classifiers. Early neural network methods and sensor fusion attempts provided some improvements but were limited by training data and model capacity. The advent of DL such as Convolutional Neural Networks (CNNs)~\cite{alexnet} and Transformers~\cite{ViT} significantly boosted classification accuracy by extracting hierarchical features and learning complex spectral-spatial patterns. However, even advanced DL networks can hit performance plateaus on complex scenes or when faced with very heterogeneous data. While CNNs excel at capturing local spatial patterns, they may struggle with long-range dependencies in large-scale remote sensing imagery. This is where MoE offers a further boost: by partitioning the feature space among specialized expert networks, MoE classifiers can handle a wider variety of inputs more effectively.

\subsubsection{Land Use and Land Cover Mapping}

In Earth observation, the MoE paradigm has been used mainly for pixel-level labeling tasks, including semantic segmentation, land-cover and land-use (LULC) mapping, and other forms of pixel-wise classification. The central idea is to let multiple experts specialize in different features (e.g., sensor types, spatial scales, or scene conditions) and then combine their outputs through a gating mechanism to obtain more accurate and robust pixel predictions.

Early studies already explored MoE-like designs for geospatial pixel labeling. Kussul et al.~\cite{2016moeKussul} employed an ensemble of neural-network experts to produce large-scale land-cover maps. This model first clusters and denoises multi-sensor time series using self-organizing maps, and then feeds each partition into an ensemble of multilayer perceptrons for classification and optical-radar fusion, generating 30-m LULC maps for Ukraine from 1990 to 2010 and 2015. Yüksel and Gader~\cite{2012moeLandmine} considered landmine detection in ground-penetrating radar, modeling each radar trace as a sequence to be classified. They introduced a MoE framework using Hidden Markov Model experts, where each expert specializes in a particular landmine signature or soil condition, and a gating scheme combines their probabilistic outputs to distinguish landmine responses from clutter. Nagarajan and Slatton~\cite{nagarajan2009multiscale} applied an MoE approach to multiscale segmentation of elevation data, e.g. Light Detection and Ranging (LiDAR)-derived Digital Elevation Models (DEMs). In their framework, experts operate at different spatial scales to segment terrain features, and a gating function merges these expert segmentations into a final elevation segmentation map. Taken together, these works show that assigning pixel- or sample-level decisions to multiple specialized experts can improve robustness across sensors, acquisition conditions, and spatial scales.

More recent DL–based MoE architectures adopt the same principle for high-resolution remote sensing imagery. Chen et al.~\cite{chen2024sparseformer} proposed SparseFormer, a dual-CNN-expert-guided transformer for semantic segmentation. The model comprises three branches, where two CNN branches employ different attention mechanisms: convolutional block attention module and coordinate attention, to encourage diverse outputs and extract complementary features, including fine-grained spatial details and global context. A credible assessment mechanism combines the CNN outputs to produce high-quality pseudolabels that supervise a CNN-Transformer hybrid branch, which integrates global representations with local features for precise segmentation. \major{On the Zurich Summer dataset, SparseFormer~\cite{chen2024sparseformer} reports mF1 and mIoU scores of 75.07\% and 64.85\%, respectively, with overall accuracy exceeding other weak supervision approaches, confirming the effectiveness of the dual-CNN-expert design for sparse annotation scenarios.} Sun et al.~\cite{sun2025multi} extended the idea to panoptic segmentation, jointly modeling object instances and semantic regions in remote sensing images. Their multi-scale feature interaction and adaptive experts framework introduces an adaptive disturbance sparse MoE module based on Transformer, where adaptive noise is introduced to disturb expert selection, enhancing randomness and exploration to improve generalization and robustness while reducing computational overhead. He et al.~\cite{he2024moe_semseg} further presented a resource-efficient MoE-based semantic segmentation network for remote sensing images, illustrating that expert specialization can also be leveraged under tight computational constraints. Ren et al.~\cite{2025moeRenclass} proposed UNetMoE, a MoE-based semantic segmentation model for remote sensing images. The model employs category-specific expert submodels, where each expert is trained to perform binary classification for a single land-cover category, and a gating system based on ResNet with self-attention mechanisms dynamically allocates weights to combine expert outputs. This category-specific expert design enables targeted feature extraction for each land-cover type, improving per-class recognition precision while the gating system synthesizes outputs from all experts to optimize overall segmentation accuracy. Complementing these designs, Chen et al.~\cite{chen2025generalizable} introduce Land-MoE, a frequency-aware mixture of low-rank~\cite{hu2022lora} token experts used as parameter-efficient adapters on top of vision foundation models to improve multispectral land-cover classification under cross-sensor and cross-geospatial domain shifts, showing that token-level MoE adapters and shared frequency-aware filters can substantially enhance the generalization of LULC maps across sensors and regions. Building on this idea of region-aware specialization, Lee et al.~\cite{lee2025generalizableslumdetectionsatellite} proposed GRAM, a region-aware MoE framework for slum segmentation that learns city-specific expert adapters on a large multi-city satellite dataset while a shared backbone captures universal informal-settlement morphology, and uses a region classifier plus cross-expert prediction consistency at test time to select reliable pseudo-labels for self-training, thereby improving the generalization of slum maps to previously unseen cities.

Beyond classification, MoE has also been used for specific pixel-wise segmentation applications. For building footprint extraction, Xu et al.~\cite{xu2025multi} proposed MSLoRA-Net, which integrates a LoRA-based~\cite{hu2022lora} MoE into a segmentation model. Low-rank adaptation experts are injected into the layers of a Vision Transformer~\cite{ViT}, and a routing network decides which LoRA~\cite{hu2022lora} expert weights to apply for each input patch. This effectively fine-tunes a large pre-trained model to specialize in building structures, enabling dynamic selection of adaptation parameters for different building shapes and sizes. In agricultural mapping, Lu et al.~\cite{lu2025uncertaintymoe} developed an uncertainty-aware difficulty-based MoE framework (DMoE-ViT) for long-tail crop type mapping, where an autoencoder first estimates sample difficulty via reconstruction loss, crops are stratified into easy, moderate and hard subsets, and three Vision Transformer~\cite{ViT} experts are trained on different difficulty levels and fused by a gating network with evidence-based uncertainty weighting; experiments on heterogeneous agricultural regions show that this design improves classification accuracy and robustness for rare and hard crop samples compared with single-backbone CNN~\cite{alexnet}, LSTM~\cite{lstm}, UNet~\cite{unet} and ViT~\cite{ViT} baselines. In agricultural mapping, Li et al.~\cite{li2025u} introduced U-MoE Mamba~\cite{unet,mamba,mamba2}, a hybrid expert segmentation model for segmenting crop fields (specifically cabbage) from high-resolution drone imagery. The model integrates three expert paradigms, i.e., multi-scale convolution, attention mechanisms, and Mamba pathways through a lightweight gating network that dynamically combines expert outputs for adaptive feature aggregation. By leveraging expert diversity, the model better accommodates seasonal and appearance variations in crops than single-backbone approaches, leading to improved delineation of field boundaries across different growth stages.

\subsubsection{Scene Classification}

Beyond pixel-level labeling, MoE has also been widely explored for scene-level classification and object recognition in remote sensing. In this setting, multiple experts are typically designed to handle different levels of difficulty, class distributions, sensor modalities, or deployment environments, and a gating mechanism aggregates their predictions into a final scene label.

Xie et al.~\cite{xie2022stacked} introduced a stacked MoE network for fast aerial scene classification. In their design, multiple expert subnetworks are organized in a stacked architecture, where later experts refine the representations and predictions produced by earlier ones. By reusing shared features across stages and concentrating expert capacity on more discriminative processing in deeper layers, the model achieves a favorable trade-off between classification accuracy and computational cost for large-scale aerial scene classification. A related idea appears in wildlife monitoring: Mensah et al.~\cite{2024moeWildlife} proposed an MoE-based mobile vision transformer for fine-grained bird species classification on edge devices. Their model modifies MobileViTV2~\cite{mehta2023separable} to include patch-level MoE layers, where a router clusters patch embeddings and routes each image patch to a small subset of transformer experts, enabling conditional computation within a single network. Together, these approaches illustrate how MoE can trade off accuracy and efficiency by routing different parts of the input to specialized experts under resource and deployment constraints.

MoE-style expert frameworks have also been explored to alleviate class imbalance and hard categories in remote sensing recognition. Wang et al. proposed DMRS~\cite{DMRS}, a diversity-oriented expert framework for long-tailed remote sensing scene recognition, where some categories appear much less frequently than others. DMRS builds multiple diversity experts together with a semantic-aware mixing strategy that exploits category-level semantic information to combine their outputs. By encouraging different experts to capture complementary semantic patterns across head and tail classes, the method reduces the bias toward frequent categories and markedly improves the recognition performance of rare classes compared with single-backbone baselines. \major{On the NWPU-RESISC45~\cite{7891544} dataset, DMRS~\cite{DMRS} reports 88.3\% overall accuracy, exceeding the next best method (MDCS~\cite{MDCS} at 81.6\%) by 6.7 percentage points, with tail class performance reaching 84.2\%, substantially higher than traditional methods that typically achieve below 65\%, highlighting the effectiveness of diversity experts for long-tailed recognition.}

\begin{figure}[t]
    \centering
    \includegraphics[width=0.8\linewidth]{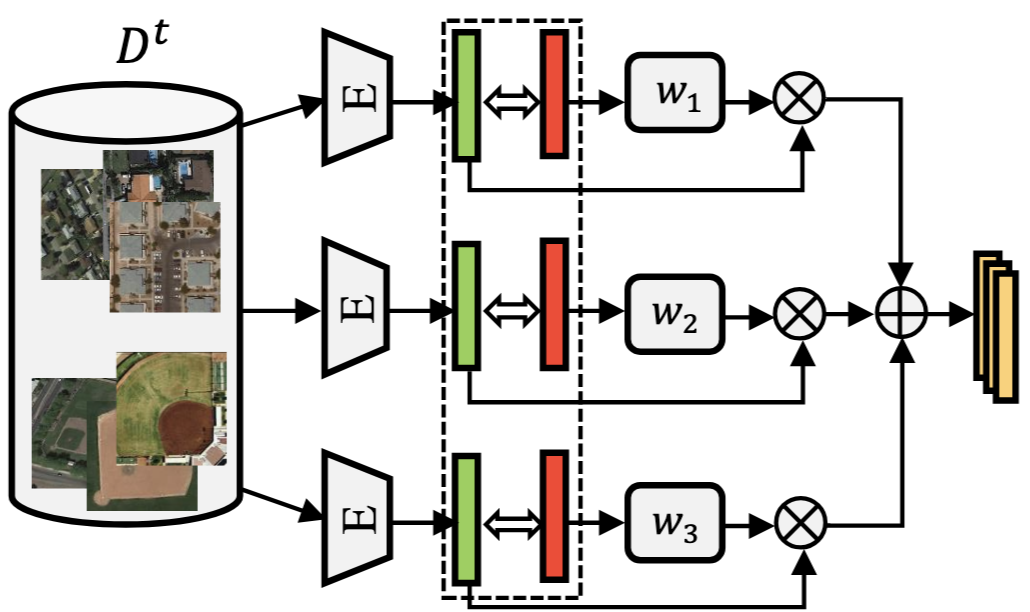}
    \caption{\major{MoE in adaptive mixture-of-experts distillation (AMoED)~\cite{fu2025adaptive} for cross-satellite generalizable incremental scene classification.}}
    \label{fig:amoed}
\end{figure}

To address domain shifts and multi-source data variability, MoE has also been applied in cross-domain adaptation scenarios. Fu et al.\ propose AMoED~\cite{fu2025adaptive}, an adaptive MoE distillation framework for cross-satellite generalizable incremental scene classification, where multiple domain-specific experts trained on different source domains provide coordinated guidance through a high-level semantic learning pipeline. The expert predictions are adaptively integrated based on domain-agnostic confidence measures to form universal class concepts, enabling stable knowledge acquisition without direct exposure to raw data streams. As shown in Fig.~\ref{fig:amoed}, the framework trains a set of domain-specific experts $E = \{E_m\}_{m=1}^M$ independently on data from $M$ distinct source domains, each specializing in recognizing newly emerged classes within its respective domain. The cross-domain classification confidence of each expert is then evaluated in a domain-agnostic manner, and the expert knowledge is adaptively mixed based on these confidence measures. An equi-partite subset is constructed by combining exemplars of previously learned classes with equally sampled instances of new classes from each source domain. Based on this subset, generalizable knowledge is acquired under the coordinated guidance of the mixed expert predictions through knowledge distillation, while knowledge consolidation is performed via class label supervision. Throughout the training process, a shallow style-mixing operation is applied to reduce geospatial and sensor-induced deviations, effectively mitigating domain shift and catastrophic forgetting across satellites. Guo et al.~\cite{guo2025confidence} addressed radar target recognition by fusing multiple radar modalities through a confidence fusion framework with representation distribution modeling and single-modal mixture of experts (sMoE). The sMoE structure employs a router to dynamically select appropriate feedforward network experts for processing different input tokens, while the representation distribution module extends point embeddings to distributional embeddings that capture uncertainty. The confidence fusion module then weights single-modal predictions based on relative confidence levels derived from evidential learning, yielding improved target classification accuracy over single-model baselines. These approaches show that MoE can handle the spectral, modal, and distributional variability inherent in multi-source remote sensing data by assigning domains or modalities to specialized experts.

\subsubsection{Hyperspectral Image Classification}

\begin{figure}[t]
    \centering
    \includegraphics[width=0.8\linewidth]{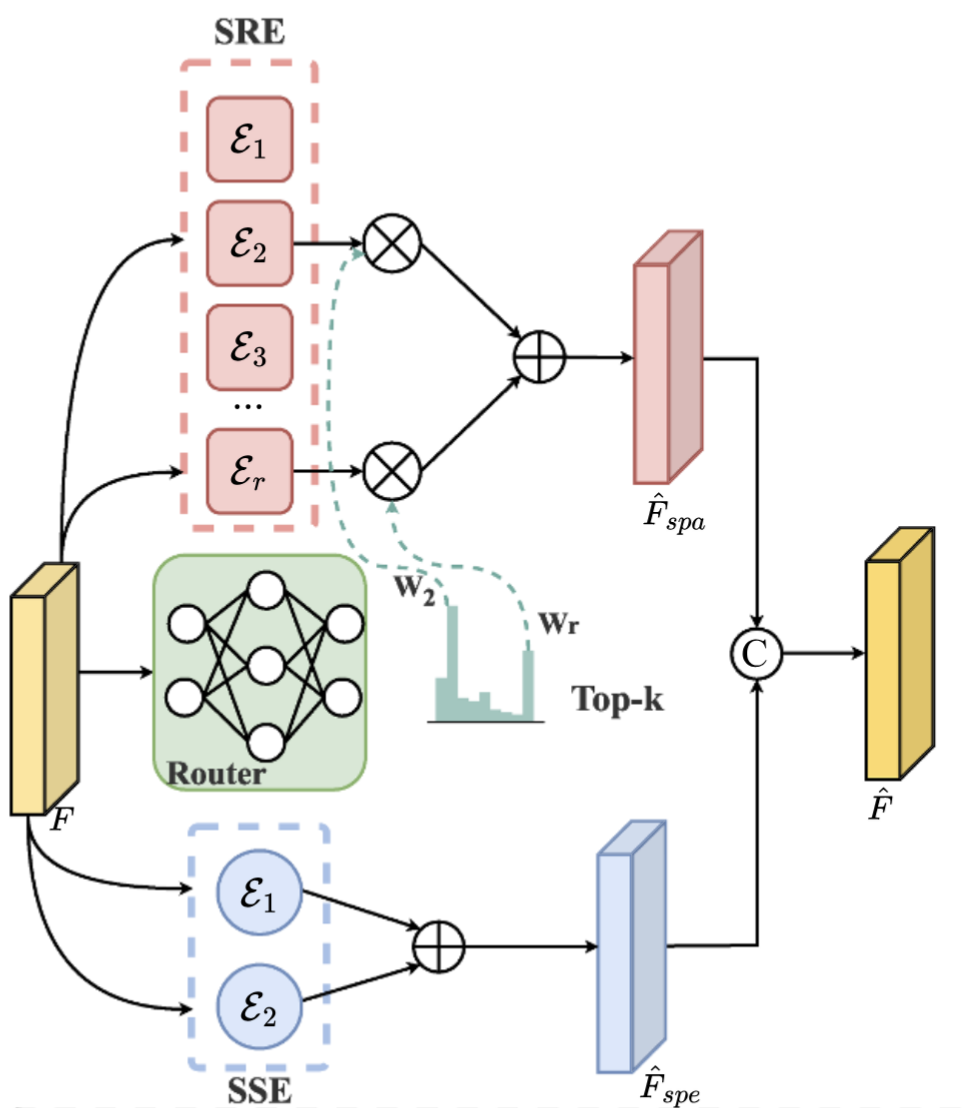}
    \caption{\major{MoE in mixture-of-spectral-spatial-experts state space model (MambaMoE)~\cite{xu2025mambamoe} for hyperspectral image classification.}}
    \label{fig:mambamoe}
\end{figure}

Hyperspectral image (HSI) classification by distinguishing materials or land-cover types from high-dimensional spectral data has greatly benefited from MoE strategies as well. Hyperspectral data is characterized by hundreds of spectral bands, and different subsets of bands often carry complementary information. Several works therefore use experts to focus on different spectral or spatial attributes. HyperTransXNet~\cite{HyperTransXNet} is a representative HSI classification model that employs a dual-branch Transformer~\cite{vaswani2017attention} with dynamic token mixers, effectively functioning as experts specializing in global vs. local spectral pattern. By learning separate expert transformers for broad spectral trends and fine-grained local variations, and then adaptively merging their outputs, this approach captures both global and local spectral dynamics to improve accuracy in HSI classification. In another study, Xu et al.\ developed MambaMoE~\cite{xu2025mambamoe}, a MoE state-space model~\cite{mamba,mamba2} for HSI classification. As illustrated in Fig.~\ref{fig:mambamoe}, MambaMoE employs a mixture of Mamba expert block (MoMEB) that splits input features $F$ into spatial and spectral views. The spatial routed expert module contains multiple Mamba experts configured with distinct spatial scanning directions, where a router network dynamically selects the top-$k$ experts and assigns routing weights to adaptively capture directional spatial context, producing spatially refined features $\hat{F}_{\mathrm{spa}}$. The spectral shared expert module employs two spectral-directional Mamba branches (forward and backward) to extract shared spectral representations, generating $\hat{F}_{\mathrm{spe}}$. The spatial and spectral features are then concatenated and integrated through a $1\times1$ convolution to form the final representation $\hat{F}$. This dual-branch design enables adaptive extraction of spectral-spatial joint features tailored to diverse land-cover characteristics. Both HyperTransXNet~\cite{HyperTransXNet} and MambaMoE~\cite{xu2025mambamoe} showed that dividing the spectral feature space among specialized experts can significantly boost classification performance on HSI benchmarks compared to conventional single-expert models. \major{On the Pavia University dataset, MambaMoE~\cite{xu2025mambamoe} yields a 3.67\% improvement in baseline overall accuracy (OA), achieving 95.20\% OA, demonstrating the effectiveness of this spectral-spatial expert decomposition approach.} MixtureRS by Liu et al.~\cite{liu2025mixturers} integrates LiDAR and hyperspectral imagery for land-cover classification using a multimodal MoE design. In MixtureRS~\cite{liu2025mixturers}, heterogeneous convolutional networks extract spectral-spatial features from hyperspectral data and elevation attributes from LiDAR data, which are then tokenized and processed through a cross-modality transformer encoder. The encoder incorporates multi-head cross-attention for inter-modal interaction and a sparse MoE feed-forward layer that selectively activates the most relevant experts for each token via top-$k$ routing. This mixture-of-modalities approach achieved superior land classification accuracy by leveraging complementary spectral and 3D structural information from the two sensors, especially for complex landscapes. Notably, MixtureRS~\cite{liu2025mixturers} highlighted that naive fusion can struggle when data are not perfectly co-registered, which can be mitigated by expert specializations for each modality. \major{On a 15-class urban benchmark, MixtureRS~\cite{liu2025mixturers} reports an overall accuracy of 88.64\%, an average accuracy of 90.23\%, and a Cohen's Kappa of 0.8767, exceeding the best homogeneous transformer by over 12 percentage points, demonstrating the advantage of modality-specific expert specialization.} Beyond these architectures, Gao et al.\ proposed a Missing-aware Mixture-of-Loras (MaMOL) framework~\cite{MaMOL}, which treats multimodal, hyperspectral-based land-cover classification with incomplete modalities as an expert-selection problem rather than explicit modality reconstruction. MaMOL inserts lightweight LoRA-based~\cite{hu2022lora} dynamic and static experts into a frozen Transformer~\cite{ViT} backbone and uses a dual routing strategy (task-aware dynamic pattern experts plus shared and modality-specific static experts) to adapt to arbitrary hyperspectral, Synthetic Aperture Radar (SAR), and Light Detection and Ranging (LiDAR) missing-patterns, achieving strong robustness and accuracy under high missing rates on benchmark datasets.

\subsection{Object Detection}

Object detection in remote sensing involves identifying and locating objects of interest (such as vehicles, buildings, ships, or airplanes) in aerial or satellite images. The task is challenging because remote sensing images are often high-resolution with small objects, complex backgrounds, and varied object orientations and scales. MoE models have started to be explored as a way to enhance detection by providing adaptable feature processing pipelines that adjust to different object types or image regions.

In object detection, one common strategy is to use experts at different feature pyramid levels or for different object orientations. Chai et al.~\cite{chai2025scalable} developed a scalable MoE attention feature pyramid network (SAFPN) to detect objects in remote sensing images. Each level of the feature pyramid has a group of expert subnetworks that attend to features of a specifical resolution. A gating mechanism then fuses information across levels, ensuring that small and large objects are all well-represented. \major{SAFPN's~\cite{chai2025scalable} multi-level expert design results in mean average precision (mAP) improvements for detection and instance segmentation from 71.3\% and 62.4\% to 82.7\% and 71.1\%, respectively, on the Airbus Ship dataset, demonstrating the benefit of resolution-specific expert specialization.} The substantial mAP gains highlight the effectiveness of multi-level expert architectures for object detection tasks. Similarly, Chen et al. proposed FAMHE-Net~\cite{chen2025famhe} for detecting oriented objects (such as rotated cars or ships) using a mixture of heterogeneous experts. FAMHE-Net~\cite{chen2025famhe} includes multiple expert detectors, each specializing in a particular object orientation or aspect ratio, and a gating network that adaptively combines their outputs. By augmenting multi-scale feature maps and letting experts handle different rotation angles, this model significantly improved detection accuracy for oriented bounding boxes compared to single-expert detectors.
In object detection pipelines, MoE has also been integrated into the region proposal and classification stages. Lin et al.~\cite{lin2025multiple} introduced a multiple region proposal experts network for detecting small objects over large areas. They trained a committee of Region Proposal Networks (RPNs) as experts, where each expert focuses on proposals in a specific image region or for specific object sizes. A gating module then merges the proposals, and subsequent classification is done by an ensemble of expert classifiers as well. The \textit{multiple experts at multiple stages} approach helped detect objects over wide scenes (e.g., detecting all cars in a city-scale image) by dividing the task among specialized RPN experts and classification experts, resulting in more reliable detection in wide-area images with experts covering different sub-scenes and object scales.

\begin{figure}[t]
    \centering
    \includegraphics[width=\linewidth]{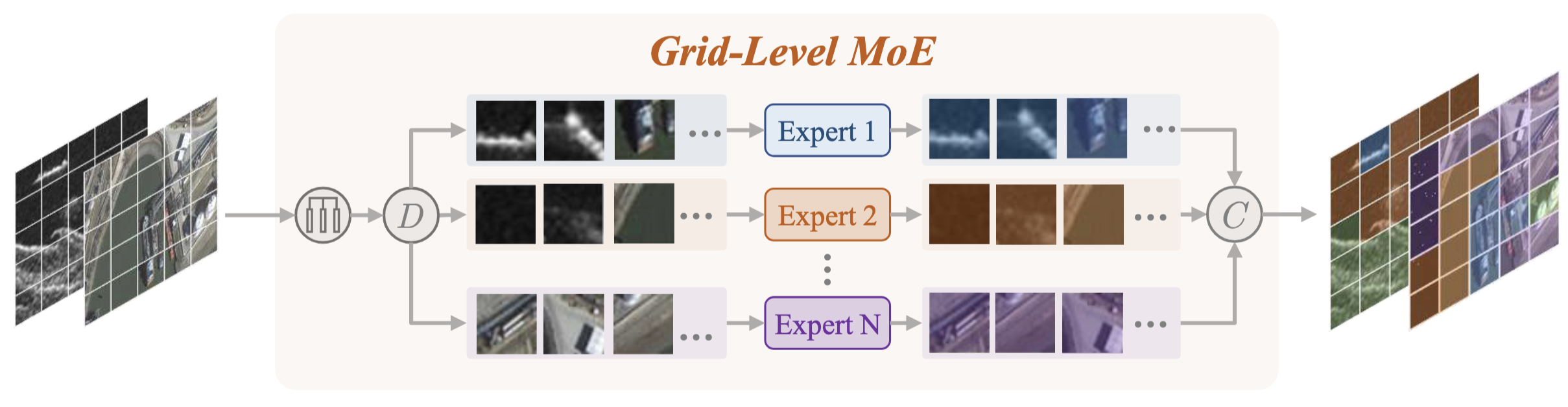}
    \caption{\major{Grid-level MoE backbone used in single model for multi-modal datasets and multi-task object detection (SM3Det)~\cite{li2024sm3det} for multi-modal remote sensing object detection.}}
    \label{fig:sm3det}
\end{figure}

For cross-domain and multimodal object detection, MoE has also been leveraged to improve performance. Li et al.~\cite{li2024sm3det} presented SM3Det, a unified model for multi-modal remote sensing object detection that uses a grid-level sparse MoE backbone. Unlike traditional approaches that route entire images to modality-specific experts, SM3Det~\cite{li2024sm3det} operates at the feature grid level, enabling experts to process local spatial features adaptively. As shown in Fig.~\ref{fig:sm3det}, multi-modal feature maps are first divided into grids and dispatched by a dispatcher $D$ to different experts based on local feature characteristics, and a collector $C$ then reassembles the processed grids so that each spatial region benefits from the expert best suited to its local patterns and appearance. This grid-level routing mechanism allows SM3Det~\cite{li2024sm3det} to capture both shared knowledge and modality-specific representations simultaneously, as experts can specialize in different local patterns across modalities while still learning common spatial structures. The model integrates a dynamic learning rate adjustment strategy to handle varying learning difficulties across different modalities and tasks. By leveraging grid-level MoE, SM3Det~\cite{li2024sm3det} achieved strong results on detecting objects across SAR, optical, and infrared modalities, highlighting MoE's utility in unified multi-modal object detection. In a related vein, Lin et al.~\cite{lin2021mednet} developed MEDNet, a multi-expert detection network that addresses the challenge of leveraging diverse distinctive information for remote sensing object detection. MEDNet~\cite{lin2021mednet} employs multiple feature pyramids (MFPs) and multiple detection experts (MDEs), where a loss distance-based k-experts clustering (LD-kEC) strategy dynamically assigns training samples to different detection experts in an unsupervised manner based on loss distances. This strategy allows each expert to specialize in detecting objects with similar characteristics (e.g., appearance, internal texture, or external context) without requiring manual expert labels. Such specialized experts significantly improved detection performance compared to a single detection pipeline approach.

Beyond traditional objects, MoE-based detection extends to forgery detection in remote sensing imagery and anomalous target localization in marine monitoring. Zhang et al.~\cite{zhang2025challenging} tackled the problem of copy-move forgery understanding in satellite images using a multimodal gated MoE model (CM-MMoE) for the remote sensing copy-move question answering task. In such forgeries, a region of an image is duplicated elsewhere to conceal information, e.g., cloning part of a satellite image to cover something. CM-MMoE employs hierarchical visual representations that include the source region, tampered region, background, and original image features, which are integrated with textual question features. A multimodal auxiliary gating network, guided by cross-attention between visual and textual features, dynamically routes inputs to multiple forgery expert networks. Each expert processes the hierarchical visual features to achieve multi-level understanding of image semantics, enabling accurate answers to diverse questions about tampering scenarios. This multi-modal MoE approach outperformed single-expert methods in identifying subtle forgeries, underlining MoE's potential in remote sensing image forensics. Qian et al.~\cite{qian2025multi} employed a multi-task, multi-expert and multi-gate (MEG) framework for underwater target detection and localization using acoustic signals. Their approach uses multiple expert networks with independent parameter spaces to specialize in different aspects of underwater acoustic signal processing. Unlike traditional multi-task learning with shared parameters, MEG employs multiple gating layers, where each task (recognition and localization) has its own gating network that dynamically learns task-specific weights to linearly combine expert outputs. This multi-gate design allows each task to obtain task-specific representations by adaptively selecting and weighting experts, while the top-$k$ gating mechanism improves efficiency by activating only the most relevant experts. This multi-task MoE strategy improved underwater object detection performance by enabling specialized feature learning for classification and localization tasks while maintaining computational efficiency.

\subsection{Spatiotemporal Modeling and Analysis}

Many remote sensing applications involve analyzing how the Earth changes over time. Temporal analysis of remote sensing data may involve modeling time series of images to detect trends, seasonal variations, or anomalies. These tasks present unique challenges: one must account for differences in imaging conditions, align multi-temporal data, and distinguish meaningful changes (like deforestation or urban growth) from irrelevant ones (like seasonal vegetation cycles or shadows moving). MoE models can assist by allocating different experts to different temporal contexts or change types, and by handling multi-modal temporal input.

\subsubsection{Change Detection}

Change detection typically uses imagery from two or more time points to identify what has changed (for example, urban expansion, deforestation, or disaster impact). MoE models can contribute here by providing experts that specialize in particular change patterns or sensor modalities, which is particularly useful when dealing with multi-temporal and multi-source data.

Liu et al.~\cite{liu2025m} proposed M$^2$CD, a unified multimodal framework for optical-SAR change detection that leverages MoE modules integrated into the backbone network. Optical (e.g., visible spectrum) and SAR imagery have very different characteristics, and changes (like urban development or flooding) might appear differently in each modality. In M$^2$CD~\cite{liu2025m}, MoE layers are inserted after each backbone block, where multiple experts adaptively handle images from different temporal phases through a gating function that selects the top-$k$ experts based on similarity scores between input features and expert embeddings. The framework also introduces an optical-to-SAR path (O2SP) that generates simulated SAR images from pre-event optical images using the fully developed speckle assumption, serving as an intermediate representation to bridge the two modalities. During training, self-distillation is applied to minimize the feature space discrepancy between the optical path (OP), O2SP, and SAR path (SP), while O2SP is omitted during inference to avoid additional computational overhead. This mixture-of-experts approach addresses the problem of heterogeneous change detection, where one modality might be affected by season or weather (optical) while the other might suffer from speckle or different imaging geometry (SAR). By creating a sparser feature space that enables distinct representation learning for optical and SAR modalities, M$^2$CD~\cite{liu2025m} achieved more accurate change detection on optical and SAR datasets than methods that simply concatenate or transform one modality into the other, \major{showing gains of 0.15\% in OA, 0.36\% in mF1, and 0.57\% in mIoU compared to TTP, validating the effectiveness of modality-adaptive expert routing for cross-modal change detection.} Another recent work by Seydi et al.~\cite{seydi2024novel} focused on detecting environmental changes due to disasters, specifically burned area mapping after wildfires, using a Siamese-based mixture of experts (SMoE) framework. They developed a deep siamese network that takes pre- and post-fire multispectral Sentinel-2 images as input through two deep feature extractor channels. The framework employs MoE layers where each expert is implemented as a convolution layer, and a gating network dynamically routes inputs to the most relevant experts based on local efficiency. The extracted features are further processed through position and channel attention modules before being fed into a dense-MoE layer for final classification. This approach yielded highly accurate burned area maps, as the mixture of experts could adaptively process different aspects of the bi-temporal data while the attention mechanisms enhanced feature representation. The MoE-based siamese model outperformed conventional change detectors by reducing false alarms and missed detections, especially in heterogeneous landscapes with complex backgrounds.

\subsubsection{Time-Series Analysis}

\begin{figure}[t]
    \centering
    \includegraphics[width=0.9\linewidth]{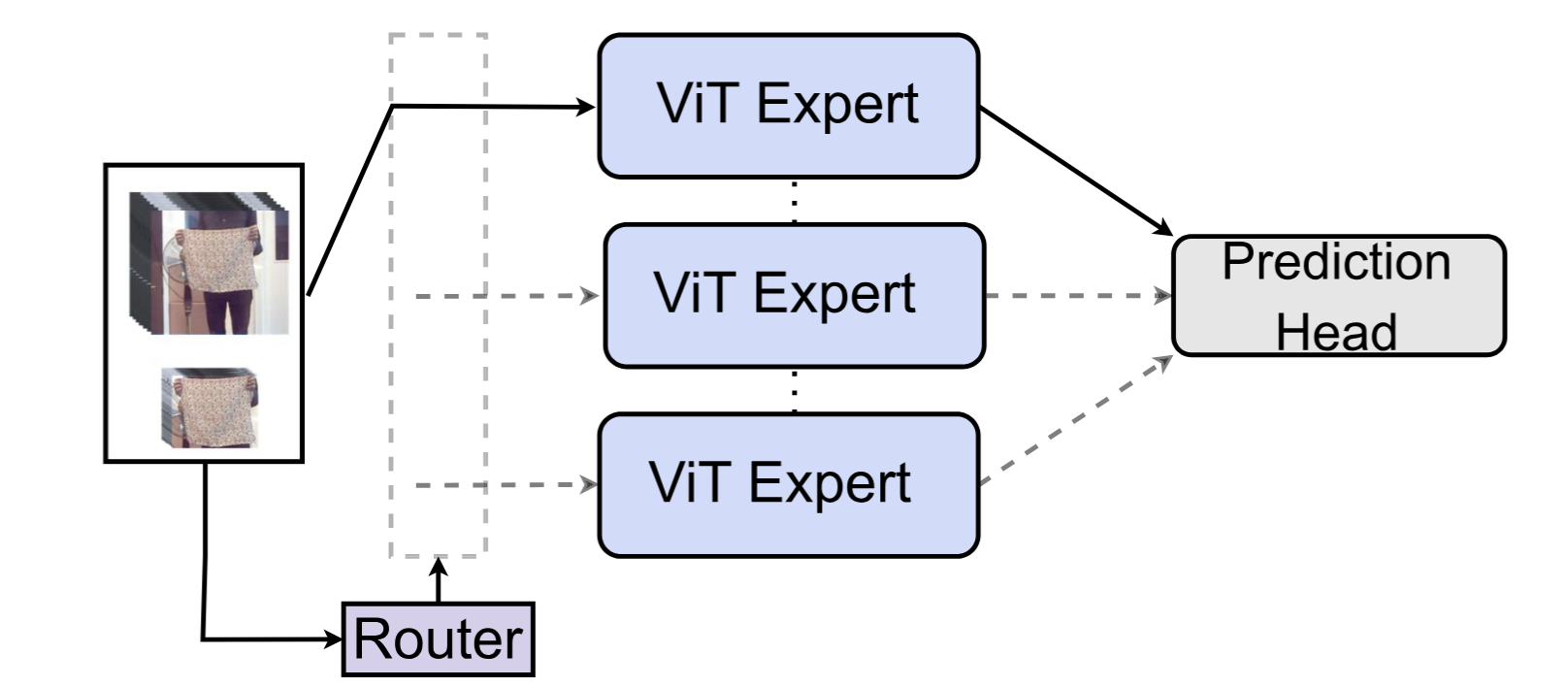}
    \caption{\major{Sparse mixture-of-experts for hyperspectral object tracking (HotMoE)~\cite{sun2025hotmoe} framework for hyperspectral object tracking.}}
    \label{fig:hotmoe}
\end{figure}

Remote sensing often involves time series analysis, such as predicting environmental variables over time or tracking moving objects across image sequences. While many MoE applications in remote sensing have focused on static imagery, a few have tackled challenges in the temporal dimension by exploiting MoE's ability to assign experts to different temporal patterns or tasks. A notable example is in climate and weather-related predictions. Jiang et al. proposed a knowledge-guided adaptive MoE model for precipitation prediction~\cite{jiang2025knowledge}. This model processes multi-source time-series data including meteorological satellite observations, reanalysis data, and ground sensor readings. The framework employs 16 independent MLP experts, with features organized into six physically meaningful categories based on domain knowledge: Momentum, Temperature, Moisture, Mass, Cloud, and Radiation. During training, a selective training strategy is applied where only two randomly selected experts are updated per epoch while others remain frozen, encouraging diversity through a custom loss function that penalizes weight similarity between active experts. A dynamic router learns to assign weights to expert outputs, computing final predictions as weighted sums of individual expert predictions. This knowledge-guided approach enables each expert to specialize in coherent subsets of climate features, leading to improved predictive accuracy and interpretability compared to baseline models. This demonstrates MoE's promise for complex time-series forecasting tasks in Earth science, where different dynamics govern different times or places. Another domain of temporal analysis is object tracking in aerial videos or sequential images. Sun et al. introduced HotMoE~\cite{sun2025hotmoe}, which explores a sparse MoE architecture for hyperspectral object tracking. As illustrated in Fig.~\ref{fig:hotmoe}, HotMoE processes hyperspectral image inputs (template and search region) through a router module that dynamically selects the most suitable ViT expert from a pool of multiple experts. The router analyzes the input characteristics and routes all tokens to a single selected expert using a hard routing strategy, where only one expert is activated per inference. Each ViT expert consists of multiple transformer encoder layers with multi-head attention mechanisms, enabling specialized processing for different tracking scenarios. The outputs from the selected expert are then fed into a prediction head to generate the final tracking results. By sparsely activating only one expert per inference instead of computing all experts, HotMoE~\cite{sun2025hotmoe} achieved efficient and robust tracking performance on hyperspectral video data, achieving 43.7 FPS with an AUC of 0.704 on the HOT2022 dataset. This approach illustrates how MoE can effectively handle high-dimensional hyperspectral data while maintaining computational efficiency through conditional expert activation.

Some recent foundation models hint at incorporating temporal expertise. Albughdadi et al.~\cite{albughdadi2025lightweight}, for example, in their MoE-MAE pretraining incorporated geo-temporal conditioning by encoding latitude, longitude, week-of-year, and hour-of-day as sinusoidal pairs to preserve cyclic structure. These metadata tokens are concatenated with patch embeddings and processed through MoE layers with NoisyTop-k routing, enabling the model to exploit spatio-temporal regularities inherent in Earth observation data. While that work is primarily about representation learning, it suggests a future direction where MoE experts could specialize not just on modalities, but on temporal segments, e.g., an expert for detecting changes in summer vs winter imagery. Time series analysis in remote sensing can also include phenological trend analysis, anomaly detection over time, and data assimilation. While not many works explicitly use MoE for these yet, the potential is clear. For example, an MoE could be devised for crop monitoring where one expert handles normal seasonal growth curves and another detects deviations (drought stress signals), with gating based on the current growth stage. In data assimilation for climate models, one could imagine experts specialized in different regimes (e.g., monsoon vs arid climate assimilation) being mixed.  Li et al. proposed STF-MoE~\cite{li2025stfmoe}, a spatio–temporal fusion MoE that couples an LSTM-Transformer~\cite{lstm,vaswani2017attention} hybrid framework with a heterogeneous mixture-of-experts mechanism. The model employs two parallel branches: a Transformer branch with multi-head self-attention to capture long-range temporal dependencies, and a bidirectional LSTM branch to extract local contextual features. The concatenated outputs from both branches are dynamically routed through an adaptive gating network to five structurally heterogeneous expert networks, with Top-2 sparse activation for computational efficiency. The model integrates multi-source remote sensing features (e.g., near-infrared reflectance vegetation index NIRv, fraction of photosynthetically active radiation absorption Fpar) and environmental variables (e.g., relative humidity, digital elevation model) for county-level wheat yield estimation across six major Chinese provinces, achieving R² = 0.827 and RMSE = 547.7 kg/ha in the most recent estimation year. This confirms that expert-based fusion can benefit quantitative agro-ecosystem prediction tasks built on multi-source remote sensing data.

In time-series and spatio-temporal remote sensing tasks, MoE applications show considerable promise. By employing experts that specialize in different temporal patterns or tracking subtasks, these models can address the high variability over time in Earth observation data. The precipitation prediction MoE~\cite{jiang2025knowledge} achieved improved accuracy and interpretability through knowledge-guided feature grouping that organizes climate variables into physically meaningful categories (e.g., Momentum, Temperature, Moisture, Mass, Cloud, Radiation), enabling each expert to specialize in coherent feature subsets. The hyperspectral tracker~\cite{sun2025hotmoe} enhanced robustness by dynamically routing inputs to specialized experts that handle distinct data distributions and address different tracking challenges, allowing the model to adapt seamlessly to various scenarios. These results indicate that MoEs can play a valuable role in temporal remote sensing analysis where different time-dependent processes need to be modeled concurrently.

\subsection{Multi-Modal Fusion and Adaptation}

Remote sensing often involves multi-modal data fusion by combining information from different sensors (e.g., optical, radar, LiDAR, multispectral), as well as adapting models across different data domains (e.g., different satellites or geographic regions). MoE models are naturally well-suited for these challenges, as they can assign dedicated experts to each modality or domain and learn how to best integrate them. A surge of recent work has applied MoE to multi-modal and cross-domain problems in remote sensing.

\subsubsection{Multi-Modal Foundation Models}

\begin{figure}[t]
    \centering
    \includegraphics[width=\linewidth]{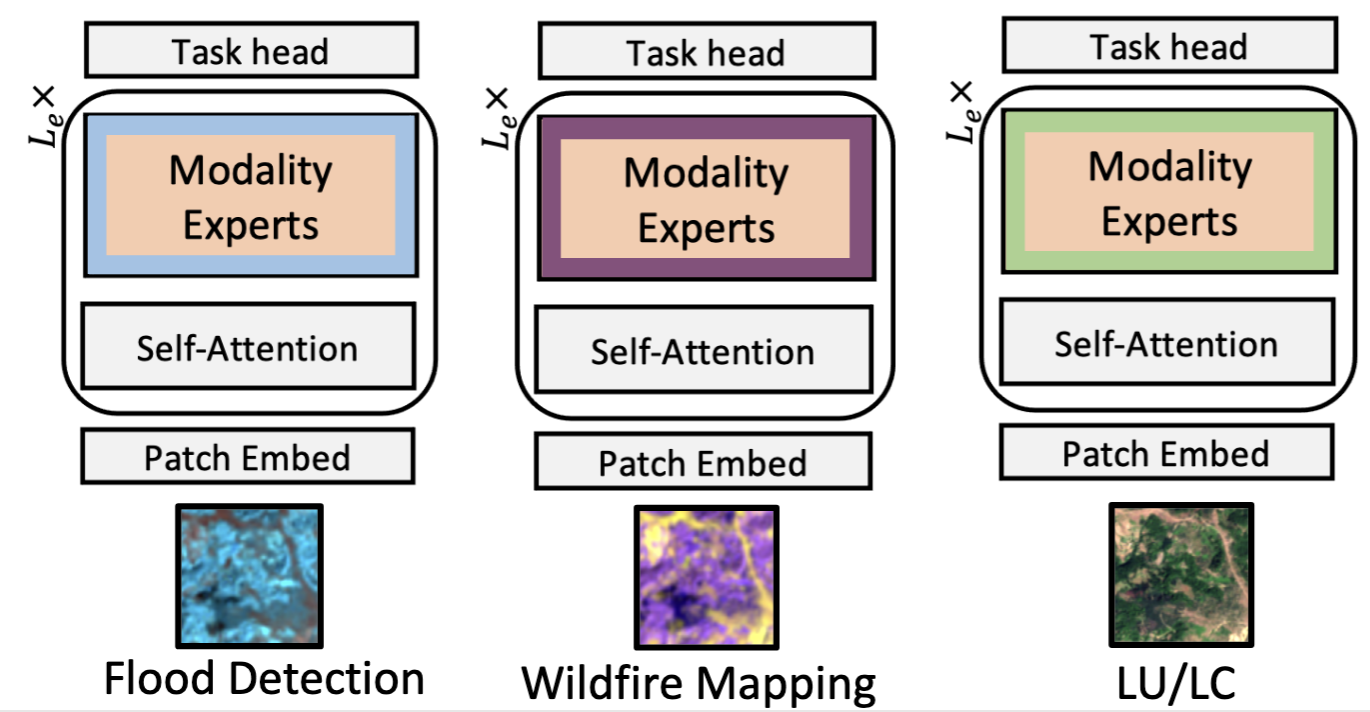}
    \caption{\major{Modality-aware pruning of experts (MAPEX)~\cite{hanna2025mapex} for multi-modal remote sensing foundation models.}}
    \label{fig:mapex}
\end{figure}

A clear example is in the construction of large remote sensing foundation models that incorporate multiple data modalities. Bi et al.~\cite{bi2025ringmoe} introduced RingMoE~\cite{bi2025ringmoe}, a MoE-based modality experts model with an enormous 14.7 billion parameters, designed as a multi-modal foundation model for universal remote sensing image interpretation. RingMoE~\cite{bi2025ringmoe} employs a hierarchical MoE architecture called RMoE (Ring MoE) that incorporates three specialized expert types: modal-specialized experts that capture fine-grained intra-modal representations for each modality (optical, SAR, multispectral, elevation), collaborative experts that model inter-modal correlations across related modalities, and a shared expert that distills common knowledge across all modalities. During pretraining on massive datasets comprising 400 million multi-modal images from nine satellites, modality-specific routing networks in RingMoE~\cite{bi2025ringmoe} learn to route tokens from each modality to its corresponding modal-specialized experts, while collaborative experts enable knowledge transfer between related modalities. This enables the model to capture a rich representation that spans visible, infrared, and radar signatures of Earth surfaces. RingMoE~\cite{bi2025ringmoe} achieved state-of-the-art results on a variety of downstream tasks (classification, segmentation, etc.) by virtue of its ability to flexibly combine modalities via expert specializations. Its success demonstrates how MoE is a powerful paradigm for scaling up multi-modal integration in remote sensing, ensuring that each data source is handled by networks optimized for its characteristics while still contributing to a unified representation. Similarly, Zhang et al. developed SkySense V2~\cite{zhang2025skysense}, a unified foundation model for multi-modal remote sensing that employs a single transformer backbone to handle multiple modalities. SkySense V2~\cite{zhang2025skysense} integrates MoE modules into the last L transformer blocks, replacing the original feed-forward network layers with M experts and a learnable gating network that performs top-k expert selection. This design enables the model to scale up capacity efficiently while maintaining computational efficiency through sparse activation. This reflects a trend of expert pruning and selection in multi-modal MoE models: Hanna et al.'s MAPEX~\cite{hanna2025mapex} specifically addresses this by performing modality-aware pruning of experts. As illustrated in Fig.~\ref{fig:mapex}, MAPEX processes input images through patch embedding, self-attention layers with modality experts (where MoE replaces the feed-forward network layers), and task-specific heads for each downstream task (e.g., flood detection, wildfire mapping, land-use/land-cover mapping). MAPEX~\cite{hanna2025mapex} employs a modality-conditioned token routing mechanism that uses learnable modality embeddings to route tokens from each modality to the same subset of experts, ensuring consistent expert-modality relationships. After pre-training, experts corresponding to unavailable modalities are pruned, and only the top-k experts for the downstream task modalities are retained, creating specialized models that are significantly smaller and easier to fine-tune. This kind of flexibility maintains many experts during training for generality, but selects a subset for a specific use, illustrating how MoE models can be both extensible and adaptable in multi-modal settings.

\subsubsection{Vision-Language Models}

Text is a special modality in remote sensing vision-language systems: unlike images or sensor signals, it typically appears as human-readable descriptions, questions, or ancillary metadata (e.g., scene tags, land-use labels, or reports), and encodes high-level semantics, reasoning cues, and task instructions rather than raw physical measurements. When combined with imagery, textual inputs enable more interactive forms of analysis such as querying, explanation, and semantic retrieval, but also introduce strong modality heterogeneity and require models that can align visual and linguistic information in a structured way. Cross-modal MoE designs are also prevalent in Visual Question Answering (VQA) and vision-language tasks for remote sensing. Liu et al.~\cite{liu2023unified} proposed a unified transformer with cross-modal MoE for remote sensing VQA. In their model, cross-modal MoE experts (CMMEs) incorporate visual and textual experts that replace conventional feed-forward networks, with shared self-attention and cross-modal attention layers to capture intricate interactions between visual and language features. The modality experts generate fused features through cross-modal interaction, and their outputs are concatenated for answer prediction. This design improved performance on remote sensing visual question answering benchmarks by effectively modeling cross-modal attention and capturing complex semantic relationships between questions and images, rather than attempting to cram all modalities into one latent space. The effectiveness of these systems demonstrates the potential of MoE for interactive remote sensing analysis. Similarly, Lin et al.~\cite{lin2025rs} and Liu \& Lian ~\cite{liu2024rsunivlm} developed large-scale vision-language models for remote sensing (RS-MoE and RS-UniVLM, respectively) that use MoE layers to handle diverse inputs like captions, scene descriptions, and images. RS-MoE~\cite{lin2025rs} employs an instruction router that dynamically generates task-specific prompts and multiple lightweight Large Language Models (LLMs) as expert models, where each expert focuses on distinct aspects of captioning: theme comprehension, object recognition, and relationship inference. \major{On the RSIEval~\cite{lin2025rs} dataset, RS-MoE-7B~\cite{lin2025rs} leads other models across all evaluation metrics, while the lightweight RS-MoE-1B~\cite{lin2025rs} variant surpasses BLIP2-13B~\cite{BLIP-2} in most of the criterion scores, demonstrating remarkable efficiency of the instruction routing mechanism.} RSUniVLM~\cite{liu2024rsunivlm} introduces a granularity-oriented MoE (G-MoE) architecture with three experts specialized for different visual granularity levels: image-level expert for holistic understanding, region-level expert for localized patterns, and pixel-level expert for fine-grained semantic information. Building on the same granularity-aware specialization method, SkyMoE~\cite{SkyMoE} further argues that a key bottleneck of remote sensing vision-language models is the persistent tension between global context reliance and local detail discrimination, and addresses it with an adaptive router that generates task- and granularity-aware routing instructions to activate specialized LLM experts. The development of these models represents a significant advancement in making remote sensing data more accessible through natural language interfaces. By leveraging LLM capabilities for natural language understanding, these models enable more intuitive interaction with remote sensing imagery. To explicitly encourage expert decoupling across local and global semantics, SkyMoE additionally introduces a context-disentangled augmentation strategy that forms contrastive local/global training pairs, and benchmarks generalization under multi-task, multi-granularity settings via MGRS-Bench. These vision-language MoEs are enabling more semantic-level interpretation of remote sensing data, moving beyond pure pixel classification to answering complex queries about images.

\subsubsection{Cross-Domain Adaptation}

Another area where MoE shines is cross-domain adaptation by ensuring models work well on data from different sources or distributions. We discussed how Land-MoE improved spectral domain generalization in classification~\cite{chen2025generalizable} and how Fu et al.~\cite{fu2025adaptive} and Lin et al.~\cite{lin2021mednet} approached domain adaptation via experts. Ngo et al.~\cite{ngo2022collaboration} also explored a related concept: using multiple experts for knowledge adaptation across multiple sources. In their approach, a shared feature extractor is combined with multiple domain-specific classifiers, where each classifier is trained on a particular source domain to specialize in that domain's characteristics. These domain-specific experts provide different views on the target domain, and collaborative learning is employed to connect these views by leveraging consistency regularization, enabling the experts to teach each other and recover missing label information. This collaboration of multiple experts was shown to yield better adaptation than single-source transfer, as each expert captured unique features of its source domain which, when combined through collaborative learning, provided a richer representation for the target. Zhang et al.~\cite{zhang2025spectralx} addressed domain generalization in the spectral domain with SpectralX, introducing parameter-efficient MoE fine-tunings for spectral shifts. SpectralX employs an attribute-oriented mixture of adapter that consists of an attribute-specific router bank and an attribute-shared adapter bank, where routing schemes dynamically allocate spatial and spectral attribute knowledge to different adapters. Rather than assigning experts to specific sensors, the model routes tokens based on their spatial and spectral attributes, enabling effective adaptation to diverse spectral imagery from different regions or seasons without retraining the whole foundation model. This demonstrates MoE's strength in capturing domain-specific nuances through attribute-oriented routing and applying them selectively.

\subsubsection{Multi-Sensor Data Fusion}

Historical uses of MoE in multi-sensor fusion provide important context. One of the earliest examples, Pasika et al. ~\cite{pasika1999neural}, applied multiple neural network methods (RBF, SVM, NDEKF-MLP, and backpropagation) to fuse diverse sensor measurements (including multispectral satellite data, temperature, and humidity) for cloud base height prediction. They demonstrated that combining information from multiple sensors could improve prediction accuracy, laying the foundation for later MoE-based fusion approaches. Aggarwal et al.~\cite{aggarwal2004multiple} proposed a multiscale data fusion method regulated by a MoE network. Their approach employed multiple multiscale Kalman filters as experts, each with different parameter vectors, to estimate topography from InSAR and ALSM data. A gating network selected the most appropriate expert for each input, effectively adapting to non-stationary terrain variations. These pioneering efforts showed that adaptive expert selection could improve fusion performance across diverse data characteristics.

For multi-sensor data fusion in classification and mapping tasks, MoE approaches have also been key. Liu et al. proposed MixtureRS, which integrates hyperspectral and LiDAR data through a cross-modality transformer enhanced with sparse MoE layers~\cite{liu2025mixturers}. The MoE layers replace conventional dense feed-forward blocks in the transformer, using Top-k routing to selectively activate the most relevant experts for each token, thereby improving model capacity without significant computational overhead. Kong et al.~\cite{kong2025joint} tackled joint classification of hyperspectral and LiDAR data using a dual MoE framework. They designed a mixture of multimodal fusion experts mechanism, where multiple fusion experts correspond to different fusion strategies, and a gating network selects and mixes these experts to achieve diverse feature fusion. The gating mechanism learns to adaptively weight different fusion strategies based on the input characteristics, enabling effective complementary learning between hyperspectral and LiDAR modalities. He et al. extended this work with an adaptive expert learning framework for HSI and MSI fusion~\cite{he2025adaptive}. They introduced a modality-guided complementary module to establish bidirectional cross-attention pathways between HSI and MSI features, followed by an attribute-aware mixture of fusion experts module. In AMoFE, the fused features are decomposed into spectral, spatial, and edge attribute subspaces, each modeled by a dedicated expert network; a soft routing mechanism then dynamically adjusts each expert's contribution based on contextual cues (e.g., regional texture complexity). This design significantly improved fusion quality by ensuring that spectral details, spatial textures, and object boundaries were each given due attention by the respective experts, rather than forcing a single model to learn all at once.

\subsection{Image Restoration and Enhancement}

Beyond classification and detection, MoE models can also be applied into image restoration and enhancement tasks in remote sensing. Such tasks include super-resolution (enhancing image spatial resolution), dehazing or denoising (improving image clarity), pansharpening (fusion of panchromatic and multispectral images), and general image quality improvement under various degradations. The rationale for using MoE is that different experts can be trained to handle different types or levels of degradations, and a gating network can then apply the appropriate experts to any given image or region.

Super-resolution (SR) is a critical task for remote sensing, aiming to increase the resolution of satellite imagery. Chen et al. proposed a heterogeneous MoE framework for remote sensing image super-resolution~\cite{chen2025heterogeneous}. Their model organizes experts into multiple expert groups, where experts within each group share identical structures while maintaining heterogeneity across groups. They introduced a multi-level feature aggregation strategy that aggregates multi-level features from the backbone network to estimate expert activation probabilities, and a dual-routing mechanism that first selects the most suitable expert group, then determines the optimal expert within the selected group. Experts across different groups utilize convolution kernels of varying sizes, enabling inter-group heterogeneous experts to provide different reconstruction scales while intra-group homogeneous experts offer multiple reconstruction patterns within the same scale. By doing so, the model was able to produce high-resolution images with both crisp details and low noise, outperforming single-expert SR networks especially on diverse terrain types. Rossi et al. took a related approach with Swin2-MoSE, a single-image super-resolution model for remote sensing~\cite{rossi2025swin2}. They integrated MoE-SM, an enhanced sparsely-gated mixture-of-experts layer, to replace the feed-forward networks inside all Transformer blocks. The MoE-SM is designed with a smart merger layer to merge the output of individual experts, and employs a per-example strategy instead of the commonly used per-token one, where all tokens of an example are processed by the same experts. \major{Swin2-MoSE's~\cite{rossi2025swin2} per-example routing strategy yields gains of up to 0.377–0.958 dB in Peak Signal-to-Noise Ratio (PSNR) and 0.0006–0.0031 in Structural Similarity Index Measure (SSIM) over any Swin-derived models on tasks of 2×, 3×, and 4× resolution upscaling, demonstrating superior performance especially for more complex tasks.} The improvements in both PSNR and SSIM metrics indicate enhanced perceptual quality in the super-resolved images. Image restoration under multiple degradations is another scenario where MoE shines. Dong et al. introduced PhyDAE, a physics-guided degradation-adaptive expert model for all-in-one remote sensing image restoration~\cite{dong2025phydae}. Instead of training separate models for denoising, deblurring, super-resolving, etc., PhyDAE~\cite{dong2025phydae} uses a set of physics-aware expert modules, each explicitly designed for a specific degradation type: dehazing, denoising, deblurring, and low-light enhancement. Each expert is rigorously designed based on corresponding physical models (e.g., atmospheric scattering model for dehazing, Retinex theory for low-light enhancement). During training, images with various simulated degradations (noise, haze, blur, low-resolution) are fed, and the gating network learns to route each degraded image to the expert best suited to restore it. Importantly, PhyDAE~\cite{dong2025phydae} incorporates physics-based constraints (e.g., sensor noise models, point spread functions) into the experts, guiding them to adhere to known degradation processes. The result is a single MoE model that can restore remote sensing images suffering from different problems. It can clean up a hazy image or sharpen a blurry one or super-resolve a coarse image, by internally selecting the appropriate expert pathway. This approach offers a flexible and efficient alternative to having separate models for each task, and the inclusion of physics knowledge in experts improved the realism and reliability of restorations.

In the realm of atmospheric correction and dehazing, MoE techniques have also been explored. Shen et al. proposed a spatial-frequency adaptive network for remote sensing image dehazing that effectively employs an MoE principle~\cite{shen2024spatial}. Their method incorporates a mixture of modulation experts in the spatial domain and a decoupled frequency learning block (DFLB) in the frequency domain. The DFLB employs a dual-branch structure to facilitate independent learning of low-frequency and high-frequency features, where low-frequency features are processed for global haze removal while high-frequency features are enhanced for detail reconstruction. A mixture of fusion experts then adaptively combines the outputs of these frequency-domain branches for the final haze-free image. By doing so, the model can address both the global effects of haze (which require low-frequency corrections) and the local effects (which benefit from high-frequency detail enhancement). This frequency-domain decoupled learning approach achieved clearer and more information-rich results on hazy satellite images than single-technique methods, highlighting how expert specialization in frequency domains is advantageous.
Another important enhancement task is pan-sharpening, where a high-resolution panchromatic image is fused with lower-resolution multispectral images to produce a high-res multispectral image. He et al. presented a frequency-adaptive pan-sharpening method using MoE~\cite{he2024frequency}. In their approach, a frequency mask predictor generates adaptive frequency masks that separate the image into high-frequency and low-frequency parts. The frequency experts module employs two MoE components: low-frequency MoE and high-frequency MoE, which exclusively process low-frequency and high-frequency information, respectively. An experts mixture module then dynamically fuses the high-frequency and low-frequency features, as well as PAN and MS features, using multiple fusion experts with adaptive gating. A gating network, guided by frequency analysis of the input, decides how to weight the fusion experts. This ensures that texture-rich areas where high-frequency detail is crucial get more contribution from the high-frequency expert, whereas homogeneous areas rely more on the low-frequency expert. This MoE-based pan-sharpening yielded images with both sharp details and accurate spectral colors, reducing typical artifacts like spectral distortion or ringing, and surpassed traditional pan-sharpening algorithms in evaluations.

In image restoration and enhancement, MoE approaches enable adaptive processing that is well-suited to the varied and complex degradations encountered in remote sensing imagery. By having experts that each excel at certain conditions, e.g., a particular noise level, frequency band, or degradation type, and gating between them, a single MoE model becomes highly versatile. This versatility is crucial for remote sensing, where images can be affected by different atmosphere, sensor, and resolution issues. The success of MoE in super-resolution, dehazing, pan-sharpening, and multi-problem restoration shows that expert specialization can yield higher fidelity outputs than one-fits-all networks, providing users with clearer, more detailed imagery.

\subsection{Other Specialized Applications}

In addition to the major categories above, MoE models show potential in several other specialized remote sensing applications. These often involve physical modeling, geoscientific data analysis, and novel tasks that benefit from expert decomposition.

One such area is geophysical parameter retrieval using remote sensing data to estimate environmental or geophysical variables via inverse modeling. Loyola et al. provided an early demonstration of MoE in this context, applying neural network MoE to the processing of satellite data for atmospheric parameter retrieval~\cite{loyola2006applications}. In their work, the complex inversion problem of predicting parameters like total column ozone from satellite radiances was broken into sub-problems handled by different networks. The input space was divided into three independent regions: aerosol types (maritime and rural), total ozone levels across different latitude bands (low, mid, and high), and satellite viewing angles (normal-view and polar-view). A total of 12 neural networks were combined via a gating network that weighted each expert's contribution based on its distance to the center of the overlapping region. This modular approach yielded both accurate and fast retrievals, as each expert could be simpler and more tuned to a subset of the problem, compared to a monolithic inversion model. It essentially proved that MoE can seamlessly merge data-driven models with the divide-and-conquer strategy often employed in physical sciences.

Seismic inversion, another form of geophysical remote sensing using seismic data to infer subsurface properties, has also adopted MoE. Li et al.\ proposed a pertinent multi-gate MoE for prestack seismic inversion, aiming to estimate multiple subsurface parameters from seismic signals~\cite{li2022pertinent}. In their model, the expert network consists of one shared expert and three special experts, each corresponding to a specific parameter (P-wave velocity, S-wave velocity, and density). Each task has its own gating network that assigns weights to the shared expert and its corresponding special expert. The shared expert receives seismic data and all three-parameter initial models, while each special expert receives seismic data and only its corresponding initial model. This design ensures that each task can learn task-specific knowledge from its special expert while benefiting from shared information through the shared expert. This approach improves inversion accuracy for each parameter and makes the overall model more robust to variations in seismic data, compared with a single network that attempts to predict all parameters simultaneously. It illustrates how MoE can effectively handle multi-output regression problems in which different outputs follow different patterns in the data.

\begin{figure}[t]
    \centering
    \includegraphics[width=\linewidth]{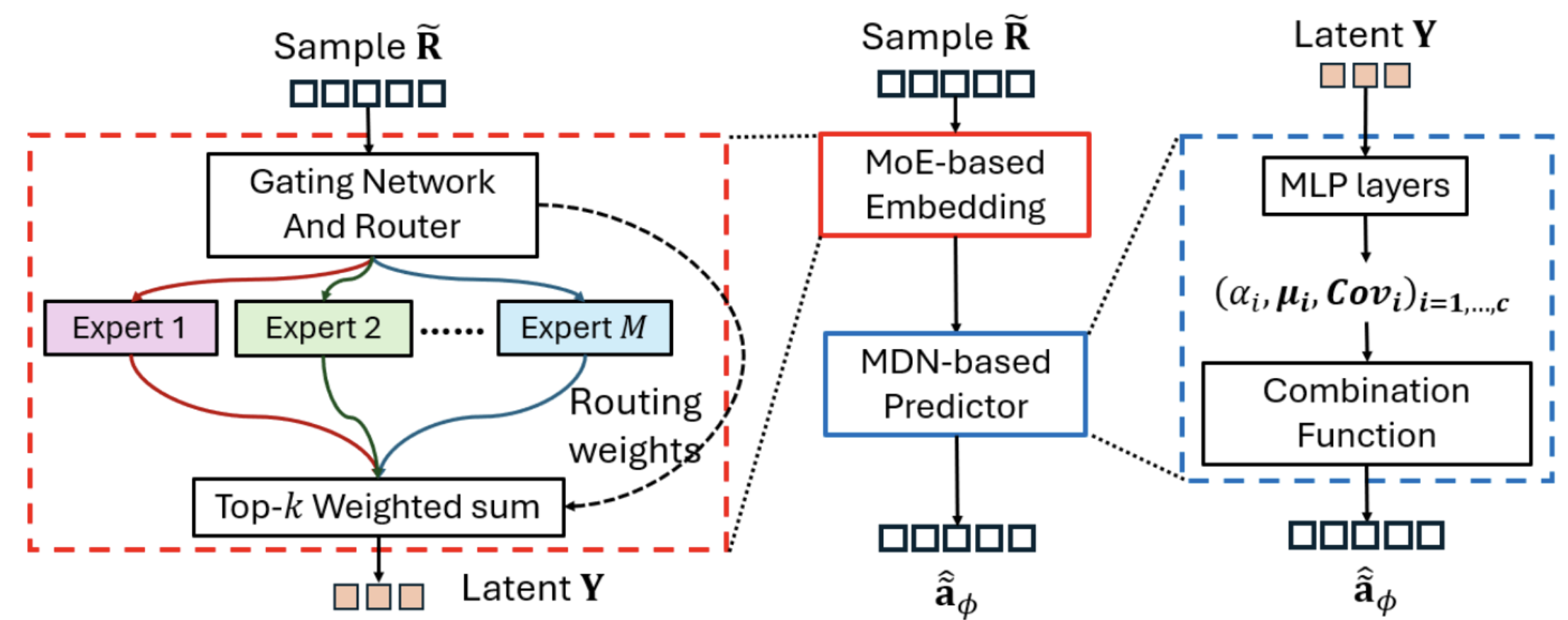}
    \caption{\major{Phytoplankton absorption mixture-of-experts (PhA-MoE)~\cite{wang2025pha} architecture for hyperspectral retrieval of phytoplankton absorption coefficients.}}
    \label{fig:phamoe}
\end{figure}

In the domain of oceanography, Wang et al. introduced PhA-MoE, an MoE model to enhance the retrieval of phytoplankton absorption coefficients from hyperspectral imagery~\cite{wang2025pha}. Retrieving ocean biochemical properties from spectral data is challenging due to variable water conditions, data scarcity, and heterogeneity. PhA-MoE addresses this by using a noisy top-$k$ gating network that dynamically selects the most relevant experts for each input, where each expert network handles a specific subset of the data distributions. As depicted in Fig.~\ref{fig:phamoe}, each pre-processed reflectance sample $\tilde{\mathbf{R}}$ is first embedded by an MoE-based embedding module into a latent vector $\mathbf{Y}$ via top-$k$ routing over multiple expert networks. The gating network determines the relevance of each expert based on the input reflectance, and the top-$k$ selected experts form a weighted sum to produce the latent embedding. The latent vector $\mathbf{Y}$ is then fed into an MDN-based predictor, which uses MLP layers to generate mixture parameters $(\alpha_i,\boldsymbol{\mu}_i,\mathbf{Cov}_i)$ that model the conditional probability distribution of absorption coefficients. A combination function then produces the final prediction $\hat{\mathbf{a}}_{\phi}$ by selecting the mean of the Gaussian component with the largest mixing coefficient. \major{PhA-MoE's~\cite{wang2025pha} MoE-based embedding approach leads to superior performance compared to other state-of-the-art models, with Normalized Root Mean Square Error (NRMSE) of 1.11 and a slope of 0.98, indicating no overestimation or underestimation, and demonstrates the effectiveness of the MoE structure in handling data heterogeneity across different evaluation metrics.} The low NRMSE value confirms the model's accuracy in retrieving phytoplankton absorption coefficients. This yielded more accurate and generalizable estimates of phytoplankton absorption across diverse water types, illustrating MoE's ability to handle heterogeneous data distributions in ocean color remote sensing.

The MoE paradigm extends to a variety of specialized remote sensing applications. It provides a framework for tackling complex inverse problems by splitting them into simpler ones handled by experts, and for injecting domain knowledge into learning via dedicated experts for known conditions. It also resonates with the ensemble nature of many remote sensing analyses, offering a trainable mechanism to combine different models or data sources. The versatility observed, from atmospheric and ocean parameter retrieval to seismic inversion and beyond, indicates that wherever a remote sensing task can be divided into sub-tasks or conditioned on context, MoE could be a beneficial approach. The next section discusses the challenges that remain in using MoE for remote sensing and potential future research directions.

\section{\major{Discussion}}
\label{sec:discussion}

\major{The evidence surveyed here indicates that the performance impact of Mixture-of-Experts (MoE) in Remote Sensing (RS) varies across datasets, modality pairings, and task formulations. RS data are heterogeneous due to sensor physics, acquisition geometry, spatial scale variation, seasonality, and regional domain shift, which provides a plausible basis for conditional specialization. These same factors also increase the risk that routing decisions reflect correlations unrelated to the target task rather than task-relevant structure, and they complicate reproducible evaluation at scale~\cite{zhu2017deeplearning,huang2025rsfm}.}

\subsection{\major{Assessment of MoE models}}

\major{A rigorous assessment of MoE requires comparisons to simpler alternatives under matched capacity and matched compute. In practice, a substantial portion of reported gains can also be obtained by scaling a dense backbone, adding explicit multi-branch or multi-scale pathways, or using ensembles at inference time. For object-centric RS tasks, multi-expert detection and proposal mechanisms often resemble structured multi-branch designs that already encourage specialization across scales~\cite{lin2021mednet,lin2025multiple}. For multi-source settings, carefully designed fusion pipelines can remain competitive when modality heterogeneity is moderate or supervision is limited~\cite{gao2020fusion}. MoE is most appropriate when the regimes that benefit from specialization are difficult to predefine, including wide geographic coverage, mixed sensors, and multi-granularity vision-language supervision, as reflected in recent RS foundation models and vision-language models that introduce modality-oriented or granularity-oriented experts~\cite{SkyMoE,bi2025ringmoe,zhang2025skysense,lin2025rs,liu2024rsunivlm}.}

\major{Sparse activation is often motivated as a way to increase parameter capacity without proportionally increasing floating-point operations~\cite{shazeer2017sparsemoe,lepikhin2021gshard,SwitchTransformers}. For RS workloads, the practical trade-offs are shaped by large spatial tensors, multi-temporal stacks, and multi-modal inputs, which amplify memory pressure and inter-device communication costs. Routing introduces dispatch and combination overhead, and its system-level impact depends on the distributed strategy and implementation. Efficient MoE training and inference therefore relies on optimized kernels and communication-aware designs~\cite{rajbhandari2022deepspeedmoe,hwang2023tutel,gale2023megablocks,hwang2024pregatedmoe}. For this reason, accuracy comparisons are more interpretable when accompanied by throughput, peak memory, and wall-clock training cost under consistent hardware and parallelism settings.}

\paragraph{\major{Assessment of effect sizes in MoE.}}
\major{Effect sizes in MoE are tightly coupled to expert count and sparsity settings. Increasing the number of experts expands representational capacity, but it also increases the risk of under-trained experts, routing collapse, and severe imbalance, especially when training data are long-tailed across regions, land-cover types, and acquisition conditions~\cite{shazeer2017sparsemoe,SwitchTransformers,riquelme2021vmoe,zoph2022stmoe}. In RS, this trade-off can be unfavorable when additional experts fragment supervision without sufficient per-expert diversity. This is consistent with RS studies that focus on usage control, regularization, or pruning rather than increasing expert counts without constraint~\cite{hanna2025mapex,chen2025heterogeneous,kong2025joint}.}

\major{To make expert-count effects interpretable, reporting should include scaling curves over expert count and top-$k$ routing, together with usage summaries such as load balance, assignment entropy, and per-expert patch counts. Router design and balancing strategy often dominate these curves. Expert-choice routing, assignment-based balancing, auxiliary-loss-free balancing, and router rebalancing explicitly target utilization issues that otherwise worsen as expert count grows. These considerations are particularly relevant for RS foundation models trained on heterogeneous global corpora, where naive routing can correlate with factors such as sensor identity, orbit, or latitude rather than task-relevant semantics~\cite{SkyMoE,bi2025ringmoe}.}

\paragraph{\major{Statistical significance and robust evaluation.}}
\major{Small reported gains are difficult to interpret without statistical evidence. MoE models can exhibit higher run-to-run variance than comparable dense models because routing stochasticity and uneven expert training introduce additional sources of instability. Robust claims therefore benefit from repeated runs with confidence intervals and, when comparing across multiple datasets or tasks, appropriate multiple-comparison procedures~\cite{demsar2006stat,dror2018hitchhiker}.} \major{Evaluation design is equally important. Random splits in geospatial data can be overly optimistic due to spatial autocorrelation, obscuring whether improvements transfer to new regions and acquisition regimes. Spatially or environmentally separated validation is often more aligned with deployment settings and reduces leakage between training and evaluation folds~\cite{roberts2017cv,valavi2019blockcv}. For time series products, temporally separated evaluation can help distinguish robust phenological learning from short-term correlations~\cite{dou2021tsclassification}. Taken together, significance analysis is most informative when paired with split strategies that match the intended transfer setting, including geographic transfer, seasonal transfer, and cross-sensor transfer.}

\paragraph{\major{Interpretability and diagnosing expert specialization.}}
\major{MoE provides a direct handle for model inspection via routing decisions. Expert assignment maps over space, time, or modality, combined with usage statistics, can reveal whether specialization aligns with meaningful regimes~\cite{pavlitskaya2020moesemseg,chen2022understanding}. However, routing-based interpretability is not guaranteed. Routers may respond to correlations that are not related to the target task, and routing patterns can change across retraining, which complicates monitoring and maintenance~\cite{zoph2022stmoe,wang2024auxiliarylossfree}. Complementary explanation tools can help verify that expert selection is supported by semantically plausible evidence, rather than artifacts, including gradient-based localization and attribution methods~\cite{selvaraju2017gradcam,sundararajan2017axiomatic,ribeiro2016whytrust}. For operational RS systems, routing diagnostics are also more actionable when combined with calibrated uncertainty estimates and structured failure auditing~\cite{guo2017calibration,ovadia2019trust}.}

\subsection{\major{Challenges of MoE in remote sensing}}

\major{The same heterogeneity that motivates MoE in RS also makes training and deployment demanding, because it affects supervision quality, routing stability, and the operational reliability of the resulting system.}

\paragraph{\major{Data heterogeneity, preprocessing, and supervision constraints.}}
\major{RS data differ from natural images in radiometric calibration, atmospheric effects, viewing geometry, and sensor-specific noise, all of which introduce structured variability that can dominate learning signals~\cite{zhu2017deeplearning}. Multi-modal settings add co-registration and resolution harmonization requirements; imperfect alignment can cause routers to specialize to misregistration patterns rather than geophysical content~\cite{gao2020fusion,he2025adaptive}. Supervision is often sparse, noisy, or updated infrequently, and domain shift across regions can be substantial. Under these conditions, specialization is beneficial only when each expert receives sufficiently consistent supervision within its regime; otherwise experts can diverge into under-constrained solutions, and performance can become sensitive to sampling choices.}

\paragraph{\major{Failure cases under domain, scale, and modality shift.}}
\major{RS systems are frequently evaluated under distribution shift, including new geography, different seasonal dynamics, and atypical atmospheric conditions. In MoE models, such shifts can lead to routing failures when out-of-distribution samples are assigned to experts that were not trained for the relevant regimes, leading to confident but incorrect predictions. This risk increases in global and multi-sensor pipelines, where missing or degraded modalities, including cloud-obscured optical imagery and gaps in Synthetic Aperture Radar (SAR) acquisitions, can cause routing to rely on spurious signals~\cite{bi2025ringmoe,MaMOL}. These observations motivate explicit shift evaluation and conservative decision rules in operational workflows, particularly when downstream actions are sensitive to false positives and false negatives.}

\paragraph{\major{Training stability and system-level constraints.}}
\major{MoE introduces stability issues that are amplified at RS scales, including expert imbalance and representational collapse, and these issues can intensify as expert count increases or as the training corpus becomes more heterogeneous. Large imagery and spatiotemporal stacks further increase memory and communication demands, making efficient dispatch and optimized kernels important for feasibility~\cite{rajbhandari2022deepspeedmoe,hwang2023tutel,gale2023megablocks}. Advances in balancing and router optimization show that utilization can be improved without extensive tuning of auxiliary losses, but systematic validation in RS-specific pipelines remains limited~\cite{lewis2021base,wang2024auxiliarylossfree,thaman2025rebalancing}.}

\paragraph{\major{Operational reliability, monitoring, and maintenance.}}
\major{Deployment introduces requirements beyond average benchmark performance. Sparse routing can yield non-uniform behavior across regions, which increases the need for geographically stratified validation, drift detection, and periodic recalibration~\cite{guo2017calibration,ovadia2019trust}. Accountability and auditing also matter in RS applications, where users may need to diagnose why outputs differ across regions or sensors, and routing summaries alone may be insufficient without additional evidence. Maintenance is also more complex than in dense models because routing patterns can change under modest data or code updates, altering expert specialization and downstream product consistency. These considerations support deployment practices that couple model-centric diagnostics with geospatially stratified evaluation and explicit update protocols~\cite{roberts2017cv,valavi2019blockcv,huang2025rsfm}.}

\subsection{\major{Benchmarking protocols and reproducibility}}\label{sec:benchmark}

\major{In Computer Vision (CV) and Natural Language Processing (NLP), MoE benchmarking has largely converged on reporting practices that disentangle model capacity from realized computation, because sparse routing makes parameter count and inference cost non-equivalent. Canonical MoE studies therefore evaluate against dense baselines under matched training budgets, and report both total parameters and the activated subset per token, together with routing-related diagnostics such as expert utilization and load balance~\cite{shazeer2017sparsemoe,SwitchTransformers,lepikhin2021gshard,riquelme2021vmoe,zhou2022expertchoice}. RS MoE research has started to adopt similar protocol choices, while also introducing domain-specific benchmarks that reflect geospatial heterogeneity and multimodality. SkyMoE explicitly proposes MGRS-Bench as a geospatial vision-language benchmark intended to stress multi-granularity interpretation and cross-domain generalization under an MoE foundation-model setting~\cite{SkyMoE}. For multi-modal detection, SM3Det establishes a unified benchmark dataset to enable single-model evaluation across modalities and detection tasks, and uses a sparse MoE backbone to support joint training without collapsing modality-specific representations~\cite{li2024sm3det}. For integrity and forensics, Zhang et al.\ release a challenging dataset for RS copy-move forgery understanding and pair it with a multimodal gated MoE model to benchmark robustness under realistic manipulations~\cite{zhang2025challenging}. Beyond explicitly new benchmarks, several RS MoE works emphasize breadth-of-evaluation as a benchmarking signal: RingMoE reports universal interpretation performance across many existing RS benchmarks under a mixture-of-modality-experts design~\cite{bi2025ringmoe}, and SkySense V2 reports results on a wide collection of multimodal datasets under a unified foundation-model interface~\cite{zhang2025skysense}; similarly, RS-MoE and RSUniVLM benchmark vision-language MoE designs on established RS captioning and visual question answering datasets to measure generalization across tasks and prompts~\cite{lin2025rs,liu2024rsunivlm,liu2023unified}.}

\major{Reproducibility for MoE models depends on faithfully specifying the routing algorithm and its training-time constraints, because small changes in capacity factors, top-$k$ routing, auxiliary balancing objectives, and token batching can shift both convergence and expert specialization. In CV and NLP, reproducibility has been strengthened by releasing optimized training stacks that make sparse dispatch deterministic and scalable, including DeepSpeed-MoE, Tutel, FastMoE, and MegaBlocks~\cite{rajbhandari2022deepspeedmoe,hwang2023tutel,he2021fastmoe,gale2023megablocks}. The lessons learned from NLP and CV communities provide valuable guidance for establishing reproducibility standards in RS MoE research. RS amplifies these sensitivities through additional degrees of freedom that sit outside the network: geospatial tiling, coordinate reference systems, cross-sensor co-registration, temporal alignment, and modality-dependent normalization can all interact with routing and produce different expert usage patterns even when the model code is unchanged~\cite{SkyMoE,bi2025ringmoe,hanna2025mapex}. As a result, several RS MoE papers foreground artifact release as part of their experimental methodology. MAPEX provides an MoE foundation model with modality-conditioned routing and modality-aware pruning, and the authors release code to support faithful replication of the full pre-training and adaptation pipeline~\cite{hanna2025mapex}. Likewise, SM3Det releases code alongside its benchmark construction and sparse MoE training recipe, which is particularly important because the protocol couples multi-dataset sampling with sparse expert activation~\cite{li2024sm3det}. These practices do not eliminate all sources of variance, but they materially reduce ambiguity around the two dominant reproducibility bottlenecks for RS MoE studies: the routing configuration that determines conditional computation, and the geospatial preprocessing pipeline that determines what tokens the router actually sees~\cite{SwitchTransformers,zhou2022expertchoice,bi2025ringmoe,SkyMoE}.}

\section{Future Directions}

Given the established progress and diverse applications of Mixture-of-Experts (MoE) in remote sensing, several promising avenues and critical challenges emerge for future research. This section outlines key directions for advancing the capabilities, efficiency, and applicability of MoE models within the unique context of Earth observation.

\subsection{Unified Multi-Modal and Multi-Task MoE}

Remote sensing workflows are increasingly multi-modal (optical, Synthetic Aperture Radar (SAR), hyperspectral, Light Detection and Ranging (LiDAR), Digital Elevation Model (DEM), text/metadata) and multi-task (classification, segmentation, detection, change detection, retrieval, captioning, Visual Question Answering (VQA)), but most existing MoE models are still designed for relatively narrow settings (e.g., a single downstream task and a small number of modalities). Integrating DEM data with optical and SAR imagery can provide complementary topographic information that enhances land-cover classification and terrain analysis. For instance, MixtureRS focuses on fusing hyperspectral imagery and LiDAR for land-use classification via modality-specific experts and cross-attention fusion, while the model proposed by Liu et al.~\cite{liu2023unified} adopts cross-modal mixture experts for remote-sensing visual question answering. These works clearly show that MoE is well suited to handle heterogeneous inputs, but they do not yet deliver a single, unified \textit{all-in-one} architecture that can serve as a foundation model across modalities and tasks.

A key future direction is the design of unified multi-modal, multi-task MoE frameworks tailored to Earth observation. Such a framework could include a backbone with separate expert pools for different modalities (e.g., optical, SAR, hyperspectral, LiDAR, text) and another set of experts for different task families (e.g., dense prediction, detection, sequence modeling). A hierarchical gating mechanism would first route tokens based on modality and acquisition metadata (sensor, orbit, season, incidence angle) and then further route within task-specific experts depending on the query (classification, segmentation, captioning, VQA, etc.). Joint pre-training of such a model with mixed objectives (masked reconstruction, contrastive alignment, language modeling, detection and segmentation heads) could produce a single MoE backbone reusable for a wide range of remote sensing tasks, including those with missing modalities or weak supervision.

In addition, future multi-modal MoE models could explicitly exploit metadata-aware routing (e.g., using acquisition time, location or orbit as router inputs) to activate experts specialized for specific regions, seasons or imaging conditions. This would be particularly valuable for large-scale monitoring, where the same architecture must handle very different landscapes and sensor configurations while sharing as much capacity as possible. Lessons from large-scale MoE in Natural Language Processing (NLP) and vision, such as sparsely-gated layers and switch-style routing~\cite{SwitchTransformers}, provide a strong blueprint for building such unified remote sensing MoE backbones at scale.

\subsection{Expert Specialization for Low-Level Vision}

Most MoE applications in remote sensing have so far emphasized mid- and high-level tasks such as classification, semantic segmentation, or VQA. Low-level vision problems, such as super-resolution, denoising, dehazing, pansharpening, atmospheric correction, and multi-degradation restoration—are equally important in operational pipelines, yet only a few works exploit MoE explicitly. A recent example is the multi-level feature guided heterogeneous MoE~\cite{chen2025heterogeneous} for remote sensing image super-resolution, which introduces heterogeneous experts in the upsampling stage and a dual-routing mechanism guided by multi-level features to adapt reconstruction to different ground-object characteristics. This example indicates that MoE can be very effective in handling the diverse textures, frequencies and structures that appear within a single remote sensing scene.

Future research can push this idea much further by designing expert sets that explicitly specialize in: (i) different degradation types (sensor noise, blur, haze, downsampling kernels, compression artefacts), (ii) different spectral types (visible, NIR, SWIR, LiDAR, etc.), and (iii) different spatial structures (water, vegetation, urban, mountainous terrain). In such models, the gating network could operate at patch- or even pixel-level to assign each region to the most appropriate combination of restoration experts. Physics-guided experts that embed sensor models, radiative transfer approximations, or priors on atmospheric scattering could coexist with purely data-driven experts, with the router learning where each is most reliable (e.g., physics-based experts for clear-sky radiometry, data-driven experts for heavy haze or mixed pixels).

Another promising line is to couple low-level MoE with downstream tasks. For example, one could jointly train a restoration MoE and a segmentation or detection head, allowing some experts to specialize in task-aware enhancement (e.g., sharpening building edges or small vessels) rather than purely perceptual quality. End-to-end training would encourage experts to learn restorations that preserve or amplify discriminative cues important for land-cover mapping, object detection, or change detection, rather than optimizing generic image quality metrics alone.

\subsection{Efficient and Robust MoE Architectures}

A well-known attraction of MoE is its ability to increase model capacity without proportionally increasing computation, by activating only a small subset of experts per input. Sparsely-gated MoE layers and Switch Transformers have shown that very large conditional-capacity models can be trained efficiently with simple routing and load-balancing strategies~\cite{SwitchTransformers}. However, remote sensing deployments impose additional constraints: models may need to run on satellites, aircraft, or edge devices with tight power and memory budgets, and they must remain robust under strong domain shifts (different sensors, regions, seasons, or acquisition geometries).

This motivates future work on resource-aware and robust MoE architectures specifically designed for Earth observation. On the efficiency side, interesting directions include: token- or patch-wise routing that prunes uninformative regions (e.g., uniform ocean), expert pruning or distillation for specific missions, and dynamic compute allocation where the router chooses not only which experts to activate but also how much computation to spend per patch. On the robustness side, domain-aware routing, conditioning on sensor ID, incidence angle, or geolocation, could help activate experts specialized for particular acquisition regimes, while auxiliary losses could regularize the router to avoid collapse (e.g., overusing one expert) and to remain stable under distribution shift. Models like ~\cite{liu2025mixturers} already hint that modality-aware expert design improves robustness to imperfect coregistration and heterogeneous landscapes; similar ideas could be extended to multi-sensor change detection or cross-satellite adaptation.

Another open issue is fault tolerance and safety in operational settings. Future MoE architectures for remote sensing should include mechanisms to detect unreliable routing decisions (e.g., when the input is far from any expert's training distribution) and fall back to simpler, more conservative experts or ensembles. This could be combined with uncertainty estimation at both expert and gating levels to support risk-aware decision-making in high-stakes applications such as disaster response or maritime surveillance.

\subsection{Effectiveness of Experts and Training Strategies}

Despite the growing number of MoE-based models, we still lack a clear understanding of what remote sensing experts actually learn, how many experts are needed, and how best to train and regularize them. General MoE work has emphasized issues such as expert specialization, load balancing, routing stability and training dynamics, proposing auxiliary losses and simplified routers (e.g., top-$k$ gating, single-expert routing) to keep training stable while encouraging diversity. Remote sensing MoEs often report ablations on the number of experts or routing variants, but systematic analyses across tasks and modalities are still rare.

A valuable research direction is to explicitly probe and visualize expert specialization in Earth observation models: for example, analyzing whether experts align with land-cover types, geographic regions, seasons, sensor families, or task types. Works such as MFG-HMoE\cite{chen2025heterogeneous} and MixtureRS~\cite{liu2025mixturers} already incorporate non-trivial routing designs (dual routing, cross-attention fusion, cross-modal mixture experts) and report improvements, but they do not fully characterize how experts partition the data space. Future studies could combine MoE with probing tasks, clustering of expert activations, and metadata conditioning to obtain more interpretable "maps of expertise" over the Earth.

Training strategies themselves are another open frontier. Beyond standard load-balancing losses, remote sensing MoE could benefit from curriculum routing (gradually increasing the number of active experts or the complexity of inputs), expert dropout (forcing robustness to missing experts), or teacher–student setups where a dense foundation model teaches a sparse MoE to allocate capacity where it matters most. In multi-task settings, it may be advantageous to share some experts across tasks while dedicating others to task-specific nuances (e.g., instance-level, pixel-level reasoning), or to learn separate routers for tasks that compete for the same features. Finally, because high-quality labels are expensive in remote sensing, semi-supervised and self-supervised pretraining for MoE using large archives of unlabeled imagery and metadata remains largely unexplored and could significantly improve expert quality and sample efficiency across downstream tasks.

\section{Conclusion}

This survey has provided a comprehensive overview of the Mixture-of-Experts (MoE) paradigm and its growing applications in Remote Sensing (RS). \major{The core advantages of MoE in RS lie in its ability to handle data heterogeneity through conditional computation and expert specialization. MoE architectures can allocate different subsets of parameters to different regions of the input space, classes, modalities, or tasks, making them natural candidates for heterogeneous RS data. By activating only a small subset of experts per input, MoE models can scale up capacity efficiently while maintaining computational efficiency, which is crucial for processing diverse Earth observation data across multiple sensors, resolutions, and temporal scales.}

\major{Despite these advantages, several unresolved key issues remain. Most existing works focus on specific tasks or limited modalities, lacking unified frameworks capable of generalizing across diverse Earth observation scenarios. The field still needs efficient and robust MoE architectures designed specifically for resource-constrained environments common in RS deployments. Understanding and visualizing expert specialization remains challenging, with limited systematic analyses of what RS experts actually learn and how they partition the data space. Training strategies that ensure stability and generalization with limited labeled data are still underdeveloped, particularly for multi-modal and multi-temporal scenarios. Additionally, benchmarking protocols and reproducibility standards need to be established to enable fair comparisons and reliable reproduction of results across different studies.}

\major{Addressing these challenges requires focused research efforts on unified multi-modal and multi-task MoE foundations, extending expert specialization to low-level vision tasks, creating resource-aware deployment strategies, and deepening the interpretability and efficiency of expert routing mechanisms. Progress in these areas will be essential for fully harnessing MoE's potential in scalable, accurate, and efficient remote sensing analysis, ultimately enabling more robust and generalizable Earth observation systems.}

\par
\vspace{10pt}


\acknowledgement
This work was supported in part by the National Natural Science Foundation of China under Grant U2243222 and Grant 61731022, in part by the National Key Research and Development Program of China under Grant 2024YFF1307204, and in part by Project of Comprehensive Site Selection System under Grant KY24004.

\conflictsofinterest
The authors declare no conflicts of interest. 

\vspace{10pt}

{\large\textbf{Ethical Approval and Consent to Participate}} 

Not applicable.


\begin{fullwidth}
\bibliographystyle{apalike}
\bibliography{ref}

@INPROCEEDINGS{2016moeKussul,
  author={Kussul, Nataliia and Shelestov, Andrii and Lavreniuk, Mykola and Butko, Igor and Skakun, Sergii},
  booktitle={2016 IEEE International Geoscience and Remote Sensing Symposium (IGARSS)}, 
  title={Deep learning approach for large scale land cover mapping based on remote sensing data fusion}, 
  year={2016},
  volume={},
  number={},
  pages={198-201},
  doi={10.1109/IGARSS.2016.7729043}}

@misc{SkyMoE,
      title={{SkyMoE}: A Vision-Language Foundation Model for Enhancing Geospatial Interpretation with Mixture of Experts}, 
      author={Jiaqi Liu and Ronghao Fu and Lang Sun and Haoran Liu and Xiao Yang and Weipeng Zhang and Xu Na and Zhuoran Duan and Bo Yang},
      year={2025},
      eprint={2512.02517},
      archivePrefix={arXiv},
      primaryClass={cs.CV},
      url={https://arxiv.org/abs/2512.02517}, 
}

@INPROCEEDINGS{2012moeLandmine,
  author={Yuksel, Seniha Esen and Gader, Paul D.},
  booktitle={2012 IEEE International Geoscience and Remote Sensing Symposium}, 
  title={Mixture of HMM Experts with applications to landmine detection}, 
  year={2012},
  volume={},
  number={},
  pages={6852-6855},
  doi={10.1109/IGARSS.2012.6352589}}

@inproceedings{2025moeRenclass,
  title={A Mixture of Experts Model for Image Classification Based on High-Resolution Remote Sensing Image},
  author={Ren, Jiwei and Zai, Kaixin and Li, Haochen and Wang, Huihui and Du, Jinjin and Mu, Weichen and Qin, Fen},
  booktitle={2025 32nd International Conference on Geoinformatics},
  pages={1--7},
  year={2025},
  organization={IEEE}
}

@article{2024moeWildlife,
  title={Towards vision mixture of experts for wildlife monitoring on the edge},
  author={Mensah, Emmanuel Azuh and Lee, Anderson and Zhang, Haoran and Shan, Yitong and Heimerl, Kurtis},
  journal={arXiv preprint arXiv:2411.07834},
  year={2024}
}

@INPROCEEDINGS{xie2022stacked,
  author={Xie, Junwei and Yu, Fan and Wang, Haonan},
  booktitle={2022 3rd International Conference on Geology, Mapping and Remote Sensing (ICGMRS)}, 
  title={Stacked Mixture-of-Expert Networks for Fast Aerial Scene Classification}, 
  year={2022},
  volume={},
  number={},
  pages={121-126},
  doi={10.1109/ICGMRS55602.2022.9849384}}

@ARTICLE{lin2025multiple,
  author={Lin, Qifeng and Huang, Haibin and Zhu, Daoye and Chen, Nuo and Fu, Gang and Yu, Yuanlong},
  journal={IEEE Transactions on Geoscience and Remote Sensing}, 
  title={Multiple Region Proposal Experts Network for Wide-Scale Remote Sensing Object Detection}, 
  year={2025},
  volume={63},
  number={},
  pages={1-16},
  doi={10.1109/TGRS.2025.3536931}}

@misc{bi2025ringmoe,
      title={{RingMoE}: Mixture-of-Modality-Experts Multi-Modal Foundation Models for Universal Remote Sensing Image Interpretation}, 
      author={Hanbo Bi and Yingchao Feng and Boyuan Tong and Mengyu Wang and Haichen Yu and Yongqiang Mao and Hao Chang and Wenhui Diao and Peijin Wang and Yue Yu and Hanyang Peng and Yehong Zhang and Kun Fu and Xian Sun},
      year={2025},
      eprint={2504.03166},
      archivePrefix={arXiv},
      primaryClass={cs.CV},
      url={https://arxiv.org/abs/2504.03166}, 
}

@inproceedings{zhang2025skysense,
  title={{SkySense V2}: A unified foundation model for multi-modal remote sensing},
  author={Zhang, Yingying and Ru, Lixiang and Wu, Kang and Yu, Lei and Liang, Lei and Li, Yansheng and Chen, Jingdong},
  booktitle={Proceedings of the IEEE/CVF International Conference on Computer Vision},
  pages={9136--9146},
  year={2025}
}

@misc{hanna2025mapex,
      title={{MAPEX}: Modality-Aware Pruning of Experts for Remote Sensing Foundation Models}, 
      author={Joelle Hanna and Linus Scheibenreif and Damian Borth},
      year={2025},
      eprint={2507.07527},
      archivePrefix={arXiv},
      primaryClass={cs.CV},
      url={https://arxiv.org/abs/2507.07527}, 
}

@article{seydi2024novel,
  title={A novel deep Siamese framework for burned area mapping Leveraging mixture of experts},
  author={Seydi, Seyd Teymoor and Hasanlou, Mahdi and Chanussot, Jocelyn},
  journal={Engineering Applications of Artificial Intelligence},
  volume={133},
  pages={108280},
  year={2024},
  publisher={Elsevier}
}

@misc{albughdadi2025lightweight,
      title={Lightweight Metadata-Aware Mixture-of-Experts Masked Autoencoder for Earth Observation}, 
      author={Mohanad Albughdadi},
      year={2025},
      eprint={2509.10919},
      archivePrefix={arXiv},
      primaryClass={cs.CV},
      url={https://arxiv.org/abs/2509.10919}, 
}

@ARTICLE{liu2025m,
  author={Liu, Ziyuan and Zhang, Jiawei and Wang, Wenyu and Gu, Yuantao},
  journal={IEEE Geoscience and Remote Sensing Letters}, 
  title={{M$^2$CD}: A Unified MultiModal Framework for Optical-SAR Change Detection With Mixture of Experts and Self-Distillation}, 
  year={2025},
  volume={22},
  number={},
  pages={1-5},
  doi={10.1109/LGRS.2025.3590959}}

@misc{zhang2025challenging,
      title={Challenging Dataset and Multi-modal Gated Mixture of Experts Model for Remote Sensing Copy-Move Forgery Understanding}, 
      author={Ze Zhang and Enyuan Zhao and Yi Jiang and Jie Nie and Xinyue Liang},
      year={2025},
      eprint={2503.18104},
      archivePrefix={arXiv},
      primaryClass={cs.MM},
      url={https://arxiv.org/abs/2503.18104}, 
}

@ARTICLE{lin2025rs,
  author={Lin, Hui and Hong, Danfeng and Ge, Shuhang and Luo, Chuyao and Jiang, Kai and Jin, Hao and Wen, Congcong},
  journal={IEEE Transactions on Geoscience and Remote Sensing}, 
  title={{RS-MoE}: A Vision–Language Model With Mixture of Experts for Remote Sensing Image Captioning and Visual Question Answering}, 
  year={2025},
  volume={63},
  number={},
  pages={1-18},
  doi={10.1109/TGRS.2025.3547988}}

@misc{liu2024rsunivlm,
      title={{RSUniVLM}: A Unified Vision Language Model for Remote Sensing via Granularity-oriented Mixture of Experts}, 
      author={Xu Liu and Zhouhui Lian},
      year={2024},
      eprint={2412.05679},
      archivePrefix={arXiv},
      primaryClass={cs.CV},
      url={https://arxiv.org/abs/2412.05679}, 
}

@article{liu2023unified,
  title={Unified transformer with cross-modal mixture experts for remote-sensing visual question answering},
  author={Liu, Gang and He, Jinlong and Li, Pengfei and Zhong, Shenjun and Li, Hongyang and He, Genrong},
  journal={Remote Sensing},
  volume={15},
  number={19},
  pages={4682},
  year={2023},
  publisher={MDPI}
}

@ARTICLE{kong2025joint,
  author={Kong, Yi and Yu, Shaocai and Cheng, Yuhu and Philip Chen, C. L. and Wang, Xuesong},
  journal={IEEE Transactions on Geoscience and Remote Sensing}, 
  title={Joint Classification of Hyperspectral Images and LiDAR Data Based on Candidate Pseudo Labels Pruning and Dual Mixture of Experts}, 
  year={2025},
  volume={63},
  number={},
  pages={1-12},
  doi={10.1109/TGRS.2025.3543498}}

@ARTICLE{he2025adaptive,
  author={He, Wangquan and Cai, Yixun and Ren, Qi and Ruze, Abuduwaili and Jia, Sen},
  journal={IEEE Transactions on Geoscience and Remote Sensing}, 
  title={Adaptive Expert Learning for Hyperspectral and Multispectral Image Fusion}, 
  year={2025},
  volume={63},
  number={},
  pages={1-15},
  doi={10.1109/TGRS.2025.3620897}}

@article{liu2025mixturers,
  title={{MixtureRS}: A Mixture of Expert Network Based Remote Sensing Land Classification},
  author={Liu, Yimei and Wu, Changyuan and Guan, Minglei and Wang, Jingzhe},
  journal={Remote Sensing},
  volume={17},
  number={14},
  pages={2494},
  year={2025},
  publisher={MDPI}
}

@article{qian2025multi,
  title={Multi-Task Mixture-of-Experts Model for Underwater Target Localization and Recognition},
  author={Qian, Peng and Wang, Jingyi and Liu, Yining and Chen, Yingxuan and Wang, Pengjiu and Deng, Yanfa and Xiao, Peng and Li, Zhenglin},
  journal={Remote Sensing},
  volume={17},
  number={17},
  pages={2961},
  year={2025},
  publisher={MDPI}
}

@article{nagarajan2009multiscale,
  title={Multiscale Segmentation of Elevation Images Using a Mixture-of-Experts Framework},
  author={Nagarajan, Karthik and Slatton, K Clint},
  journal={IEEE Geoscience and Remote Sensing Letters},
  volume={6},
  number={4},
  pages={865--869},
  year={2009},
  publisher={IEEE}
}

@misc{li2024sm3det,
      title={{SM3Det}: A Unified Model for Multi-Modal Remote Sensing Object Detection}, 
      author={Yuxuan Li and Xiang Li and Yunheng Li and Yicheng Zhang and Yimian Dai and Qibin Hou and Ming-Ming Cheng and Jian Yang},
      year={2025},
      eprint={2412.20665},
      archivePrefix={arXiv},
      primaryClass={cs.CV},
      url={https://arxiv.org/abs/2412.20665}, 
}

@article{li2025u,
  title={{U-MoEMamba}: A Hybrid Expert Segmentation Model for Cabbage Heads in Complex UAV Low-Altitude Remote Sensing Scenarios},
  author={Li, Rui and Ding, Xue and Peng, Shuangyun and Cai, Fapeng},
  journal={Agriculture},
  volume={15},
  number={16},
  pages={1723},
  year={2025},
  publisher={MDPI}
}

@inproceedings{xu2025multi,
  title={Multi-Scale Mixture-of-Experts With Lora for Building Extraction from Optical Remote-Sensing Images},
  author={Xu, Huina and Xue, Bowei and Liu, Runjie and Zhang, Qionggui and Lu, Wenjing},
  booktitle={2025 32nd International Conference on Geoinformatics},
  pages={1--9},
  year={2025},
  organization={IEEE}
}

@misc{MaMOL,
      title={Rethinking Efficient Mixture-of-Experts for Remote Sensing Modality-Missing Classification}, 
      author={Qinghao Gao and Jiahui Qu and Yunsong Li and Wenqian Dong},
      year={2025},
      eprint={2511.11460},
      archivePrefix={arXiv},
      primaryClass={cs.CV},
      url={https://arxiv.org/abs/2511.11460}, 
}

@article{lstm,
  title={Long short-term memory},
  author={Hochreiter, Sepp and Schmidhuber, J{\"u}rgen},
  journal={Neural computation},
  volume={9},
  number={8},
  pages={1735--1780},
  year={1997},
  publisher={MIT press}
}

@misc{lee2025generalizableslumdetectionsatellite,
      title={Generalizable Slum Detection from Satellite Imagery with Mixture-of-Experts}, 
      author={Sumin Lee and Sungwon Park and Jeasurk Yang and Jihee Kim and Meeyoung Cha},
      year={2025},
      eprint={2511.10300},
      archivePrefix={arXiv},
      primaryClass={cs.CV},
      url={https://arxiv.org/abs/2511.10300}, 
}

@article{mehta2023separable,
title={Separable Self-attention for Mobile Vision Transformers},
author={Sachin Mehta and Mohammad Rastegari},
journal={Transactions on Machine Learning Research},
issn={2835-8856},
year={2023},
url={https://openreview.net/forum?id=tBl4yBEjKi},
note={}
}

@ARTICLE{li2025stfmoe,
AUTHOR={Li, Jian  and Kang, Junrui  and Lu, Jian  and Fu, Hongkun  and Li, Zheng  and Liu, Baoqi  and Lin, Xinglei  and Zhao, Jiawei  and Guan, Hengxu  and Liu, He  and Liu, Zhihan },
TITLE={Dynamic gating-enhanced deep learning model with multi-source remote sensing synergy for optimizing wheat yield estimation},
JOURNAL={Frontiers in Plant Science},
VOLUME={Volume 16 - 2025},
YEAR={2025},
URL={https://www.frontiersin.org/journals/plant-science/articles/10.3389/fpls.2025.1640806},
DOI={10.3389/fpls.2025.1640806},
ISSN={1664-462X},}

@inproceedings{he2024moe_semseg,
author = {Shaofeng He and Qiu Cheng and Yu Huai and Zhongke Zhu and Jie Ding},
title = {{Mixture-of-experts for semantic segmentation of remoting sensing image}},
volume = {13213},
booktitle = {International Conference on Image Processing and Artificial Intelligence (ICIPAl 2024)},
editor = {Chuan Qin and Huiyu Zhou},
organization = {International Society for Optics and Photonics},
publisher = {SPIE},
pages = {132131Z},
year = {2024},
doi = {10.1117/12.3035091},
URL = {https://doi.org/10.1117/12.3035091}
}

@inproceedings{sun2025multi,
  title={Multi-scale Feature Interaction and Adaptive Experts for Panoptic Segmentation in Remote Sensing Images},
  author={Sun, Zhenkun and Liu, Jia and Zhang, Wenhua and Liu, Fang and Yang, Jingxiang and Xiao, Liang},
  booktitle={ICASSP 2025 - IEEE International Conference on Acoustics, Speech and Signal Processing (ICASSP)},
  pages={1--5},
  year={2025},
  organization={IEEE}
}

@misc{chen2025generalizable,
      title={Generalizable Multispectral Land Cover Classification via Frequency-Aware Mixture of Low-Rank Token Experts}, 
      author={Xi Chen and Shen Yan and Juelin Zhu and Chen Chen and Yu Liu and Maojun Zhang},
      year={2025},
      eprint={2505.14088},
      archivePrefix={arXiv},
      primaryClass={cs.CV},
      url={https://arxiv.org/abs/2505.14088}, 
}

@ARTICLE{chen2024sparseformer,
  author={Chen, Yujia and Cui, Hao and Zhang, Guo and Li, Xue and Xie, Zhigang and Li, Haifeng and Li, Deren},
  journal={IEEE Transactions on Geoscience and Remote Sensing}, 
  title={{SparseFormer}: A Credible Dual-CNN Expert-Guided Transformer for Remote Sensing Image Segmentation With Sparse Point Annotation}, 
  year={2025},
  volume={63},
  number={},
  pages={1-16},
  doi={10.1109/TGRS.2024.3523537}}

@inproceedings{aggarwal2004multiple,
  title={Multiple-model multiscale data fusion regulated by a mixture-of-experts network},
  author={Aggarwal, Vikas and Nagarajan, Karthik and Slatton, K Clint},
  booktitle={IGARSS 2004. 2004 IEEE International Geoscience and Remote Sensing Symposium},
  volume={1},
  year={2004},
  organization={IEEE}
}

@inproceedings{pasika1999neural,
  title={Neural networks for sensor fusion in remote sensing},
  author={Pasika, Hugh and Haykin, Simon and Clothiaux, Eugene and Stewart, Ron},
  booktitle={IJCNN'99. International Joint Conference on Neural Networks. Proceedings (Cat. No. 99CH36339)},
  volume={4},
  pages={2772--2776},
  year={1999},
  organization={IEEE}
}

@misc{zhang2025spectralx,
      title={{SpectralX}: Parameter-efficient Domain Generalization for Spectral Remote Sensing Foundation Models}, 
      author={Yuxiang Zhang and Wei Li and Mengmeng Zhang and Jiawei Han and Ran Tao and Shunlin Liang},
      year={2025},
      eprint={2508.01731},
      archivePrefix={arXiv},
      primaryClass={cs.CV},
      url={https://arxiv.org/abs/2508.01731}, 
}

@article{ngo2022collaboration,
  title={Collaboration between multiple experts for knowledge adaptation on multiple remote sensing sources},
  author={Ngo, Ba Hung and Kim, Ju Hyun and Park, So Jeong and Cho, Sung In},
  journal={IEEE Transactions on Geoscience and Remote Sensing},
  volume={60},
  pages={1--15},
  year={2022},
  publisher={IEEE}
}

@article{lin2021mednet,
  title={{MEDNet}: Multiexpert detection network with unsupervised clustering of training samples},
  author={Lin, Qifeng and Zhao, Jianhui and Du, Bo and Fu, Gang and Yuan, Zhiyong},
  journal={IEEE Transactions on Geoscience and Remote Sensing},
  volume={60},
  pages={1--14},
  year={2021},
  publisher={IEEE}
}

@ARTICLE{guo2025confidence,
  author={Guo, Shuai and Chen, Ting and Wang, Penghui and Yan, Junkun and Liu, Hongwei},
  journal={IEEE Transactions on Aerospace and Electronic Systems}, 
  title={Confidence Fusion With Representation Distribution and Mixture of Experts for Multimodal Radar Target Recognition}, 
  year={2025},
  volume={61},
  number={5},
  pages={13251-13268},
  doi={10.1109/TAES.2025.3576777}}

@article{chen2025famhe,
  title={{FAMHE-Net}: Multi-Scale Feature Augmentation and Mixture of Heterogeneous Experts for Oriented Object Detection},
  author={Chen, Yixin and Jiang, Weilai and Wang, Yaonan},
  journal={Remote Sensing},
  volume={17},
  number={2},
  pages={205},
  year={2025},
  publisher={MDPI AG}
}

@article{wang2025pha,
  title={{PhA-MOE}: Enhancing Hyperspectral Retrievals for Phytoplankton Absorption Using Mixture-of-Experts},
  author={Wang, Weiwei and Liu, Bingqing and Gao, Song and Li, Jiang and Zhou, Yueling and Zhang, Songyang and Ding, Zhi},
  journal={Remote Sensing},
  volume={17},
  number={12},
  pages={2103},
  year={2025},
  publisher={MDPI}
}

@ARTICLE{li2022pertinent,
  author={Li, Zihang and Chen, Xiaohong and Li, Jingye and Zhang, Jian},
  journal={IEEE Transactions on Geoscience and Remote Sensing}, 
  title={Pertinent Multigate Mixture-of-Experts-Based Prestack Three-Parameter Seismic Inversion}, 
  year={2022},
  volume={60},
  number={},
  pages={1-15},
  doi={10.1109/TGRS.2022.3208226}}

@article{loyola2006applications,
  title={Applications of neural network methods to the processing of earth observation satellite data},
  author={Loyola R, Diego G},
  journal={Neural networks},
  volume={19},
  number={2},
  pages={168--177},
  year={2006},
  publisher={Elsevier}
}

@inproceedings{he2024frequency,
  title={Frequency-adaptive pan-sharpening with mixture of experts},
  author={He, Xuanhua and Yan, Keyu and Li, Rui and Xie, Chengjun and Zhang, Jie and Zhou, Man},
  booktitle={Proceedings of the AAAI Conference on Artificial Intelligence},
  volume={38},
  number={3},
  pages={2121--2129},
  year={2024}
}

@ARTICLE{shen2024spatial,
  author={Shen, Hao and Ding, Henghui and Zhang, Yulun and Cong, Xiaofeng and Zhao, Zhong-Qiu and Jiang, Xudong},
  journal={IEEE Transactions on Geoscience and Remote Sensing}, 
  title={Spatial-Frequency Adaptive Remote Sensing Image Dehazing With Mixture of Experts}, 
  year={2024},
  volume={62},
  number={},
  pages={1-14},
  doi={10.1109/TGRS.2024.3458986}}

@misc{dong2025phydae,
      title={{PhyDAE}: Physics-Guided Degradation-Adaptive Experts for All-in-One Remote Sensing Image Restoration}, 
      author={Zhe Dong and Yuzhe Sun and Haochen Jiang and Tianzhu Liu and Yanfeng Gu},
      year={2025},
      eprint={2510.08653},
      archivePrefix={arXiv},
      primaryClass={cs.CV},
      url={https://arxiv.org/abs/2510.08653}, 
}

@article{rossi2025swin2,
  title={{Swin2-MoSE}: A new single image supersolution model for remote sensing},
  author={Rossi, Leonardo and Bernuzzi, Vittorio and Fontanini, Tomaso and Bertozzi, Massimo and Prati, Andrea},
  journal={IET Image Processing},
  volume={19},
  number={1},
  pages={e13303},
  year={2025},
  publisher={Wiley Online Library}
}

@ARTICLE{chen2025heterogeneous,
  author={Chen, Bowen and Chen, Keyan and Yang, Mohan and Zou, Zhengxia and Shi, Zhenwei},
  journal={IEEE Geoscience and Remote Sensing Letters}, 
  title={Heterogeneous Mixture of Experts for Remote Sensing Image Super-Resolution}, 
  year={2025},
  volume={22},
  number={},
  pages={1-5},
  doi={10.1109/LGRS.2025.3557928}}

@ARTICLE{sun2025hotmoe,
  author={Sun, Wenfang and Tan, Yuedong and Li, Jingyuan and Hou, Shuwei and Li, Xiaobo and Shao, Yingzhao and Wang, Zhe and Song, Beibei},
  journal={IEEE Transactions on Multimedia}, 
  title={{HotMoE}: Exploring Sparse Mixture-of-Experts for Hyperspectral Object Tracking}, 
  year={2025},
  volume={27},
  number={},
  pages={4072-4083},
  doi={10.1109/TMM.2025.3535339}}

@misc{jiang2025knowledge,
      title={Knowledge-Guided Adaptive Mixture of Experts for Precipitation Prediction}, 
      author={Chen Jiang and Kofi Osei and Sai Deepthi Yeddula and Dongji Feng and Wei-Shinn Ku},
      year={2025},
      eprint={2509.11459},
      archivePrefix={arXiv},
      primaryClass={cs.AI},
      url={https://arxiv.org/abs/2509.11459}, 
}

@article{chai2025scalable,
  title={Scalable Mixture-of-Experts Attention Feature Pyramid Network for Detection and Segmentation},
  author={Chai, Bosong and Zhou, Qifan and Nie, Xuan and Qiao, Qian and Wu, Wangyu and Shi, Yongji and Li, Xuedong},
  year={2025},
  publisher={Preprints}
}

@article{xu2025mambamoe,
  title={{MambaMoE}: Mixture-of-Spectral-Spatial-Experts State Space Model for Hyperspectral Image Classification},
  author={Xu, Yichu and Wang, Di and Jiao, Hongzan and Zhang, Lefei and Zhang, Liangpei},
  journal={arXiv preprint arXiv:2504.20509},
  year={2025}
}

@article{HyperTransXNet,
  title={{HyperTransXNet}: learning both global and local dynamics with a dual dynamic token mixer for hyperspectral image classification},
  author={Dai, Xin and Li, Zexi and Li, Lin and Xue, Shuihua and Huang, Xiaohui and Yang, Xiaofei},
  journal={Remote Sensing},
  volume={17},
  number={14},
  pages={2361},
  year={2025},
  publisher={MDPI}
}

@article{DMRS,
  title={{DMRS}: Long-tailed remote sensing recognition via semantic-aware mixing and diversity experts},
  author={Wang, Yifan and Zhang, Fan and Zhao, Qihao and Hu, Wei and Ma, Fei},
  journal={International Journal of Applied Earth Observation and Geoinformation},
  volume={141},
  pages={104623},
  year={2025},
  publisher={Elsevier}
}

@ARTICLE{fu2025adaptive,
  author={Fu, Yimin and Yang, Runqing and Liu, Zhunga and Ng, Michael K.},
  journal={IEEE Transactions on Circuits and Systems for Video Technology}, 
  title={Adaptive Mixture-of-Experts Distillation for Cross-Satellite Generalizable Incremental Remote Sensing Scene Classification}, 
  year={2025},
  volume={},
  number={},
  pages={1-1},
  doi={10.1109/TCSVT.2025.3598274}}

@inproceedings{ViT,
  title     = {An Image is Worth 16x16 Words: Transformers for Image Recognition at Scale},
  author    = {Dosovitskiy, Alexey and Beyer, Lucas and Kolesnikov, Alexander and others},
  booktitle = {International Conference on Learning Representations (ICLR)},
  year      = {2021},
  url       = {https://arxiv.org/abs/2010.11929}
}

@inproceedings{alexnet,
  author    = {Krizhevsky, Alex and Sutskever, Ilya and Hinton, Geoffrey E.},
  title     = {ImageNet Classification with Deep Convolutional Neural Networks},
  booktitle = {Advances in Neural Information Processing Systems},
  year      = {2012},
  volume    = {25}
}

@inproceedings{vaswani2017attention,
author = {Vaswani, Ashish and Shazeer, Noam and Parmar, Niki and Uszkoreit, Jakob and Jones, Llion and Gomez, Aidan N. and Kaiser, \L{}ukasz and Polosukhin, Illia},
title = {Attention is all you need},
year = {2017},
isbn = {9781510860964},
publisher = {Curran Associates Inc.},
address = {Red Hook, NY, USA},
booktitle = {Proceedings of the 31st International Conference on Neural Information Processing Systems},
pages = {6000–6010},
numpages = {11},
location = {Long Beach, California, USA},
series = {NIPS'17}
}

@ARTICLE{jacobs1991adaptive,
  author={Jacobs, Robert A. and Jordan, Michael I. and Nowlan, Steven J. and Hinton, Geoffrey E.},
  journal={Neural Computation}, 
  title={Adaptive Mixtures of Local Experts}, 
  year={1991},
  volume={3},
  number={1},
  pages={79-87},
  doi={10.1162/neco.1991.3.1.79}}

@ARTICLE{jordan1994hme,
  author={Jordan, Michael I. and Jacobs, Robert A.},
  journal={Neural Computation}, 
  title={Hierarchical Mixtures of Experts and the EM Algorithm}, 
  year={1994},
  volume={6},
  number={2},
  pages={181-214},
  doi={10.1162/neco.1994.6.2.181}}

@InProceedings{jiang1999hmeglm,
  title = 	 {Hierarchical Mixtures-of-Experts for Generalized Linear Models: Some Results on Denseness and Consistency},
  author =       {Jiang, Wenxin and Tanner, Martin A.},
  booktitle = 	 {Proceedings of the Seventh International Workshop on Artificial Intelligence and Statistics},
  year = 	 {1999},
  editor = 	 {Heckerman, David and Whittaker, Joe},
  volume = 	 {R2},
  series = 	 {Proceedings of Machine Learning Research},
  month = 	 {03--06 Jan},
  publisher =    {PMLR},
  url = 	 {https://proceedings.mlr.press/r2/jiang99a.html},
  note =         {Reissued by PMLR on 20 August 2020.}
}

@ARTICLE{yuksel2012twenty,
  author={Yuksel, Seniha Esen and Wilson, Joseph N. and Gader, Paul D.},
  journal={IEEE Transactions on Neural Networks and Learning Systems}, 
  title={Twenty Years of Mixture of Experts}, 
  year={2012},
  volume={23},
  number={8},
  pages={1177-1193},
  doi={10.1109/TNNLS.2012.2200299}}

@article{masoudnia2014survey,
  author    = {Saeid Masoudnia and Reza Ebrahimpour},
  title     = {Mixture of experts: a literature survey},
  journal   = {Artificial Intelligence Review},
  year      = {2014},
  volume    = {42},
  number    = {2},
  pages     = {275--293},
  month     = aug,
  doi       = {10.1007/s10462-012-9338-y},
  url       = {https://doi.org/10.1007/s10462-012-9338-y},
  issn      = {1573-7462},
  publisher = {Springer},
}

@article{nguyen2018overview,
author = {Nguyen, Hien D. and Chamroukhi, Faicel},
title = {Practical and theoretical aspects of mixture-of-experts modeling: An overview},
journal = {WIREs Data Mining and Knowledge Discovery},
volume = {8},
number = {4},
pages = {e1246},
doi = {https://doi.org/10.1002/widm.1246},
url = {https://wires.onlinelibrary.wiley.com/doi/abs/10.1002/widm.1246},
eprint = {https://wires.onlinelibrary.wiley.com/doi/pdf/10.1002/widm.1246},
year = {2018}
}

@ARTICLE{cai2025llmmoe,
  author={Cai, Weilin and Jiang, Juyong and Wang, Fan and Tang, Jing and Kim, Sunghun and Huang, Jiayi},
  journal={IEEE Transactions on Knowledge and Data Engineering}, 
  title={A Survey on Mixture of Experts in Large Language Models}, 
  year={2025},
  volume={37},
  number={7},
  pages={3896-3915},
  doi={10.1109/TKDE.2025.3554028}}

@article{cai2024survey,
  title   = {A Survey on Mixture of Experts},
  author  = {Cai, Weiguo and Jiang, Jialu and Wang, Fang and Tang, Jie and Kim, Sungchul and Huang, Jianbin},
  journal = {arXiv preprint},
  year    = {2024},
  eprint  = {2407.06204},
  archivePrefix = {arXiv},
  primaryClass  = {cs.LG}
}

@article{gan2025bigdata,
  title   = {Mixture of Experts ({MoE}): A Big Data Perspective},
  author  = {Gan, Weilong and Ning, Zhiyong and Qi, Zhiqiang and Yu, Philip S.},
  journal = {arXiv preprint},
  year    = {2025},
  eprint  = {2501.16352},
  archivePrefix = {arXiv},
  primaryClass  = {cs.LG}
}

@article{dimitri2025survey,
  title   = {A Survey on Mixture of Experts: Advancements, Challenges, and Future Directions},
  author  = {Dimitri, Vasily and Regina, Barbara and Alfonz, Magdolna},
  journal = {TechRxiv Preprints},
  year    = {2025}
}

@inproceedings{shazeer2017sparsemoe,
title={Outrageously Large Neural Networks: The Sparsely-Gated Mixture-of-Experts Layer},
author={Noam Shazeer and Azalia Mirhoseini and Krzysztof Maziarz and Andy Davis and Quoc Le and Geoffrey Hinton and Jeff Dean},
booktitle={International Conference on Learning Representations},
year={2017},
url={https://openreview.net/forum?id=B1ckMDqlg}
}

@inproceedings{lepikhin2021gshard,
  title     = {{GShard}: Scaling Giant Models with Conditional Computation and Automatic Sharding},
  author    = {Lepikhin, Dmitry and Lee, HyoukJoong and Xu, Yuanzhong and Chen, Dehao and Firat, Orhan and Huang, Yanping and Krikun, Maxim and Shazeer, Noam and Chen, Zhifeng},
  booktitle = {International Conference on Learning Representations},
  year      = {2021}
}

@article{SwitchTransformers,
  author  = {William Fedus and Barret Zoph and Noam Shazeer},
  title   = {Switch {Transformers}: Scaling to Trillion Parameter Models with Simple and Efficient Sparsity},
  journal = {Journal of Machine Learning Research},
  year    = {2022},
  volume  = {23},
  number  = {120},
  pages   = {1--39},
  url     = {http://jmlr.org/papers/v23/21-0998.html}
}

@InProceedings{du2022glam,
  title = 	 {{GL}a{M}: Efficient Scaling of Language Models with Mixture-of-Experts},
  author =       {Du, Nan and Huang, Yanping and Dai, Andrew M and Tong, Simon and Lepikhin, Dmitry and Xu, Yuanzhong and Krikun, Maxim and Zhou, Yanqi and Yu, Adams Wei and Firat, Orhan and Zoph, Barret and Fedus, Liam and Bosma, Maarten P and Zhou, Zongwei and Wang, Tao and Wang, Emma and Webster, Kellie and Pellat, Marie and Robinson, Kevin and Meier-Hellstern, Kathleen and Duke, Toju and Dixon, Lucas and Zhang, Kun and Le, Quoc and Wu, Yonghui and Chen, Zhifeng and Cui, Claire},
  booktitle = 	 {Proceedings of the 39th International Conference on Machine Learning},
  pages = 	 {5547--5569},
  year = 	 {2022},
  editor = 	 {Chaudhuri, Kamalika and Jegelka, Stefanie and Song, Le and Szepesvari, Csaba and Niu, Gang and Sabato, Sivan},
  volume = 	 {162},
  series = 	 {Proceedings of Machine Learning Research},
  month = 	 {17--23 Jul},
  publisher =    {PMLR},
  url = 	 {https://proceedings.mlr.press/v162/du22c.html},
}

@inproceedings{riquelme2021vmoe,
 author = {Riquelme, Carlos and Puigcerver, Joan and Mustafa, Basil and Neumann, Maxim and Jenatton, Rodolphe and Susano Pinto, Andr\'{e} and Keysers, Daniel and Houlsby, Neil},
 booktitle = {Advances in Neural Information Processing Systems},
 editor = {M. Ranzato and A. Beygelzimer and Y. Dauphin and P.S. Liang and J. Wortman Vaughan},
 pages = {8583--8595},
 publisher = {Curran Associates, Inc.},
 title = {Scaling Vision with Sparse Mixture of Experts},
 url = {https://proceedings.neurips.cc/paper_files/paper/2021/file/48237d9f2dea8c74c2a72126cf63d933-Paper.pdf},
 volume = {34},
 year = {2021}
}

@inproceedings{mustafa2022limoe,
 author = {Mustafa, Basil and Riquelme, Carlos and Puigcerver, Joan and Jenatton, Rodolphe and Houlsby, Neil},
 booktitle = {Advances in Neural Information Processing Systems},
 editor = {S. Koyejo and S. Mohamed and A. Agarwal and D. Belgrave and K. Cho and A. Oh},
 pages = {9564--9576},
 publisher = {Curran Associates, Inc.},
 title = {Multimodal Contrastive Learning with LIMoE: the Language-Image Mixture of Experts},
 url = {https://proceedings.neurips.cc/paper_files/paper/2022/file/3e67e84abf900bb2c7cbd5759bfce62d-Paper-Conference.pdf},
 volume = {35},
 year = {2022}
}

@article{zoph2022stmoe,
  title   = {{ST}-{MoE}: Designing Stable and Transferable Sparse Expert Models},
  author  = {Zoph, Barret and Bello, Irwan and Kumar, Sameer and Du, Nan and Huang, Yanping and Dean, Jeff and Shazeer, Noam and Fedus, William},
  journal = {arXiv preprint},
  year    = {2022},
  eprint  = {2202.08906},
  archivePrefix = {arXiv},
  primaryClass  = {cs.LG}
}

@inproceedings{zhou2022expertchoice,
 author = {Zhou, Yanqi and Lei, Tao and Liu, Hanxiao and Du, Nan and Huang, Yanping and Zhao, Vincent and Dai, Andrew M and Chen, zhifeng and Le, Quoc V and Laudon, James},
 booktitle = {Advances in Neural Information Processing Systems},
 editor = {S. Koyejo and S. Mohamed and A. Agarwal and D. Belgrave and K. Cho and A. Oh},
 pages = {7103--7114},
 publisher = {Curran Associates, Inc.},
 title = {Mixture-of-Experts with Expert Choice Routing},
 url = {https://proceedings.neurips.cc/paper_files/paper/2022/file/2f00ecd787b432c1d36f3de9800728eb-Paper-Conference.pdf},
 volume = {35},
 year = {2022}
}

@inproceedings{chen2022understanding,
 author = {Chen, Zixiang and Deng, Yihe and Wu, Yue and Gu, Quanquan and Li, Yuanzhi},
 booktitle = {Advances in Neural Information Processing Systems},
 editor = {S. Koyejo and S. Mohamed and A. Agarwal and D. Belgrave and K. Cho and A. Oh},
 pages = {23049--23062},
 publisher = {Curran Associates, Inc.},
 title = {Towards Understanding the Mixture-of-Experts Layer in Deep Learning},
 url = {https://proceedings.neurips.cc/paper_files/paper/2022/file/91edff07232fb1b55a505a9e9f6c0ff3-Paper-Conference.pdf},
 volume = {35},
 year = {2022}
}

@inproceedings{hazimeh2021dselectk,
 author = {Hazimeh, Hussein and Zhao, Zhe and Chowdhery, Aakanksha and Sathiamoorthy, Maheswaran and Chen, Yihua and Mazumder, Rahul and Hong, Lichan and Chi, Ed},
 booktitle = {Advances in Neural Information Processing Systems},
 editor = {M. Ranzato and A. Beygelzimer and Y. Dauphin and P.S. Liang and J. Wortman Vaughan},
 pages = {29335--29347},
 publisher = {Curran Associates, Inc.},
 title = {{DSelect-k}: Differentiable Selection in the Mixture of Experts with Applications to Multi-Task Learning},
 url = {https://proceedings.neurips.cc/paper_files/paper/2021/file/f5ac21cd0ef1b88e9848571aeb53551a-Paper.pdf},
 volume = {34},
 year = {2021}
}

@article{he2021fastmoe,
  title   = {{FastMoE}: A Fast Mixture-of-Expert Training System},
  author  = {He, Jiaao and Qiu, Jiezhong and Zeng, Aohan and Yang, Zhilin and Zhai, Jidong and Tang, Jie},
  journal = {arXiv preprint},
  year    = {2021},
  eprint  = {2103.13262},
  archivePrefix = {arXiv},
  primaryClass  = {cs.LG}
}

@InProceedings{rajbhandari2022deepspeedmoe,
  title = 	 {{D}eep{S}peed-{M}o{E}: Advancing Mixture-of-Experts Inference and Training to Power Next-Generation {AI} Scale},
  author =       {Rajbhandari, Samyam and Li, Conglong and Yao, Zhewei and Zhang, Minjia and Aminabadi, Reza Yazdani and Awan, Ammar Ahmad and Rasley, Jeff and He, Yuxiong},
  booktitle = 	 {Proceedings of the 39th International Conference on Machine Learning},
  pages = 	 {18332--18346},
  year = 	 {2022},
  editor = 	 {Chaudhuri, Kamalika and Jegelka, Stefanie and Song, Le and Szepesvari, Csaba and Niu, Gang and Sabato, Sivan},
  volume = 	 {162},
  series = 	 {Proceedings of Machine Learning Research},
  month = 	 {17--23 Jul},
  publisher =    {PMLR},
  url = 	 {https://proceedings.mlr.press/v162/rajbhandari22a.html}
}

@article{gale2023megablocks,
  title   = {{MegaBlocks}: Efficient Sparse Training with Mixture-of-Experts},
  author  = {Gale, Trevor and Elsen, Erich and Hooker, Sara},
  journal = {arXiv preprint},
  year    = {2023},
  eprint  = {2211.15841},
  archivePrefix = {arXiv},
  primaryClass  = {cs.LG}
}

@inproceedings{he2022fastermoe,
  title={{FasterMoE}: modeling and optimizing training of large-scale dynamic pre-trained models},
  author={He, Jiaao and Zhai, Jidong and Antunes, Tiago and Wang, Haojie and Luo, Fuwen and Shi, Shangfeng and Li, Qin},
  booktitle={Proceedings of the 27th ACM SIGPLAN Symposium on Principles and Practice of Parallel Programming},
  pages={120--134},
  year={2022}
}

@InProceedings{unet,
author="Ronneberger, Olaf
and Fischer, Philipp
and Brox, Thomas",
editor="Navab, Nassir
and Hornegger, Joachim
and Wells, William M.
and Frangi, Alejandro F.",
title="U-Net: Convolutional Networks for Biomedical Image Segmentation",
booktitle="Medical Image Computing and Computer-Assisted Intervention -- MICCAI 2015",
year="2015",
publisher="Springer International Publishing",
address="Cham",
pages="234--241",
isbn="978-3-319-24574-4"
}

@article{mamba,
  title={Mamba: Linear-Time Sequence Modeling with Selective State Spaces},
  author={Gu, Albert and Dao, Tri},
  journal={arXiv preprint arXiv:2312.00752},
  year={2023}
}

@inproceedings{mamba2,
  title={Transformers are {SSM}s: Generalized Models and Efficient Algorithms Through Structured State Space Duality},
  author={Dao, Tri and Gu, Albert},
  booktitle={International Conference on Machine Learning (ICML)},
  year={2024}
}

@article{nie2022hetumoe,
  title   = {{HetuMoE}: An Efficient Trillion-scale Mixture-of-Expert Distributed Training System},
  author  = {Nie, Xiaonan and Zhao, Pinxue and Miao, Xupeng and Zhao, Tong and Cui, Bin},
  journal = {arXiv preprint},
  year    = {2022},
  eprint  = {2203.14685},
  archivePrefix = {arXiv},
  primaryClass  = {cs.DC}
}

@INPROCEEDINGS{hwang2024pregatedmoe,
  author={Hwang, Ranggi and Wei, Jianyu and Cao, Shijie and Hwang, Changho and Tang, Xiaohu and Cao, Ting and Yang, Mao},
  booktitle={2024 ACM/IEEE 51st Annual International Symposium on Computer Architecture (ISCA)}, 
  title={Pre-gated MoE: An Algorithm-System Co-Design for Fast and Scalable Mixture-of-Expert Inference}, 
  year={2024},
  volume={},
  number={},
  pages={1018-1031},
  doi={10.1109/ISCA59077.2024.00078}}

@inproceedings{hwang2023tutel,
 author = {Hwang, Changho and Cui, Wei and Xiong, Yifan and Yang, Ziyue and Liu, Ze and Hu, Han and Wang, Zilong and Salas, Rafael and Jose, Jithin and Ram, Prabhat and Chau, HoYuen and Cheng, Peng and Yang, Fan and Yang, Mao and Xiong, Yongqiang},
 booktitle = {Proceedings of Machine Learning and Systems},
 editor = {D. Song and M. Carbin and T. Chen},
 pages = {269--287},
 publisher = {Curan},
 title = {Tutel: Adaptive Mixture-of-Experts at Scale},
 url = {https://proceedings.mlsys.org/paper_files/paper/2023/file/5616d34cf8ff73942cfd5aa922842556-Paper-mlsys2023.pdf},
 volume = {5},
 year = {2023}
}

@inproceedings{ma2018mmoe,
author = {Ma, Jiaqi and Zhao, Zhe and Yi, Xinyang and Chen, Jilin and Hong, Lichan and Chi, Ed H.},
title = {Modeling Task Relationships in Multi-task Learning with Multi-gate Mixture-of-Experts},
year = {2018},
isbn = {9781450355520},
publisher = {Association for Computing Machinery},
address = {New York, NY, USA},
url = {https://doi.org/10.1145/3219819.3220007},
doi = {10.1145/3219819.3220007},
booktitle = {Proceedings of the 24th ACM SIGKDD International Conference on Knowledge Discovery \& Data Mining},
pages = {1930–1939},
numpages = {10},
location = {London, United Kingdom},
series = {KDD '18}
}

@inproceedings{tang2020ple,
author = {Tang, Hongyan and Liu, Junning and Zhao, Ming and Gong, Xudong},
title = {Progressive Layered Extraction (PLE): A Novel Multi-Task Learning (MTL) Model for Personalized Recommendations},
year = {2020},
isbn = {9781450375832},
publisher = {Association for Computing Machinery},
address = {New York, NY, USA},
url = {https://doi.org/10.1145/3383313.3412236},
doi = {10.1145/3383313.3412236},
booktitle = {Proceedings of the 14th ACM Conference on Recommender Systems},
pages = {269–278},
numpages = {10},
location = {Virtual Event, Brazil},
series = {RecSys '20}
}

@article{gupta2022sparsemtl,
  title   = {Sparsely Activated Mixture-of-Experts are Robust Multi-Task Learners},
  author  = {Gupta, Shashank and Mukherjee, Subhabrata and Subudhi, Krishan and Gonzalez, Eduardo and Jose, Damien and Awadallah, Ahmed H. and Gao, Jianfeng},
  journal = {arXiv preprint},
  year    = {2022},
  eprint  = {2204.07689},
  archivePrefix = {arXiv},
  primaryClass  = {cs.LG}
}

@INPROCEEDINGS{chen2023modsquad,
  author={Chen, Zitian and Shen, Yikang and Ding, Mingyu and Chen, Zhenfang and Zhao, Hengshuang and Learned-Miller, Erik and Gan, Chuang},
  booktitle={2023 IEEE/CVF Conference on Computer Vision and Pattern Recognition (CVPR)}, 
  title={{Mod-Squad}: Designing Mixtures of Experts As Modular Multi-Task Learners}, 
  year={2023},
  volume={},
  number={},
  pages={11828-11837},
  doi={10.1109/CVPR52729.2023.01138}}

@inproceedings{kudugunta2021taskmoe,
    title = "Beyond {Distillation}: Task-level Mixture-of-Experts for Efficient Inference",
    author = "Kudugunta, Sneha  and
      Huang, Yanping  and
      Bapna, Ankur  and
      Krikun, Maxim  and
      Lepikhin, Dmitry  and
      Luong, Minh-Thang  and
      Firat, Orhan",
    editor = "Moens, Marie-Francine  and
      Huang, Xuanjing  and
      Specia, Lucia  and
      Yih, Scott Wen-tau",
    booktitle = "Findings of the Association for Computational Linguistics: EMNLP 2021",
    month = nov,
    year = "2021",
    address = "Punta Cana, Dominican Republic",
    publisher = "Association for Computational Linguistics",
    url = "https://aclanthology.org/2021.findings-emnlp.304/",
    doi = "10.18653/v1/2021.findings-emnlp.304",
    pages = "3577--3599",
}

@inproceedings{dai2024deepseekmoe,
    title = "{D}eep{S}eek{M}o{E}: Towards Ultimate Expert Specialization in Mixture-of-Experts Language Models",
    author = "Dai, Damai  and
      Deng, Chengqi  and
      Zhao, Chenggang  and
      Xu, R.x.  and
      Gao, Huazuo  and
      Chen, Deli  and
      Li, Jiashi  and
      Zeng, Wangding  and
      Yu, Xingkai  and
      Wu, Y.  and
      Xie, Zhenda  and
      Li, Y.k.  and
      Huang, Panpan  and
      Luo, Fuli  and
      Ruan, Chong  and
      Sui, Zhifang  and
      Liang, Wenfeng",
    editor = "Ku, Lun-Wei  and
      Martins, Andre  and
      Srikumar, Vivek",
    booktitle = "Proceedings of the 62nd Annual Meeting of the Association for Computational Linguistics (Volume 1: Long Papers)",
    month = aug,
    year = "2024",
    address = "Bangkok, Thailand",
    publisher = "Association for Computational Linguistics",
    url = "https://aclanthology.org/2024.acl-long.70/",
    doi = "10.18653/v1/2024.acl-long.70",
    pages = "1280--1297",
}

@article{jiang2024mixtral,
  title   = {Mixtral of Experts},
  author  = {Jiang, Albert Q. and Sablayrolles, Alexandre and Roux, Antoine and Mensch, Arthur and Savary, Blanche and Bamford, Chris and Chaplot, Devendra Singh and de las Casas, Diego and Bou Hanna, Emma and Bressand, Florian and others},
  journal = {arXiv preprint},
  year    = {2024},
  eprint  = {2401.04088},
  archivePrefix = {arXiv},
  primaryClass  = {cs.CL}
}

@article{muennighoff2024olmoe,
  title   = {{OLMoE}: Open Mixture-of-Experts Language Models},
  author  = {Muennighoff, Niklas and Soldaini, Luca and Groeneveld, Dirk and Lo, Kyle and Morrison, Jacob and Min, Sewon and Shi, Weijia and Walsh, Pete and Tafjord, Oyvind and Lambert, Nathan and others},
  journal = {arXiv preprint},
  year    = {2024},
  eprint  = {2409.02060},
  archivePrefix = {arXiv},
  primaryClass  = {cs.CL}
}

@inproceedings{hu2022lora,
title={Lo{RA}: Low-Rank Adaptation of Large Language Models},
author={Edward J Hu and Yelong Shen and Phillip Wallis and Zeyuan Allen-Zhu and Yuanzhi Li and Shean Wang and Lu Wang and Weizhu Chen},
booktitle={International Conference on Learning Representations},
year={2022},
}

@article{TTLoRAMoE,
  title={{TT-LoRA MoE}: Unifying Parameter-Efficient Fine-Tuning and Sparse Mixture-of-Experts},
  author={Kunwar, Pradip and Vu, Minh N and Gupta, Maanak and Abdelsalam, Mahmoud and Bhattarai, Manish},
  journal={arXiv preprint arXiv:2504.21190},
  year={2025}
}

@article{wu2024mole,
  title   = {Mixture of {LoRA} Experts},
  author  = {Wu, Yiming and others},
  journal = {arXiv preprint},
  year    = {2024},
  eprint  = {2404.13628},
  archivePrefix = {arXiv},
  primaryClass  = {cs.LG}
}

@inproceedings{liao2024hmora,
  title={{HMoRA}: Making LLMs More Effective with Hierarchical Mixture of LoRA Experts},
  author={Liao, Mengqi and Chen, Wei and Shen, Junfeng and Guo, Shengnan and Wan, Huaiyu},
  booktitle={The Thirteenth International Conference on Learning Representations},
  year={2025}
}

@inproceedings{huang2024harder,
    title = "Harder Task Needs More Experts: Dynamic Routing in {M}o{E} Models",
    author = "Huang, Quzhe  and
      An, Zhenwei  and
      Zhuang, Nan  and
      Tao, Mingxu  and
      Zhang, Chen  and
      Jin, Yang  and
      Xu, Kun  and
      Xu, Kun  and
      Chen, Liwei  and
      Huang, Songfang  and
      Feng, Yansong",
    editor = "Ku, Lun-Wei  and
      Martins, Andre  and
      Srikumar, Vivek",
    booktitle = "Proceedings of the 62nd Annual Meeting of the Association for Computational Linguistics (Volume 1: Long Papers)",
    month = aug,
    year = "2024",
    address = "Bangkok, Thailand",
    publisher = "Association for Computational Linguistics",
    url = "https://aclanthology.org/2024.acl-long.696/",
    doi = "10.18653/v1/2024.acl-long.696",
    pages = "12883--12895",
}

@InProceedings{lewis2021base,
  title = 	 {{BASE} Layers: Simplifying Training of Large, Sparse Models},
  author =       {Lewis, Mike and Bhosale, Shruti and Dettmers, Tim and Goyal, Naman and Zettlemoyer, Luke},
  booktitle = 	 {Proceedings of the 38th International Conference on Machine Learning},
  pages = 	 {6265--6274},
  year = 	 {2021},
  editor = 	 {Meila, Marina and Zhang, Tong},
  volume = 	 {139},
  series = 	 {Proceedings of Machine Learning Research},
  month = 	 {18--24 Jul},
  publisher =    {PMLR},
  url = 	 {https://proceedings.mlr.press/v139/lewis21a.html}
}

@inproceedings{roller2021hash,
author = {Roller, Stephen and Sukhbaatar, Sainbayar and Szlam, Arthur and Weston, Jason},
title = {Hash layers for large sparse models},
year = {2021},
isbn = {9781713845393},
publisher = {Curran Associates Inc.},
address = {Red Hook, NY, USA},
booktitle = {Proceedings of the 35th International Conference on Neural Information Processing Systems},
articleno = {1343},
numpages = {12},
series = {NIPS '21}
}

@inproceedings{zuo2022stochastic,
title={Taming Sparsely Activated Transformer with Stochastic Experts},
author={Simiao Zuo and Xiaodong Liu and Jian Jiao and Young Jin Kim and Hany Hassan and Ruofei Zhang and Jianfeng Gao and Tuo Zhao},
booktitle={International Conference on Learning Representations},
year={2022},
url={https://openreview.net/forum?id=B72HXs80q4}
}

@inproceedings{komatsuzaki2023sparseupcycling,
title={Sparse {Upcycling}: Training Mixture-of-Experts from Dense Checkpoints},
author={Aran Komatsuzaki and Joan Puigcerver and James Lee-Thorp and Carlos Riquelme Ruiz and Basil Mustafa and Joshua Ainslie and Yi Tay and Mostafa Dehghani and Neil Houlsby},
booktitle={The Eleventh International Conference on Learning Representations },
year={2023},
url={https://openreview.net/forum?id=T5nUQDrM4u}
}

@inproceedings{zhu2022uniperceiver,
author = {Zhu, Jinguo and Zhu, Xizhou and Wang, Wenhai and Wang, Xiaohua and Li, Hongsheng and Wang, Xiaogang and Dai, Jifeng},
title = {{Uni-perceiver-MoE}: learning sparse generalist models with conditional MoEs},
year = {2022},
isbn = {9781713871088},
publisher = {Curran Associates Inc.},
address = {Red Hook, NY, USA},
booktitle = {Proceedings of the 36th International Conference on Neural Information Processing Systems},
articleno = {193},
numpages = {15},
location = {New Orleans, LA, USA},
series = {NIPS '22}
}

@article{ho2022gaussianmoe,
  author  = {Nhat Ho and Chiao-Yu Yang and Michael I. Jordan},
  title   = {Convergence Rates for Gaussian Mixtures of Experts},
  journal = {Journal of Machine Learning Research},
  year    = {2022},
  volume  = {23},
  number  = {323},
  pages   = {1--81},
  url     = {http://jmlr.org/papers/v23/20-1129.html}
}

@inproceedings{nguyen2023demystify,
 author = {Nguyen, Huy and Nguyen, TrungTin and Ho, Nhat},
 booktitle = {Advances in Neural Information Processing Systems},
 editor = {A. Oh and T. Naumann and A. Globerson and K. Saenko and M. Hardt and S. Levine},
 pages = {4624--4652},
 publisher = {Curran Associates, Inc.},
 title = {Demystifying Softmax Gating Function in Gaussian Mixture of Experts},
 url = {https://proceedings.neurips.cc/paper_files/paper/2023/file/0ef6ffcb85a2d238fc4761860c31ded4-Paper-Conference.pdf},
 volume = {36},
 year = {2023}
}

@inproceedings{nguyen2023topk,
title={Statistical Perspective of Top-K Sparse Softmax Gating Mixture of Experts},
author={Huy Nguyen and Pedram Akbarian and Fanqi Yan and Nhat Ho},
booktitle={The Twelfth International Conference on Learning Representations},
year={2024},
}

@inproceedings{nguyen2023generalgating,
title={A General Theory for Softmax Gating Multinomial Logistic Mixture of Experts},
author={Huy Nguyen and Pedram Akbarian and TrungTin Nguyen and Nhat Ho},
booktitle={Forty-first International Conference on Machine Learning},
year={2024},
url={https://openreview.net/forum?id=2Sl0lPF6ka}
}

@inproceedings{shi2025timemoe,
title={Time-MoE: Billion-Scale Time Series Foundation Models with Mixture of Experts},
author={Xiaoming Shi and Shiyu Wang and Yuqi Nie and Dianqi Li and Zhou Ye and Qingsong Wen and Ming Jin},
booktitle={The Thirteenth International Conference on Learning Representations},
year={2025},
url={https://openreview.net/forum?id=e1wDDFmlVu}
}

@article{zhong2024lory,
  title   = {{Lory}: Fully Differentiable Mixture-of-Experts for Autoregressive Language Model Pre-training},
  author  = {Zhong, Zexuan and Xia, Mengzhou and Chen, Danqi and Lewis, Mike},
  journal = {arXiv preprint},
  year    = {2024},
  eprint  = {2405.03133},
  archivePrefix = {arXiv},
  primaryClass  = {cs.CL}
}

@misc{wang2024auxiliarylossfree,
  title         = {Auxiliary-Loss-Free Load Balancing Strategy for Mixture-of-Experts},
  author        = {Lean Wang and Huazuo Gao and Chenggang Zhao and Xu Sun and Damai Dai},
  year          = {2024},
  eprint        = {2408.15664},
  archivePrefix = {arXiv},
  primaryClass  = {cs.LG},
  url           = {https://arxiv.org/abs/2408.15664}
}

@inproceedings{thaman2025rebalancing,
title={One Must Imagine Experts Happy: Rebalancing Neural Routers via Constrained Optimization},
author={Kushal Thaman},
booktitle={Sparsity in LLMs (SLLM): Deep Dive into Mixture of Experts, Quantization, Hardware, and Inference},
year={2025},
url={https://openreview.net/forum?id=gsAkhArtfT}
}

@misc{omi2025similarityrouter,
  title         = {Load Balancing Mixture of Experts with Similarity Preserving Routers},
  author        = {Nabil Omi and Siddhartha Sen and Ali Farhadi},
  year          = {2025},
  eprint        = {2506.14038},
  archivePrefix = {arXiv},
  primaryClass  = {cs.LG},
  url           = {https://arxiv.org/abs/2506.14038}
}

@inproceedings{do2025simsmoe,
    title = "{S}im{SM}o{E}: Toward Efficient Training Mixture of Experts via Solving Representational Collapse",
    author = "Do, Giang  and
      Le, Hung  and
      Tran, Truyen",
    editor = "Chiruzzo, Luis  and
      Ritter, Alan  and
      Wang, Lu",
    booktitle = "Findings of the Association for Computational Linguistics: NAACL 2025",
    month = apr,
    year = "2025",
    address = "Albuquerque, New Mexico",
    publisher = "Association for Computational Linguistics",
    url = "https://aclanthology.org/2025.findings-naacl.107/",
    doi = "10.18653/v1/2025.findings-naacl.107",
    pages = "2012--2025",
    ISBN = "979-8-89176-195-7",
}

@inproceedings{chen2023smoedropout,
title={Sparse MoE as the New Dropout: Scaling Dense and Self-Slimmable Transformers},
author={Tianlong Chen and Zhenyu Zhang and AJAY KUMAR JAISWAL and Shiwei Liu and Zhangyang Wang},
booktitle={The Eleventh International Conference on Learning Representations },
year={2023},
url={https://openreview.net/forum?id=w1hwFUb_81}
}

@article{xie2023moec, 
title={MoEC: Mixture of Expert Clusters}, 
author={Xie, Yuan and Huang, Shaohan and Chen, Tianyu and Wei, Furu},
journal={Proceedings of the AAAI Conference on Artificial Intelligence},
year={2023},
volume={37},
number={11},
pages={13807-13815},
url={https://ojs.aaai.org/index.php/AAAI/article/view/26617}
}

@inproceedings{zuo2022moebert,
    title = "{M}o{EBERT}: from {BERT} to Mixture-of-Experts via Importance-Guided Adaptation",
    author = "Zuo, Simiao  and
      Zhang, Qingru  and
      Liang, Chen  and
      He, Pengcheng  and
      Zhao, Tuo  and
      Chen, Weizhu",
    editor = "Carpuat, Marine  and
      de Marneffe, Marie-Catherine  and
      Meza Ruiz, Ivan Vladimir",
    booktitle = "Proceedings of the 2022 Conference of the North American Chapter of the Association for Computational Linguistics: Human Language Technologies",
    month = jul,
    year = "2022",
    address = "Seattle, United States",
    publisher = "Association for Computational Linguistics",
    url = "https://aclanthology.org/2022.naacl-main.116/",
    doi = "10.18653/v1/2022.naacl-main.116",
    pages = "1610--1623",
}

@inproceedings{szatkowski2024d2dmoe,
author = {Szatkowski, Filip and W\'{o}jcik, Bartosz and Pi\'{o}rczy\'{n}ski, Miko\l{}aj and Scardapane, Simone},
title = {Exploiting activation sparsity with dense to dynamic-k mixture-of-experts conversion},
year = {2024},
isbn = {9798331314385},
publisher = {Curran Associates Inc.},
address = {Red Hook, NY, USA},
booktitle = {Proceedings of the 38th International Conference on Neural Information Processing Systems},
articleno = {1370},
numpages = {29},
location = {Vancouver, BC, Canada},
series = {NIPS '24}
}

@misc{gao2025tomoe,
  title         = {{ToMoE}: Converting Dense Large Language Models to Mixture-of-Experts through Dynamic Structural Pruning},
  author        = {Shangqian Gao and Ting Hua and Reza Shirkavand and Chi-Heng Lin and Zhen Tang and Zhengao Li and Longge Yuan and Fangyi Li and Zeyu Zhang and Alireza Ganjdanesh and Lou Qian and Xu Jie and Yen-Chang Hsu},
  year          = {2025},
  eprint        = {2501.15316},
  archivePrefix = {arXiv},
  primaryClass  = {cs.LG},
  url           = {https://arxiv.org/abs/2501.15316}
}

@misc{nussbaum2025trainingsparse,
  title         = {Training Sparse Mixture Of Experts Text Embedding Models},
  author        = {Zach Nussbaum and Brandon Duderstadt},
  year          = {2025},
  eprint        = {2502.07972},
  archivePrefix = {arXiv},
  primaryClass  = {cs.CL},
  url           = {https://arxiv.org/abs/2502.07972}
}

@misc{gu2025elasticmoe,
  title         = {Elastic {MoE}: Unlocking the Inference-Time Scalability of Mixture-of-Experts},
  author        = {Naibin Gu and Zhenyu Zhang and Yuchen Feng and Yilong Chen and Peng Fu and Zheng Lin and Shuohuan Wang and Yu Sun and Hua Wu and Weiping Wang and Haifeng Wang},
  year          = {2025},
  eprint        = {2509.21892},
  archivePrefix = {arXiv},
  primaryClass  = {cs.LG},
  url           = {https://arxiv.org/abs/2509.21892}
}

@misc{ma2025stabilizingmoerl,
  title         = {Stabilizing MoE Reinforcement Learning by Aligning Training and Inference Routers},
  author        = {Wenhan Ma and Hailin Zhang and Liang Zhao and Yifan Song and Yudong Wang and Zhifang Sui and Fuli Luo},
  year          = {2025},
  eprint        = {2510.11370},
  archivePrefix = {arXiv},
  primaryClass  = {cs.LG},
  url           = {https://arxiv.org/abs/2510.11370}
}

@misc{gururangan2023scalingexpert,
  title         = {Scaling Expert Language Models with Unsupervised Domain Discovery},
  author        = {Suchin Gururangan and Margaret Li and Mike Lewis and Weijia Shi and Tim Althoff and Noah A. Smith and Luke Zettlemoyer},
  year          = {2023},
  eprint        = {2303.14177},
  archivePrefix = {arXiv},
  primaryClass  = {cs.CL},
  url           = {https://arxiv.org/abs/2303.14177}
}

@inproceedings{jawahar2023automoe,
    title = "{A}uto{M}o{E}: Heterogeneous Mixture-of-Experts with Adaptive Computation for Efficient Neural Machine Translation",
    author = "Jawahar, Ganesh  and
      Mukherjee, Subhabrata  and
      Liu, Xiaodong  and
      Kim, Young Jin  and
      Abdul-Mageed, Muhammad  and
      Lakshmanan, V.S., Laks  and
      Awadallah, Ahmed Hassan  and
      Bubeck, Sebastien  and
      Gao, Jianfeng",
    editor = "Rogers, Anna  and
      Boyd-Graber, Jordan  and
      Okazaki, Naoaki",
    booktitle = "Findings of the Association for Computational Linguistics: ACL 2023",
    month = jul,
    year = "2023",
    address = "Toronto, Canada",
    publisher = "Association for Computational Linguistics",
    url = "https://aclanthology.org/2023.findings-acl.580/",
    doi = "10.18653/v1/2023.findings-acl.580",
    pages = "9116--9132",
}

@inproceedings{BERT,
    title = "{BERT}: Pre-training of Deep Bidirectional Transformers for Language Understanding",
    author = "Devlin, Jacob  and
      Chang, Ming-Wei  and
      Lee, Kenton  and
      Toutanova, Kristina",
    editor = "Burstein, Jill  and
      Doran, Christy  and
      Solorio, Thamar",
    booktitle = "Proceedings of the 2019 Conference of the North {A}merican Chapter of the Association for Computational Linguistics: Human Language Technologies, Volume 1 (Long and Short Papers)",
    month = jun,
    year = "2019",
    address = "Minneapolis, Minnesota",
    publisher = "Association for Computational Linguistics",
    url = "https://aclanthology.org/N19-1423/",
    doi = "10.18653/v1/N19-1423",
    pages = "4171--4186",
}

@ARTICLE{zhu2017deeplearning,
  author={Zhu, Xiao Xiang and Tuia, Devis and Mou, Lichao and Xia, Gui-Song and Zhang, Liangpei and Xu, Feng and Fraundorfer, Friedrich},
  journal={IEEE Geoscience and Remote Sensing Magazine}, 
  title={Deep Learning in Remote Sensing: A Comprehensive Review and List of Resources}, 
  year={2017},
  volume={5},
  number={4},
  pages={8-36},
  doi={10.1109/MGRS.2017.2762307}}

@article{ma2019deeprs,
title = {Deep learning in remote sensing applications: A meta-analysis and review},
journal = {ISPRS Journal of Photogrammetry and Remote Sensing},
volume = {152},
pages = {166-177},
year = {2019},
issn = {0924-2716},
doi = {https://doi.org/10.1016/j.isprsjprs.2019.04.015},
url = {https://www.sciencedirect.com/science/article/pii/S0924271619301108},
author = {Lei Ma and Yu Liu and Xueliang Zhang and Yuanxin Ye and Gaofei Yin and Brian Alan Johnson},
}

@Article{jiang2022cdsurvey,
AUTHOR = {Jiang, Huiwei and Peng, Min and Zhong, Yuanjun and Xie, Haofeng and Hao, Zemin and Lin, Jingming and Ma, Xiaoli and Hu, Xiangyun},
TITLE = {A Survey on Deep Learning-Based Change Detection from High-Resolution Remote Sensing Images},
JOURNAL = {Remote Sensing},
VOLUME = {14},
YEAR = {2022},
NUMBER = {7},
ARTICLE-NUMBER = {1552},
URL = {https://www.mdpi.com/2072-4292/14/7/1552},
ISSN = {2072-4292},
DOI = {10.3390/rs14071552}
}

@ARTICLE{gao2020fusion,
  author={Gao, Jing and Li, Peng and Chen, Zhikui and Zhang, Jianing},
  journal={Neural Computation}, 
  title={A Survey on Deep Learning for Multimodal Data Fusion}, 
  year={2020},
  volume={32},
  number={5},
  pages={829-864},
  doi={10.1162/neco_a_01273}}

@article{osco2021uav,
title = {A review on deep learning in UAV remote sensing},
journal = {International Journal of Applied Earth Observation and Geoinformation},
volume = {102},
pages = {102456},
year = {2021},
issn = {1569-8432},
doi = {https://doi.org/10.1016/j.jag.2021.102456},
url = {https://www.sciencedirect.com/science/article/pii/S030324342100163X},
author = {Lucas Prado Osco and José {Marcato Junior} and Ana Paula {Marques Ramos} and Lúcio André {de Castro Jorge} and Sarah Narges Fatholahi and Jonathan {de Andrade Silva} and Edson Takashi Matsubara and Hemerson Pistori and Wesley Nunes Gonçalves and Jonathan Li},
}

@Article{shafique2022cdsurvey,
AUTHOR = {Shafique, Ayesha and Cao, Guo and Khan, Zia and Asad, Muhammad and Aslam, Muhammad},
TITLE = {Deep Learning-Based Change Detection in Remote Sensing Images: A Review},
JOURNAL = {Remote Sensing},
VOLUME = {14},
YEAR = {2022},
NUMBER = {4},
ARTICLE-NUMBER = {871},
URL = {https://www.mdpi.com/2072-4292/14/4/871},
ISSN = {2072-4292},
DOI = {10.3390/rs14040871}
}

@ARTICLE{huang2025rsfm,
  author={Lu, Siqi and Guo, Junlin and Zimmer-Dauphinee, James R. and Nieusma, Jordan M. and Wang, Xiao and VanValkenburgh, Parker and Wernke, Steven A. and Huo, Yuankai},
  journal={IEEE Geoscience and Remote Sensing Magazine}, 
  title={Vision Foundation Models in Remote Sensing: A survey}, 
  year={2025},
  volume={13},
  number={3},
  pages={190-215},
  doi={10.1109/MGRS.2025.3541952}}

@Article{zhao2023lulcreview,
AUTHOR = {Zhao, Shengyu and Tu, Kaiwen and Ye, Shutong and Tang, Hao and Hu, Yaocong and Xie, Chao},
TITLE = {Land Use and Land Cover Classification Meets Deep Learning: A Review},
JOURNAL = {Sensors},
VOLUME = {23},
YEAR = {2023},
NUMBER = {21},
ARTICLE-NUMBER = {8966},
URL = {https://www.mdpi.com/1424-8220/23/21/8966},
PubMedID = {37960665},
ISSN = {1424-8220},
DOI = {10.3390/s23218966}
}

@article{chamroukhi2017skewt,
title = {Skew $t$ mixture of experts},
journal = {Neurocomputing},
volume = {266},
pages = {390-408},
year = {2017},
issn = {0925-2312},
doi = {https://doi.org/10.1016/j.neucom.2017.05.044},
url = {https://www.sciencedirect.com/science/article/pii/S0925231217308949},
author = {F. Chamroukhi},
}

@article{fung2022moe,
title = {Mixture of experts models for multilevel data: Modeling framework and approximation theory},
journal = {Neurocomputing},
volume = {626},
pages = {129357},
year = {2025},
issn = {0925-2312},
doi = {https://doi.org/10.1016/j.neucom.2025.129357},
url = {https://www.sciencedirect.com/science/article/pii/S0925231225000293},
author = {Tsz Chai Fung and Spark C. Tseung},
}

@inproceedings{miller1997moe,
author = {Miller, David J. and Uyar, Hasan S.},
title = {A mixture of experts classifier with learning based on both labelled and unlabelled data},
year = {1996},
publisher = {MIT Press},
address = {Cambridge, MA, USA},
booktitle = {Proceedings of the 10th International Conference on Neural Information Processing Systems},
pages = {571–577},
numpages = {7},
location = {Denver, Colorado},
series = {NIPS'96}
}

@InProceedings{wang2020deepmoe,
  title = 	 {Deep Mixture of Experts via Shallow Embedding},
  author =       {Wang, Xin and Yu, Fisher and Dunlap, Lisa and Ma, Yi-An and Wang, Ruth and Mirhoseini, Azalia and Darrell, Trevor and Gonzalez, Joseph E.},
  booktitle = 	 {Proceedings of The 35th Uncertainty in Artificial Intelligence Conference},
  pages = 	 {552--562},
  year = 	 {2020},
  editor = 	 {Adams, Ryan P. and Gogate, Vibhav},
  volume = 	 {115},
  series = 	 {Proceedings of Machine Learning Research},
  month = 	 {22--25 Jul},
  publisher =    {PMLR},
  url = 	 {https://proceedings.mlr.press/v115/wang20d.html},
}

@inproceedings{liang2022m3vit,
author = {Liang, Hanxue and Fan, Zhiwen and Sarkar, Rishov and Jiang, Ziyu and Chen, Tianlong and Zou, Kai and Cheng, Yu and Hao, Cong and Wang, Zhangyang},
title = {M3ViT: mixture-of-experts vision transformer for efficient multi-task learning with model-accelerator co-design},
year = {2022},
isbn = {9781713871088},
publisher = {Curran Associates Inc.},
address = {Red Hook, NY, USA},
booktitle = {Proceedings of the 36th International Conference on Neural Information Processing Systems},
articleno = {2062},
numpages = {17},
location = {New Orleans, LA, USA},
series = {NIPS '22}
}

@INPROCEEDINGS{pavlitskaya2020moesemseg,
  author={Pavlitskaya, Svetlana and Hubschneider, Christian and Weber, Michael and Moritz, Ruby and Hüger, Fabian and Schlicht, Peter and Zöllner, J. Marius},
  booktitle={2020 IEEE/CVF Conference on Computer Vision and Pattern Recognition Workshops (CVPRW)}, 
  title={Using Mixture of Expert Models to Gain Insights into Semantic Segmentation}, 
  year={2020},
  volume={},
  number={},
  pages={1399-1406},
  doi={10.1109/CVPRW50498.2020.00179}}

@article{raffel2020t5,
  author  = {Colin Raffel and Noam Shazeer and Adam Roberts and Katherine Lee and Sharan Narang and Michael Matena and Yanqi Zhou and Wei Li and Peter J. Liu},
  title   = {Exploring the Limits of Transfer Learning with a Unified Text-to-Text Transformer},
  journal = {Journal of Machine Learning Research},
  year    = {2020},
  volume  = {21},
  number  = {140},
  pages   = {1--67},
  url     = {http://jmlr.org/papers/v21/20-074.html}
}

@Article{lu2025uncertaintymoe,
AUTHOR = {Lu, Qiuye and Zhao, Wenzhi and Chen, Jiage and Chen, Xuehong and Zhang, Liqiang},
TITLE = {Uncertainty Mixture of Experts Model for Long Tail Crop Type Mapping},
JOURNAL = {Remote Sensing},
VOLUME = {17},
YEAR = {2025},
NUMBER = {22},
ARTICLE-NUMBER = {3752},
URL = {https://www.mdpi.com/2072-4292/17/22/3752},
ISSN = {2072-4292},
DOI = {10.3390/rs17223752}
}

@article{ding2025cddatascarce,
   title={A Survey of Sample-Efficient Deep Learning for Change Detection in Remote Sensing: Tasks, strategies, and challenges},
   volume={13},
   ISSN={2473-2397},
   url={http://dx.doi.org/10.1109/MGRS.2025.3533605},
   DOI={10.1109/mgrs.2025.3533605},
   number={3},
   journal={IEEE Geoscience and Remote Sensing Magazine},
   publisher={Institute of Electrical and Electronics Engineers (IEEE)},
   author={Ding, Lei and Hong, Danfeng and Zhao, Maofan and Chen, Hongruixuan and Li, Chenyu and Deng, Jie and Yokoya, Naoto and Bruzzone, Lorenzo and Chanussot, Jocelyn},
   year={2025},
   month=sep, pages={164–189} }

@article{peng2025cdoptical,
title = {Deep learning change detection techniques for optical remote sensing imagery: Status, perspectives and challenges},
journal = {International Journal of Applied Earth Observation and Geoinformation},
volume = {136},
pages = {104282},
year = {2025},
issn = {1569-8432},
doi = {https://doi.org/10.1016/j.jag.2024.104282},
url = {https://www.sciencedirect.com/science/article/pii/S1569843224006381},
author = {Daifeng Peng and Xuelian Liu and Yongjun Zhang and Haiyan Guan and Yansheng Li and Lorenzo Bruzzone},
}

@article{dou2021tsclassification,
title = {Time series remote sensing image classification framework using combination of deep learning and multiple classifiers system},
journal = {International Journal of Applied Earth Observation and Geoinformation},
volume = {103},
pages = {102477},
year = {2021},
issn = {1569-8432},
doi = {https://doi.org/10.1016/j.jag.2021.102477},
url = {https://www.sciencedirect.com/science/article/pii/S0303243421001847},
author = {Peng Dou and Huanfeng Shen and Zhiwei Li and Xiaobin Guan}
}

@Article{jia2018urbanlanduse,
AUTHOR = {Jia, Yuanxin and Ge, Yong and Ling, Feng and Guo, Xian and Wang, Jianghao and Wang, Le and Chen, Yuehong and Li, Xiaodong},
TITLE = {Urban Land Use Mapping by Combining Remote Sensing Imagery and Mobile Phone Positioning Data},
JOURNAL = {Remote Sensing},
VOLUME = {10},
YEAR = {2018},
NUMBER = {3},
ARTICLE-NUMBER = {446},
URL = {https://www.mdpi.com/2072-4292/10/3/446},
ISSN = {2072-4292},
DOI = {10.3390/rs10030446}
}

@INPROCEEDINGS{he2016resnet,
  author={He, Kaiming and Zhang, Xiangyu and Ren, Shaoqing and Sun, Jian},
  booktitle={2016 IEEE Conference on Computer Vision and Pattern Recognition (CVPR)}, 
  title={Deep Residual Learning for Image Recognition}, 
  year={2016},
  volume={},
  number={},
  pages={770-778},
  doi={10.1109/CVPR.2016.90}}

@INPROCEEDINGS{long2015fcn,
  author={Long, Jonathan and Shelhamer, Evan and Darrell, Trevor},
  booktitle={2015 IEEE Conference on Computer Vision and Pattern Recognition (CVPR)}, 
  title={Fully convolutional networks for semantic segmentation}, 
  year={2015},
  volume={},
  number={},
  pages={3431-3440},
  doi={10.1109/CVPR.2015.7298965}}

@article{roberts2017cv,
author = {Roberts, David R. and Bahn, Volker and Ciuti, Simone and Boyce, Mark S. and Elith, Jane and Guillera-Arroita, Gurutzeta and Hauenstein, Severin and Lahoz-Monfort, José J. and Schröder, Boris and Thuiller, Wilfried and Warton, David I. and Wintle, Brendan A. and Hartig, Florian and Dormann, Carsten F.},
title = {Cross-validation strategies for data with temporal, spatial, hierarchical, or phylogenetic structure},
journal = {Ecography},
volume = {40},
number = {8},
pages = {913-929},
doi = {https://doi.org/10.1111/ecog.02881},
url = {https://nsojournals.onlinelibrary.wiley.com/doi/abs/10.1111/ecog.02881},
eprint = {https://nsojournals.onlinelibrary.wiley.com/doi/pdf/10.1111/ecog.02881},
year = {2017}
}

@article{valavi2019blockcv,
author = {Valavi, Roozbeh and Elith, Jane and Lahoz-Monfort, José J. and Guillera-Arroita, Gurutzeta},
title = {blockCV: An r package for generating spatially or environmentally separated folds for k-fold cross-validation of species distribution models},
journal = {Methods in Ecology and Evolution},
volume = {10},
number = {2},
pages = {225-232},
doi = {https://doi.org/10.1111/2041-210X.13107},
url = {https://besjournals.onlinelibrary.wiley.com/doi/abs/10.1111/2041-210X.13107},
eprint = {https://besjournals.onlinelibrary.wiley.com/doi/pdf/10.1111/2041-210X.13107},
year = {2019}
}

@article{demsar2006stat,
author = {Dem\v{s}ar, Janez},
title = {Statistical Comparisons of Classifiers over Multiple Data Sets},
year = {2006},
issue_date = {12/1/2006},
publisher = {JMLR.org},
volume = {7},
issn = {1532-4435},
journal = {J. Mach. Learn. Res.},
month = dec,
pages = {1–30},
numpages = {30}
}

@inproceedings{dror2018hitchhiker,
    title = "The Hitchhiker{'}s Guide to Testing Statistical Significance in Natural Language Processing",
    author = "Dror, Rotem  and
      Baumer, Gili  and
      Shlomov, Segev  and
      Reichart, Roi",
    editor = "Gurevych, Iryna  and
      Miyao, Yusuke",
    booktitle = "Proceedings of the 56th Annual Meeting of the Association for Computational Linguistics (Volume 1: Long Papers)",
    month = jul,
    year = "2018",
    address = "Melbourne, Australia",
    publisher = "Association for Computational Linguistics",
    url = "https://aclanthology.org/P18-1128/",
    doi = "10.18653/v1/P18-1128",
    pages = "1383--1392"
}

@INPROCEEDINGS{selvaraju2017gradcam,
  author={Selvaraju, Ramprasaath R. and Cogswell, Michael and Das, Abhishek and Vedantam, Ramakrishna and Parikh, Devi and Batra, Dhruv},
  booktitle={2017 IEEE International Conference on Computer Vision (ICCV)}, 
  title={Grad-CAM: Visual Explanations from Deep Networks via Gradient-Based Localization}, 
  year={2017},
  volume={},
  number={},
  pages={618-626},
  doi={10.1109/ICCV.2017.74}}

@inproceedings{sundararajan2017axiomatic,
author = {Sundararajan, Mukund and Taly, Ankur and Yan, Qiqi},
title = {Axiomatic attribution for deep networks},
year = {2017},
publisher = {JMLR.org},
booktitle = {Proceedings of the 34th International Conference on Machine Learning - Volume 70},
pages = {3319–3328},
numpages = {10},
location = {Sydney, NSW, Australia},
series = {ICML'17}
}

@inproceedings{ribeiro2016whytrust,
author = {Ribeiro, Marco Tulio and Singh, Sameer and Guestrin, Carlos},
title = {{Why Should I Trust You?}: Explaining the Predictions of Any Classifier},
year = {2016},
isbn = {9781450342322},
publisher = {Association for Computing Machinery},
address = {New York, NY, USA},
url = {https://doi.org/10.1145/2939672.2939778},
doi = {10.1145/2939672.2939778},
booktitle = {Proceedings of the 22nd ACM SIGKDD International Conference on Knowledge Discovery and Data Mining},
pages = {1135–1144},
numpages = {10},
location = {San Francisco, California, USA},
series = {KDD '16}
}

@inproceedings{guo2017calibration,
author = {Guo, Chuan and Pleiss, Geoff and Sun, Yu and Weinberger, Kilian Q.},
title = {On calibration of modern neural networks},
year = {2017},
publisher = {JMLR.org},
booktitle = {Proceedings of the 34th International Conference on Machine Learning - Volume 70},
pages = {1321–1330},
numpages = {10},
location = {Sydney, NSW, Australia},
series = {ICML'17}
}

@inproceedings{BLIP-2,
author = {Li, Junnan and Li, Dongxu and Savarese, Silvio and Hoi, Steven},
title = {{BLIP-2}: bootstrapping language-image pre-training with frozen image encoders and large language models},
year = {2023},
publisher = {JMLR.org},
booktitle = {Proceedings of the 40th International Conference on Machine Learning},
articleno = {814},
numpages = {13},
location = {Honolulu, Hawaii, USA},
series = {ICML'23}
}

@ARTICLE{7891544,
  author={Cheng, Gong and Han, Junwei and Lu, Xiaoqiang},
  journal={Proceedings of the IEEE}, 
  title={Remote Sensing Image Scene Classification: Benchmark and State of the Art}, 
  year={2017},
  volume={105},
  number={10},
  pages={1865-1883},
  doi={10.1109/JPROC.2017.2675998}}

@INPROCEEDINGS{MDCS,
  author={Zhao, Qihao and Jiang, Chen and Hu, Wei and Zhang, Fan and Liu, Jun},
  booktitle={2023 IEEE/CVF International Conference on Computer Vision (ICCV)}, 
  title={{MDCS}: More Diverse Experts with Consistency Self-distillation for Long-tailed Recognition}, 
  year={2023},
  volume={},
  number={},
  pages={11563-11574},
  doi={10.1109/ICCV51070.2023.01065}}

@inbook{ovadia2019trust,
author = {Ovadia, Yaniv and Fertig, Emily and Ren, Jie and Nado, Zachary and Sculley, D. and Nowozin, Sebastian and Dillon, Joshua V. and Lakshminarayanan, Balaji and Snoek, Jasper},
title = {Can you trust your model's uncertainty? evaluating predictive uncertainty under dataset shift},
year = {2019},
publisher = {Curran Associates Inc.},
address = {Red Hook, NY, USA},
booktitle = {Proceedings of the 33rd International Conference on Neural Information Processing Systems},
articleno = {1254},
numpages = {12}
}
\end{fullwidth}








\end{document}